\documentclass{article}
\usepackage[table]{xcolor}
\definecolor{xxpurple}{RGB}{128, 0, 128}
\usepackage[preprint]{neurips_2026}
\usepackage[utf8]{inputenc} 
\usepackage[T1]{fontenc}    
\usepackage[dvipsnames]{xcolor}
\definecolor{Gred}{RGB}{219, 50, 54}
\definecolor{Ggreen}{RGB}{60, 186, 84}
\definecolor{Gblue}{RGB}{72, 133, 237}
\definecolor{Gyellow}{RGB}{247, 178, 16}
\definecolor{ToCgreen}{RGB}{0, 128, 0}
\definecolor{myGold}{RGB}{231,141,20}
\definecolor{myBlue}{rgb}{0.19,0.41,.65}
\definecolor{myPurple}{RGB}{175,0,124}
\usepackage[hyperfootnotes=false]{hyperref}
\hypersetup{
  colorlinks=true,
  citecolor=black,
  linkcolor=Sepia,
  filecolor=Gred
}
\usepackage{url}            
\usepackage{booktabs}       
\usepackage{amsfonts}       
\usepackage{nicefrac}       
\usepackage{microtype}      
\usepackage{xcolor}         
\usepackage{graphicx}
\usepackage{subcaption}
\usepackage[most]{tcolorbox}
\usepackage{enumitem}
\usepackage{xcolor,xspace,soul}
\usepackage[textsize=tiny]{todonotes}
\usepackage{tikzlings}
\usepackage{wrapfig}
\newcommand{\mask}{\mathbf{m}}
\def\rvx{{\mathbf{x}}}
\def\rvy{{\mathbf{y}}}
\def\rvz{{\mathbf{z}}}

\usepackage{amsmath}
\usepackage{amssymb}
\usepackage{mathtools}
\usepackage{amsthm}
\usepackage{lipsum}
\usepackage{multirow}
\usepackage{algorithm}
\usepackage{algpseudocode}

\definecolor{xblue}{HTML}{4169E1}
\definecolor{xgreen}{HTML}{036C3A}
\definecolor{xpurple}{HTML}{9838B1}
\definecolor{xslategray}{HTML}{70818F}
\definecolor{xorange}{HTML}{FF8C00}
\definecolor{xcyan}{HTML}{06AEEF}
\definecolor{xred}{HTML}{FF0000}
\definecolor{xgray}{HTML}{808080}
\definecolor{xxgreen}{HTML}{009F86}
\definecolor{xsienna}{HTML}{8B4512}
\definecolor{xxgreen}{HTML}{009F86}
\definecolor{xxpurple}{HTML}{623E99}
\definecolor{xolive}{HTML}{556B2F}

\newcommand{\xblue}[1]{\textcolor{xblue}{#1}}

\newcommand{\xxpurple}[1]{\textcolor{xxpurple}{#1}}
\newcommand{\xxgreen}[1]{\textcolor{xxgreen}{#1}}
\newcommand{\xred}[1]{\textcolor{xred}{#1}}

\newcommand{\coloredul}[2]{\textcolor{#1}{\underline{#2}}\xspace}

\title{From Interface to Inference: Eliciting Any-Order Inference from Any-Order Models}
\author{
Seunggeun Kim$^{1}$$^{\star}$\quad
Jaeyeon Kim$^{2}$$^{\star}$\quad
Taekyun Lee$^{1}$$^{\star}$\quad
Yuyuan Chen$^{2}$$^{\star}$\AND
Yilun Du$^{2}$\quad
Sham Kakade$^{2}$\quad
Sitan Chen$^{2}$
\AND
{\normalfont $^{1}$University of Texas at Austin,\,\,\, $^{2}$Harvard University,\,\,\,$^{\star}$Co-first authors}
}
\usepackage[capitalize,noabbrev]{cleveref}
\begin{document}
\maketitle
\begin{abstract}
Many discrete reasoning tasks, such as code generation, are inherently non-causal: programmers move between high-level structure and local details, a process we call \emph{any-order inference}. For autoregressive language models, which lack a native any-order interface, non-causal abilities such as infilling and next-edit prediction require hand-designed mechanisms. Can we instead design models that natively support any-order inference? Masked diffusion models have recently emerged as compelling candidates, as their any-order training objective naturally offers an any-order prediction interface. This \emph{interface}, however, does not automatically yield any-order \emph{inference}. We demonstrate that this interface-inference gap stems from \emph{positional uncertainty}: fixed-canvas, token-level models may know \emph{what} semantic component should appear without knowing \emph{where} to place it. In light of this, we propose two complementary approaches: (1) \emph{Insertion-based masked diffusion}, building on FlexMDM~\citep{kim2025any}, relaxes fixed-position commitments via insertions, enabling generation across non-contiguous regions. (2) \emph{Latent-space masked diffusion} shifts prediction to coarser semantic segments, enabling search over latent generation orders. Empirically, we train a 7B FlexMDM for Python coding and a 125M LatentMDM for GSM8K and show that both approaches induce distinct any-order inference behaviors and improve downstream performance. We release our codebase at \url{https://github.com/SeunggeunKimkr/genuine-any-order}.
\end{abstract}

\begin{figure}[h]
\centering
\vspace{-0.1in}
\includegraphics[width=1.0\linewidth]{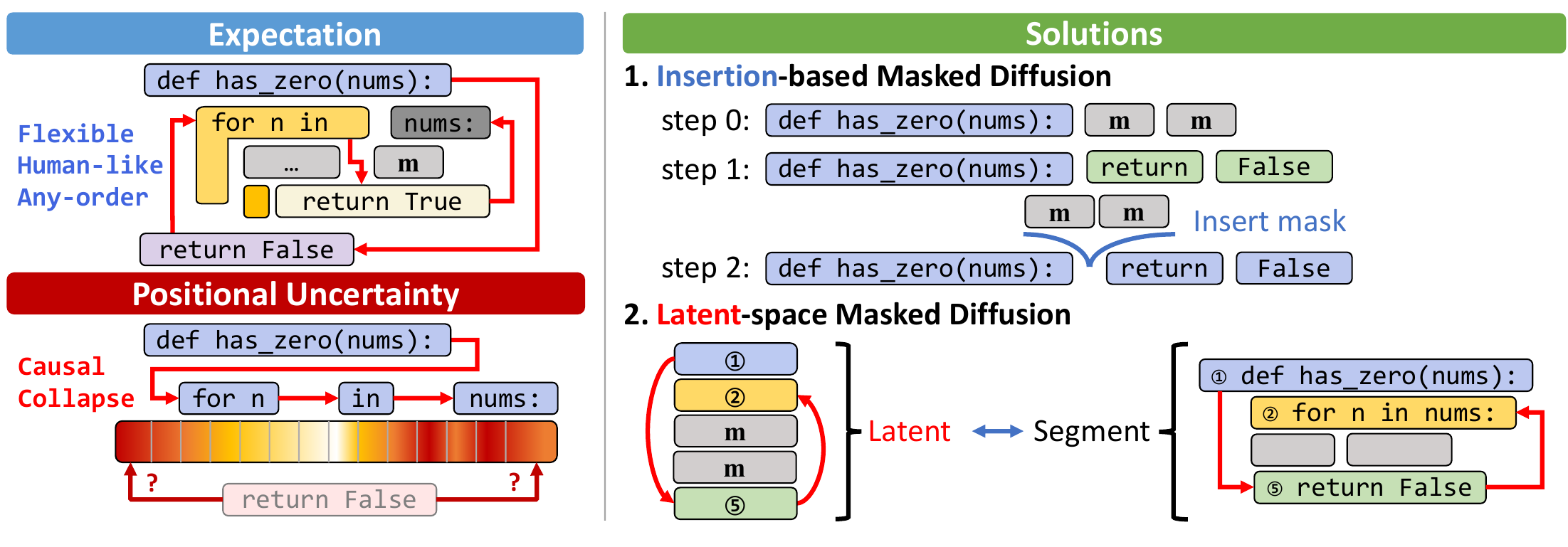}
\caption{\textbf{Illustrative overview of our work.} Masked diffusion models cannot perform genuine \emph{any-order} inference as they operate on a fixed canvas at the token level. We demonstrate that insertion-based and latent-space masked diffusion modeling overcome this limitation, enabling genuine \emph{any-order} inference.}
\label{fig:main}
\end{figure}
\section{Introduction}
Many real-world discrete reasoning tasks are inherently non-causal: when humans construct complex discrete objects such as code, they revise earlier decisions, insert missing pieces, and alternate between unfolding high-level structure and filling in local details. Autoregressive language models, the dominant paradigm in discrete generative modeling, decode left-to-right and thus do not natively support such trajectories. A substantial line of work has nevertheless extended them with non-causal mechanisms such as infilling~\citep{bavarian2022efficient}, reordering~\citep{shah2024causal}, and next-edit prediction~\citep{lu2025next}. While this shows that any-order inference is a desirable capability, the task-specific nature of these approaches motivates more native and principled generative paradigms.

We focus on masked diffusion models (MDMs)~\citep{shi2024simplified,sahoo2024simple}, a popular instantiation of diffusion language models~\citep{nie2025large,xie2025dream,gong2025diffucoder,bie2025llada20scalingdiffusionlanguage}. MDMs are natural candidates for any-order inference as their training objective endows them with an any-order prediction \emph{interface}~\citep{ou2024your,zheng2024masked,kim2025train}: given a partially masked sequence, they predict clean tokens at arbitrary masked positions. This interface makes any-order inference possible in principle, but not automatic: the decoding policy must still choose which positions to reveal at each step, and poor choices can collapse generation order back toward left-to-right completion. We therefore ask whether MDMs actually deliver on this promise: \emph{can an any-order prediction interface yield inference that is genuinely any-order in nature?}

\textbf{Our contributions.} In this work, we focus on code generation, a domain that would naturally benefit from genuine any-order inference. Our first finding is negative. We show that under confidence-based decoding, the dominant inference-time strategy for improving downstream performance, MDMs produce code that is \emph{algorithmically close} to what they would have generated via strict left-to-right decoding (Section~\ref{sec:similarity}). This resonates with recent observations of \emph{causal collapse}~\citep{ni2026flexibility,gong2025diffucoder,li2026diffusion}, but we identify a more fundamental bottleneck underlying this phenomenon: \emph{positional uncertainty}. 

Positional uncertainty arises when the model knows \emph{what} semantic component should be generated next, but remains uncertain about \emph{where} to place it: its probability mass is spread across multiple plausible locations on the fixed masked canvas. Since confidence-based unmasking ranks individual positions, these globally meaningful but \emph{positionally dispersed} components appear less confident than locally determined continuations, pushing decoding back toward left-to-right completion. These findings (Section~\ref{sec:pos_uncer}) suggest that the any-order training objective of masked diffusion models does not guarantee genuine any-order inference \emph{per se}. This motivates a broader question:
\vspace{-0.08in}
\begin{center}
\xblue{\textbf{What generative models can achieve genuine any-order inference?}}
\end{center}
\vspace{-0.08in}
In this work, we propose two complementary remedies, each sidestepping positional uncertainty in a different way and inducing \emph{distinct forms of any-order inference}. 
Our first solution, \xblue{\textbf{\emph{insertion-based masked diffusion}}} (Section~\ref{sec:flexmdm}), builds on FlexMDM~\citep{kim2025any}, which equips MDMs with the ability to insert tokens. 
This circumvents the fixed-canvas bottleneck: unmasking no longer commits a token to a specific final position, since later insertions can shift its location in the final sequence. 
With this reduced token-commitment burden, we demonstrate that FlexMDM yields a structural form of any-order inference: under the parsed program tree, generation \emph{\xxpurple{\textbf{moves back and forth across different nodes}}} rather than completing one contiguous region at a time. 
Empirically, we achieve this by significantly extending the practical scope of~\citet{kim2025any}: we fine-tune Dream-Coder 7B~\citep{xie2025dream} into a general-purpose Python-code-generating FlexMDM, which outperforms Dream-Coder 7B on HumanEval and HumanEval+~\citep{chen2021evaluating} at Pass@16, and moves ahead of it at Pass@1 on MBPP and MBPP+~\citep{austin2021program} while remaining on par overall.

Our second solution, \xblue{\textbf{\emph{latent-space masked diffusion} (LatentMDM)}} (Section~\ref{sec:lmdm}), overcomes positional uncertainty by moving masked diffusion from token space to a continuous latent space. LatentMDM is distinct from prior embedding-space approaches, such as the Flow Map Language Model~\citep{lee2026flow}, and constitutes our main modeling contribution. Rather than selecting individual token positions, the model decides which masked latent segment to decode next, shifting the interface to a coarser semantic granularity. 
This enables a distinct form of any-order inference: \xxpurple{\emph{\textbf{searching over semantic generation orders in latent space}}}, allowing the model to discover orderings beyond the left-to-right token order. 
Unlike FlexMDM, LatentMDM has no off-the-shelf pretrained base model. We therefore pretrain a 125M-parameter latent masked diffusion model on TinyGSM~\citep{liu2023tinygsm} and show that it outperforms standard MDMs, embedding-space models such as DUO~\citep{sahoo2025diffusion} and $\mathbb{S}$-FLM~\citep{deschenaux2026language}, and even an autoregressive model with KV caching under the same inference budget.
\vspace{-0.10in}
\section{Preliminaries} \label{sec:prelim}
\vspace{-0.05in}
In this section, we review masked diffusion models (MDMs)~\citep{shi2024simplified,sahoo2024simple}.

\textbf{Notation.} Suppose our goal is to learn to generate samples from the data distribution $\rvx\sim p_{\mathrm{data}}$ over length-$L$ discrete sequences with a finite vocabulary $\mathcal{V}$. Let $\rvx^i$ denote the $i$-th element of a given sequence $\rvx=(\rvx^1, \dots, \rvx^L)$ and $\Delta(\mathcal{V})$ indicate the simplex of probability distributions over $\mathcal{V}$.

\textbf{Training.} Although MDMs admit several interpretations, we adopt an \emph{any-order} language model interpretation~\citep{ou2024your,zheng2024masked}, which streamlines the prior account of MDM~\citep{sahoo2024simple,shi2024simplified}. Roughly speaking, MDMs introduce an auxiliary mask token $\mask$ and learn, for each masked position, the posterior marginals of clean tokens conditioned on a masked sequence. To learn this posterior during training, one draws a clean sequence $\rvx\sim p_{\mathrm{data}}$ and constructs a partially masked sequence $\rvz$ as follows: Sample $n\sim\mathrm{Unif}\{0,\dots,L\}$ and replace the tokens at uniformly selected $n$ indices in $\rvx$ with $\mask$. Hence, a resulting $\rvz$ has $n$ (randomly drawn) masked indices.

This masking procedure induces a joint distribution over $(\rvx,\rvz)$ and we refer to the conditional marginal $\mathrm{law}(\rvx^i\mid \rvz)$ as the \emph{unmasking posterior}. This unmasking posterior is the central object in MDMs and is modeled by a neural network $f_\theta$ that takes $\rvz$ as input and outputs a $|\mathcal{V}|\times L$ matrix. Concretely, its $i$-th column, $f_\theta^i(\cdot\,|\,\rvz)\in\Delta(\mathcal{V})$, models the unmasking posterior $f_\theta^i(v\,|\,\rvz) \approx p(\rvx^i=v\,|\,\rvz)$. To train $f_\theta$, we minimize cross-entropy loss summed over all masked indices.

\vspace{-0.12in}
{\small
\begin{equation*}
    \mathcal{L}(\theta)\colon =  \mathbb{E}_{\rvx,\rvz}\left[\frac{1}{n}\sum_{i\colon \rvz^i=\mask} -\log f_\theta^i(\rvx^i \,|\,\rvz) \right].
\end{equation*}}
\vspace{-0.12in}

\textbf{MDM inference.}  A key feature of MDMs is their \emph{any-order objective}: a pretrained MDM has learned to predict the posterior of \emph{any masked position} in a sequence. This any-order property builds upon earlier work in masked language modeling~\citep{devlin2019bert,ghazvininejad2019mask,wang2019bert}. Next, we explain how this any-order property translates to the flexibility at inference time.

MDM inference starts from a length-$L$ masked sequence $\rvx_1 = (\mask,\dots,\mask)$ or generally a given prompt $\rvx_1 = ([\texttt{prompt}],\mask,\dots,\mask)$, and proceeds over a monotonically decreasing time grid $t_0=1>\dots>t_N = 0$. At each step $t_\ell$, given a partially masked sequence $\rvx_{t_\ell}\in(\mathcal{V}\cup \{\mask\})^L$, we proceed in two steps to obtain $\rvx_{t_{\ell+1}}$: \textbf{(a)} Choose a subset of masked positions $\mathcal{S}\subseteq\{i\,|\,\rvx_{t_\ell}^i=\mask\}$ and \textbf{(b)} For each $i \in \mathcal{S}$, unmask $\rvx_{t_\ell}^i$ to a clean token $v$ sampled from $v\sim f_\theta^i(\cdot\,|\, \rvx_{t_\ell})\in\Delta(\mathcal{V})$.
Notably, as the MDM training is any-order, i.e., $f_\theta$ predicts the clean token distribution over all masked positions, there is flexibility in the choice of $\mathcal{S}$, which is central to the downstream performance. Below, we describe several strategies for choosing $\mathcal{S}$ in practice.

Training-free methods, including~\cite{zheng2024masked,kim2025train,peng2025pathplanningmaskeddiffusion,ben2025accelerated,nie2025large,wu2025fast,hayakawa2025demystifying}, typically first compute a confidence score for each masked position and select a set $\mathcal{S}$ of the positions with the highest-scoring indices, i.e.,
$\mathcal{S}\gets\mathrm{TopK}_{i:\,\rvx_t^i=\mask}\left[\mathrm{score}(i)\right]$,
where $\mathrm{score}(i)$ quantifies how certain the model prediction is at position $i$.

Common recipes for $\mathrm{score}(i)$ include the maximum predicted probability
$\max_{v\in\mathcal{V}} f_\theta^i(v\,|\,\rvx_t)$, the margin between the top two probabilities
$f_\theta^i(v_1\,|\,\rvx_t)-f_\theta^i(v_2\,|\,\rvx_t)$ (where $v_1,v_2$ are the top-2 tokens with the highest predicted probabilities), and the negative entropy of the categorical distribution. Some variants further incorporate an explicit position-dependent bias toward leftmost positions, often referred to as semi-autoregressive decoding. We refer to this broad family of training-free rules as \emph{confidence-based decoding}.

\textbf{Diffusion large language models.} After the advent of MDMs, they have been effectively scaled from 7B-parameter~\citep{nie2025large,ye2025dream,xie2025dream,song2025seed,gong2025diffucoder} to 100B~\citep{bie2025llada20scalingdiffusionlanguage,bie2026llada2} and industry-scale~\citep{gemini2025diffusion,labs2025mercury} and are often referred to as diffusion language models. Although the exact details of closed-source industrial models are not fully disclosed, all open-source diffusion language models considered in this work instantiate the masked diffusion modeling framework.
\vspace{-0.10in}
\section{The Interface-to-Inference Gap}
\vspace{-0.05in}
As discussed in the introduction, many discrete objects, such as code, admit natural non-left-to-right generation orders. MDMs are appealing candidates for modeling such non-causal procedures due to their \emph{any-order} interface. So far, however, this inference-time flexibility has primarily been used for efficient parallel decoding~\citep{gemini2025diffusion,labs2025mercury,wu2025fast2,wu2025fast,song2025seed}. These works show that MDMs can exploit parallelism, but leave open a more fundamental question: \emph{do MDMs actually exploit any-order flexibility to reason more effectively?}

Prior work gives positive evidence in restricted settings: on small, highly structured domains such as logical puzzles, MDMs can exploit their any-order interface to follow non-autoregressive generation orders and outperform autoregressive baselines~\citep{ye2024beyond,kim2025train,trainin2026discrete}. At larger scales, however, recent empirical work paints a less optimistic picture: with confidence-based decoding, the dominant inference strategy in practice, MDMs exhibit a form of \emph{causal collapse} on standard math and coding benchmarks. In particular, the token generation order remains close to autoregressive. These findings suggest that any-order training alone \xxgreen{\textbf{does not guarantee genuinely any-order inference}} once MDMs are scaled.

\textbf{Comparison to prior work.}
It is well known that, despite its flexibility in generation order, confidence-based MDM sampling often produces less diverse outputs than fixed left-to-right sampling with a nonzero token-sampling temperature~\citep{ni2026flexibility, olausson2026tale, shen2026improving, lamont2026free, fang2026locally,patel2025improved}. Degradation of diversity is harmful in Pass@K evaluation and post-training, where multiple diverse rollouts provide valuable search coverage or learning signals.

Most directly, \citet{ni2026flexibility} studies why confidence-based any-order decoding has lower entropy than AR-like decoding. Their explanation is that the flexibility of any-order interface allows the model to repeatedly select low-entropy tokens, thereby committing early to low-entropy trajectories. This question, therefore, is distinct from the causal-collapse phenomenon studied in our work: they ask why confidence-based decoding is less diverse than autoregressive, whereas we ask why confidence-based decoding, despite its \emph{any-order interface}, fails to go meaningfully beyond AR-like decoding.

A separate line of work more directly investigates causal collapse by examining why MDM generation orders frequently become left-to-right~\citep{gong2025diffucoder, li2026diffusion}. These works attribute the phenomenon to causal biases in the pretraining data: under this view, confidence-based decoding recovers a \emph{natural} ordering already present in the data, which then appears causal. Although plausible, this explanation remains incomplete. In particular, it does not readily explain why the same collapse persists in code generation, where Python programs often do not possess a unique inherent causal ordering.

Our work provides a mechanistic account of causal collapse. We quantify precisely how closely confidence-based MDM inference structurally approximates causal decoding on Python programs and investigate why this approximation arises. We trace the collapse not merely to a preference inherited from the training distribution, but to a more fundamental limitation of token-level, fixed-canvas MDM inference: \textbf{\xxgreen{\emph{positional uncertainty}}}.

\textbf{Summary of the results.} 
In Section~\ref{sec:similarity}, we first characterize the causal collapse from a \emph{structural} perspective. Beyond analyzing the generation order itself, using tree-based parsing of generated code, we show that confidence-based MDM decoding produces samples that are \xblue{\textbf{structurally similar}} to those generated in a strictly left-to-right manner, and that MDM's apparent flexible interface does not translate into inference that exploits the hierarchical structure of programs.
In Section~\ref{sec:pos_uncer}, we then formalize \emph{positional uncertainty} as the inevitable bottleneck that arises when token-level, fixed-canvas decoding must decide both \textbf{\xblue{\emph{what} should be generated}} and \textbf{\xblue{\emph{where} it should be placed}}. We show that this uncertainty persists across different MDMs and tasks.

\subsection{Semantic collapse in Python code generation}  \label{sec:similarity}
In this section, we demonstrate that confidence-based MDM decoding produces programs that are structurally similar to those generated by strict left-to-right (L2R) decoding. For each pretrained MDM, we hold the model fixed and vary only the unmasking policy. The L2R baseline reveals the leftmost remaining masked position at each step, while sampling its token from the same learned posterior $f_\theta^i(\cdot \mid \rvx_t)$. We then compare the final programs generated by each non-L2R policy against the corresponding L2R generations, using a structural similarity metric and best-match aggregation described below and detailed in Appendix~\ref{app:sim}.

\vspace{-0.10in}
\begin{figure}[h]
    \centering
    \includegraphics[width=\linewidth,trim={0, 2.0cm, 0, 2.0cm}]{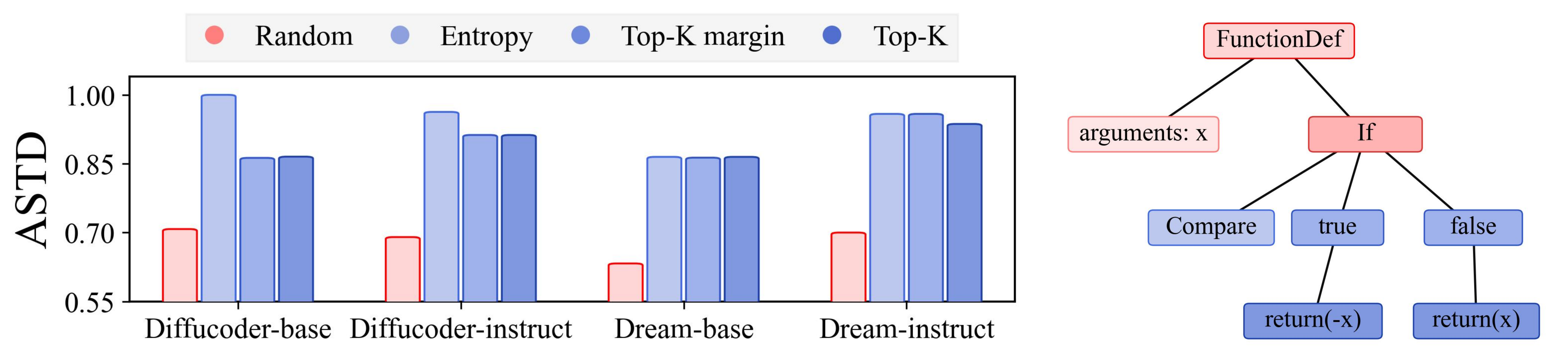}
\caption{
\textbf{(Left)} Confidence-based decoding yields code structurally close to left-to-right decoding.
\textbf{(Right)} A simplified AST of the python function that returns \texttt{-x} if \texttt{x < 0} and \texttt{x} otherwise.}
\label{fig:sim}
\vspace{-0.12in}
\end{figure}

\textbf{Code similarity metrics.} To compare program structure, we use a standard AST-based tree-edit comparison. We parse each generated program with Python’s built-in ast module, yielding an abstract syntax tree (see Figure~\ref{fig:sim} for example) whose nodes represent syntactic components such as functions, branches, loops, assignments, and returns. We then compute normalized tree-edit similarity between ASTs; higher values indicate more similar program structure. We refer to this score as \emph{abstract syntax tree distance} (ASTD) and provide precise definitions, variants, and additional results in Appendix~\ref{app:sim}.

\textbf{Experiment setup.} We take several large-scale pretrained MDMs, including their base and instruction- or RL-fine-tuned variants; Dream-7B-Base/Instruct~\citep{xie2025dream}, DiffuCoder-7B-Base/Instruct/cpGRPO~\citep{gong2025diffucoder}, and Dream-Coder-7B-Instruct~\citep{xie2025dream}. On problems from HumanEval~\citep{chen2021evaluating}, MBPP~\citep{austin2021program}, and LiveCodeBench~\citep{jain2024livecodebench}, we generate samples with sequence length $256$ and $256$ steps, using token sampling temperature $0.2$ and nucleus sampling with $p=0.95$, which exactly match the configurations from~\cite{gong2025diffucoder}.
We compare three confidence-based policies—Top-$K$, Top-$K$ margin, and entropy-based decoding—against strict L2R decoding. All generations use temperature 0.2. For each prompt, we draw one sample from each confidence-based policy, since at this low temperature, confidence-based decoding produces little sample diversity. In contrast, we draw 32 samples from L2R and random decoding. For L2R, this gives a reference set for asking whether samples from other policies are already covered by L2R-biased generation; for random decoding, it provides a calibration baseline for how structurally diverse valid or correct programs can be when the reveal order has no causal bias. The larger sample budget also offsets the lower valid/correct rate of L2R and random decoding.

\textbf{Best-match aggregation and filters.} Since L2R is our reference for structurally left-to-right-biased generation, we examine whether each sample produced by a given policy has a close counterpart among the L2R samples for the same prompt. We thus use \emph{best-match aggregation}: for each generated program, we compute its similarity to the most similar L2R program, and then average these best-match scores across prompts. A high score means that the policy mostly discovers structures already covered by L2R decoding; a low score means that it finds programs structurally outside the L2R reference set. We compute this under two filters: valid–valid, where both programs parse, and success–success, where both programs pass the unit tests. Appendix~\ref{app:sim} gives the precise definition.

\textbf{Results.}
As shown in Figure~\ref{fig:sim}, confidence-based decoding methods (\coloredul{xblue}{blue bars}) remain substantially closer to the L2R reference set than the Random-vs-L2R baseline (\coloredul{xred}{red bar}). This trend holds across models and tasks, as shown in additional results in Appendix~\ref{app:sim}. Thus, \xblue{\textbf{the programs produced by confidence-based decoding share the same global skeleton and hierarchical structure}} as L2R-generated code. By contrast, the lower Random-vs-L2R similarity shows that the benchmarks admit structurally different valid or correct programs, but confidence-based MDM decoding tends not to discover them.

\vspace{-0.1in}
\subsection{Diagnosing Positional Uncertainty in MDMs} 
\label{sec:pos_uncer}
\vspace{-0.1in}
The observation in Section~\ref{sec:similarity} raises the next question about \emph{why} this causal collapse occurs. In this section, we argue that the key bottleneck lies in the fixed-canvas, token-level nature of MDMs.

\textbf{Example: the \emph{what} and \emph{where} of a return node.}
Consider a simple code snippet with return statements, illustrated in Figure~\ref{fig:main}. In many coding tasks, the model may infer early that the program needs a fallback return, such as \texttt{return False}. However, the correct location of this return statement can remain ambiguous before the surrounding structure is resolved: placing it inside a loop, inside a branch, or after the loop can lead to different programs. Thus, a model may be confident about \emph{what} semantic component should appear---for example, a fallback return node---while remaining uncertain about \emph{where} it should be placed. In a fixed-canvas MDM, this uncertainty \emph{spreads} the overall high probability mass for the return node across several plausible locations, resulting in relatively low mass at each of those locations. Therefore, confidence-based decoding tends to select nearby local continuations first and collapse toward a left-to-right ordering.

\begin{wrapfigure}[20]{r}{0.34\textwidth}
\vspace{-0.30in}
\centering
\begin{subfigure}{0.33\textwidth}
    \centering
    \includegraphics[width=1.05\textwidth,trim={2.0cm 0.0cm 0.5cm 0.3cm}]{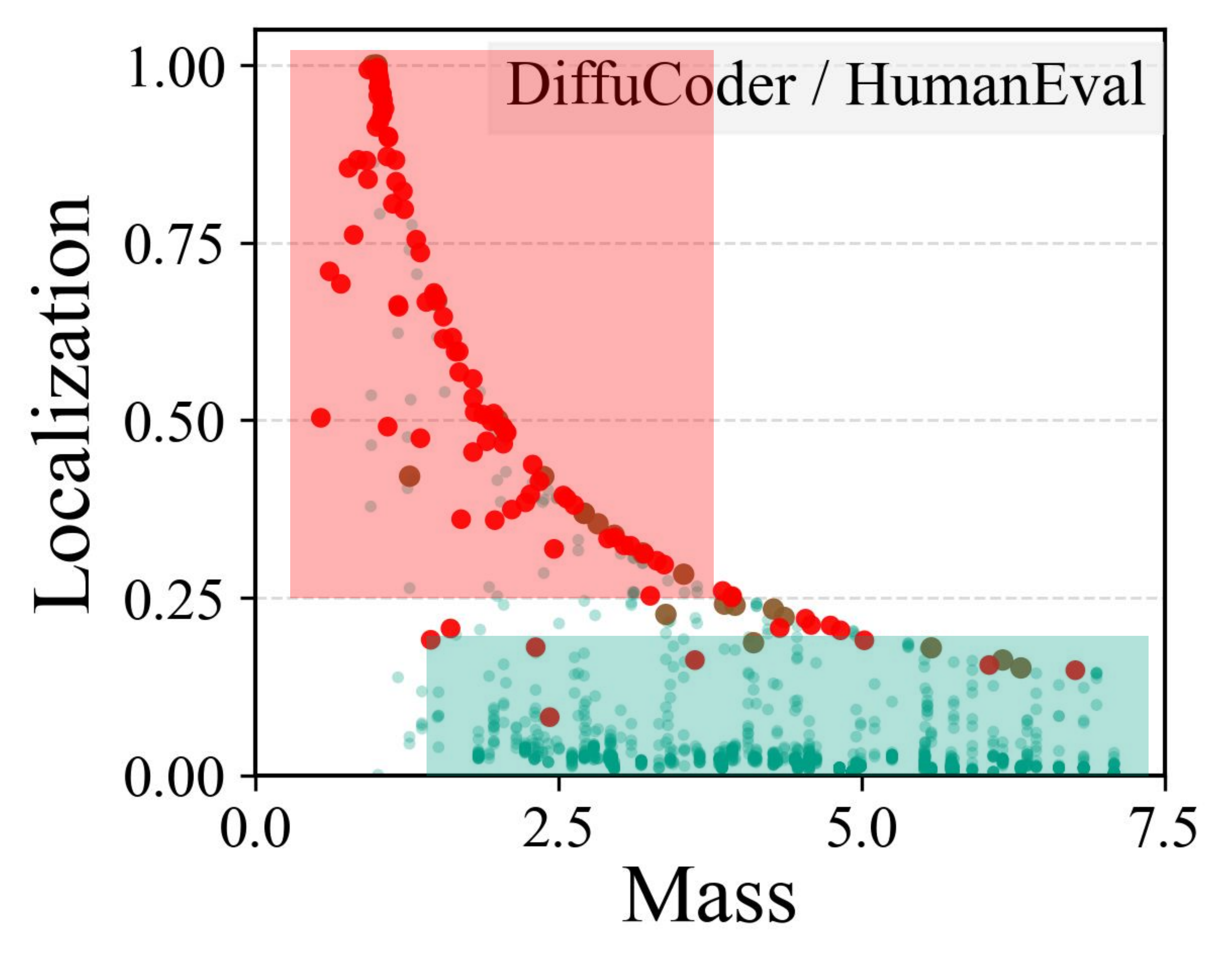}
\end{subfigure}
\begin{subfigure}{0.33\textwidth}
    \centering
    \includegraphics[width=1.05\textwidth,trim={2.0cm 1.5cm 0.5cm 0.3cm}]{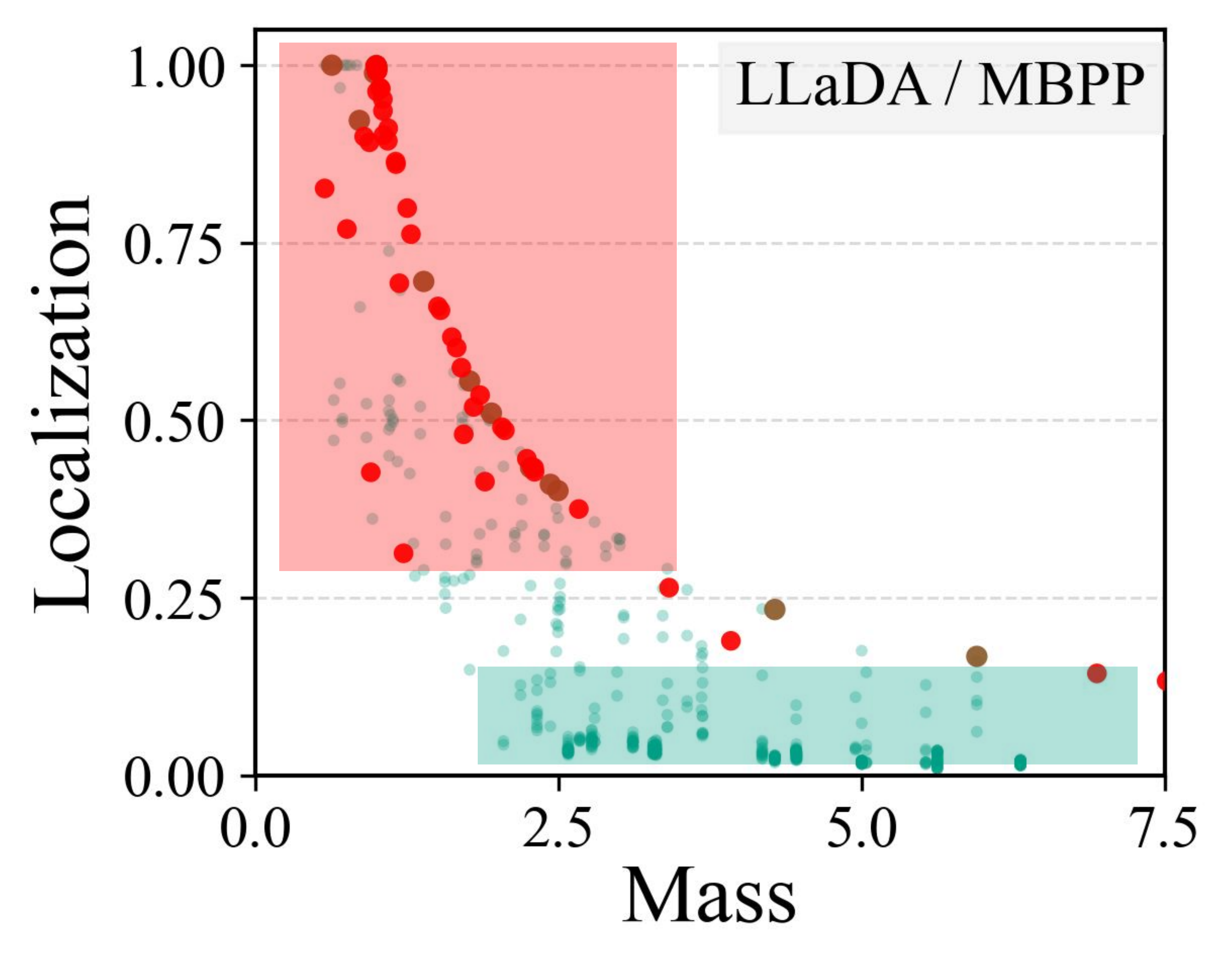}
\end{subfigure}
\vspace{-0.10in}
\caption{\textbf{Positional Uncertainty.}}
\label{fig:pos_uncert}
\end{wrapfigure}

\textbf{Positional uncertainty.} We now formalize the intuition for \emph{positional uncertainty} stated above. For a token $v$, we define its \emph{aggregate mass} as $m(v;\rvx_t) = \sum_{i:\,\rvx_t^i = \mask} f_\theta^i(v\,|\,\rvx_t)$. This measures the total probability assigned to token $v$ across all currently masked positions. To measure whether this mass is concentrated at a single location or dispersed across many plausible locations, we define the \emph{localization score}
\begin{equation*}
   \mathrm{LOC}(v;\rvx_t)\colon= \max_{i:\,\rvx_t^i=\mask} \left[\frac{f_\theta^i(v\,|\,\rvx_t)}{m(v;\rvx_t)}\right].
\end{equation*}
Thus, $\mathrm{LOC}(v;\cdot)$ is the largest fraction of token $v$’s aggregate mass assigned to any one masked position. High values indicate a clear placement; low values imply dispersal.
The case where the model knows \emph{what to generate but not where to place it} corresponds to a token $v$ with (1) high aggregated mass $m(v;\rvx_t)$ but (2) low $\mathrm{LOC}(v;\cdot)$.

\textbf{Results.} We examine whether positional uncertainty appears in practice. For each prompt and each partial sequence $\rvx_t$, we collect two types of points. First, to visualize tokens the model assigns high overall probability to, we take the top two tokens $v$ with the largest aggregate mass $m(v;\rvx_t)$ and plot $(m(v;\rvx_t), \mathrm{LOC}(v;\rvx_t))$ in \coloredul{xxgreen}{green}. Second, to visualize what the model actually commits, for each revealed token $v$ at position $i$, we plot its aggregate mass together with its committed-position localization $f_\theta^i(v\,|\,\rvx_t)/m(v;\rvx_t)$ in \coloredul{xred}{red}. We aggregate these points across prompts and provide additional plots in Appendix~\ref{app:pos_unc}.

The resulting pattern in Figure~\ref{fig:pos_uncert} supports our claim. Many high-mass tokens lie in the \coloredul{xxgreen}{lower-right} region of the plot, indicating that the model assigns them substantial total probability, but does not decode them due to low localization. In contrast, committed tokens tend to have lower aggregate mass but higher localization, appearing toward the \coloredul{xred}{upper-left} region.

\textbf{Similarity analysis for learned decoding strategies.}
A natural follow-up question is whether this bottleneck can be mitigated by post-training the unmasking policy of an MDM, as explored in~\citep{hong2025improving,jazbec2025learning,chen2025dultra,ye2024diffusion}. To examine this possibility, we take a checkpoint from~\citet{chen2025dultra},
which trains an unmasking policy model with reinforcement learning. 
We then repeat our similarity analysis using the planner-induced unmasking policy. Although the learned planner changes the distribution over decoding orders, our results in Appendix~\ref{app:dultra} show that the same trend persists: generated programs remain structurally close to those produced by strict left-to-right decoding. This suggests that, within fixed-canvas token-level MDMs, learning a decoding policy alone may be insufficient to escape causal collapse.

We do not claim this as an exhaustive evaluation of all planner-learning methods, since some of the public checkpoints are not available. Nevertheless, this result supports our broader view that positional uncertainty is a more fundamental limitation of fixed-canvas token-level MDMs for enabling genuine any-order inference, rather than merely a weakness of a particular confidence score.
\vspace{-0.10in}
\section{Circumventing Positional Uncertainty}
\vspace{-0.09in}
We identified positional uncertainty as a key bottleneck of MDMs. This suggests that genuine any-order inference may require relaxing at least one of two aspects of MDM generation: \textbf{\xxgreen{\emph{token-level}}} and \textbf{\xxgreen{\emph{fixed-canvas}}}.
In this section, we present two alternative generative models that each relax one of these aspects. We first summarize their modeling principles, then provide detailed descriptions and empirical evidence in Section~\ref{sec:flexmdm} and Section~\ref{sec:lmdm}.

\textbf{FlexMDM: relaxing fixed-canvas.} Our first approach is to employ FlexMDM~\citep{kim2025any}, an insertion-based masked diffusion model that not only decodes mask tokens but also inserts new masks during inference. This relaxes positional uncertainty since revealing a token no longer irrevocably commits it to a fixed position: later insertions can shift its location in the final sequence. 

This relaxation induces a structural form of \emph{any-order inference}. 
When we view generated code through its parsed algorithmic tree, we find that FlexMDM generation moves back and forth across different semantic nodes, rather than completing one contiguous region before moving to the next, thereby going beyond left-to-right generation.
In this sense, insertion gives a mechanism for drafting multiple semantic regions and refining them later, closer to how human programmers develop code.

\textbf{LatentMDM: relaxing token-level.} 
Our second approach moves the masked-diffusion interface from individual tokens to segment-level latent variables. We split each sequence into coarse segments, such as lines of code, and jointly train a lightweight encoder-decoder pair that maps between segments and latent vectors, together with a masked diffusion model over the sequence of segment latents.

At inference time, the resulting LatentMDM chooses which \emph{masked latent segment} to reveal next, and the decoder realizes the chosen latent as tokens. This switches the object of any-order inference from token positions to latent segments. By absorbing token-level positional uncertainty into segment-level latent variables, the revealed confidence becomes less tied to local token-position certainty and more tied to the compatibility of higher-level semantic components. \textbf{\xxgreen{Thus, the form of any-order reasoning realized by LatentMDM is semantic-order search in a more compact exploration space:}} deciding which high-level latent segment to instantiate next.

\textbf{Experiments.} We present a suite of experiments to answer the following questions:

\xblue{\texttt{Q1.}} \textbf{\emph{Do our solutions mitigate causal collapse by relaxing positional uncertainty?}}

\xblue{\texttt{Q2.}} \textbf{\emph{What new forms of any-order inference do our solutions enable?}}

\subsection{FlexMDM} \label{sec:flexmdm}
Now we describe the training and inference procedures of FlexMDM and present empirical results obtained by fine-tuning Dream-Coder 7B~\citep{xie2025dream}.

\textbf{Training.}
To enable insertion during training, FlexMDM augments the standard MDM masking procedure (recall Section~\ref{sec:prelim}) with \emph{deletion}. As a result, during training, FlexMDM observes not only masked but also deleted versions of sequences $\rvx \sim p_{\mathrm{data}}$. The masking and deletion rates are determined by two schedules: an unmasking and an insertion schedule. Given a corrupted sequence $\rvz$, a FlexMDM is trained to jointly predict the posterior distribution over masked tokens and the expected number of tokens to insert, conditioned on the current sequence $\rvz$.

\textbf{Inference.}
At inference time, given a trained FlexMDM and the corresponding schedules, generation proceeds by alternating between two operations: (1) selecting which masked tokens to reveal according to the unmasking schedule and model's predicted unmasking posterior, and (2) inserting additional mask tokens according to the model's predicted insertion counts. Importantly, FlexMDM retains the any-order flexibility of standard MDM: for unmasking, one can still use confidence-based decoding to decide which masked tokens to reveal. For insertion, we sample the number of mask tokens to insert from a Poisson distribution whose rate is determined by the training-time insertion schedule together with the model's prediction. Like unmasking, insertion carries its own inference-time temperature: the per-gap insertion rates factorize into a total count and a placement distribution, and tempering only the placement controls \emph{where} new masks open while preserving the expected generation length (Appendix~\ref{app:insertion_temp}). We defer technical details to Appendix~\ref{app:flexmdm_training}.

\vspace{-0.1in}
\begin{figure}[h]
\centering
\begin{subfigure}[t]{0.44\textwidth}
\centering
\vspace{0pt}
\includegraphics[
    width=\textwidth,
    trim={1.0cm 0 0 0},
    clip
]{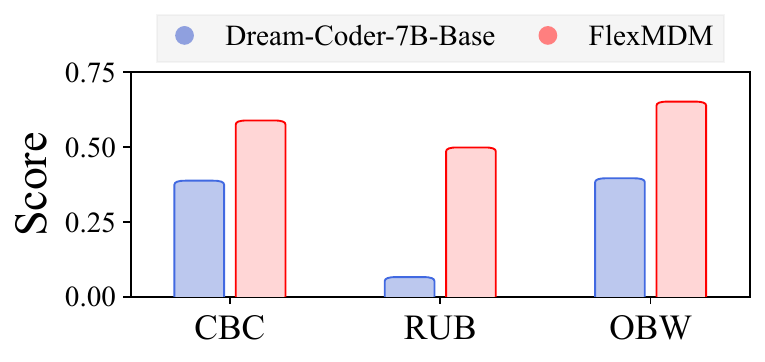}
\end{subfigure}
\begin{subfigure}[t]{0.54\textwidth}
\centering
\vspace{0.17in} 
\small
\setlength{\tabcolsep}{4pt}
\resizebox{\textwidth}{!}{
\begin{tabular}{llcccc}
\toprule
\textbf{Benchmark} 
& \textbf{Model}
& \textbf{@1} 
& \textbf{@4} 
& \textbf{@8} 
& \textbf{@16} \\
\midrule
HumanEval 
& Dream-Coder & 58.65 & 81.34 & 86.94 & 90.85 \\
& FlexMDM     & 50.65 & 78.69 & 86.86 & 92.07 \\
\midrule
HumanEval+
& Dream-Coder & 53.89 & 75.96 & 81.53 & 85.37 \\
& FlexMDM     & 46.61 & 73.83 & 82.07 & 87.80 \\
\bottomrule
\end{tabular}}
\end{subfigure}
\vspace{-0.14in}
\caption{\textbf{(Left)} We quantify any-order inference by tracking how generation moves across the tree. \textbf{(Right)} FlexMDM outperforms Dream-Coder 7B on HumanEval and HumanEval+, at Pass@16.}
\label{fig:flexmdm_tree}
\end{figure}

\vspace{-0.10in}
\textbf{Experimental setup.} Rather than training FlexMDM from scratch, we initialize from Dream-Coder 7B~\citep{xie2025dream}, while attaching an auxiliary head and an AdaLN time embedding~\citep{peebles2023scalable} for insertion prediction.
We construct the fine-tuning corpus by post-processing Python data from OpenCodeInstruct~\citep{ahmad2025opencodeinstruct}, rStar-Coder~\citep{liu2025rstarcoder}, KodCode~\citep{xu2025kodcode}, and opc-stage-2-educational~\citep{huang2025opencoder}, resulting in $\approx$2.6M training sequences.
We then fine-tune the model using the FlexMDM training loss for $\approx$1150 GPU hours.

\textbf{Results.} FlexMDM by construction addresses \xblue{\texttt{Q1}} by decoupling unmasking from fixed-position commitment. This relaxation induces an \emph{any-order inference behavior} (\xblue{\texttt{Q2}}): FlexMDM generates drafts of non-contiguous regions of the generated code. We quantify this \emph{any-order inference} by parsing generated code into an AST (as in Section~\ref{sec:similarity}) and tracking how generation moves across the tree. We use three complementary metrics: Coverage Before Commitment (CBC), the average number of sibling nodes previewed before completing the first child block; Return to Unfinished Blocks (RUB), which measures back-and-forth revisits to unfinished nodes; and Open-Block Width (OBW), the maximum number of sibling blocks kept open simultaneously. FlexMDM yields substantially higher values than Dream-Coder's standard inference across all metrics (Figure~\ref{fig:flexmdm_tree}, left), supporting our claim of achieving any-order inference with insertion-based models: they generate code across non-contiguous semantic components, whereas an MDM with confidence-based decoding largely collapses to an autoregressive order. We defer precise definitions and additional results to Appendix~\ref{app:flexmdm_anyorder}.

Next, we test whether this any-order behavior translates into downstream gains. On HumanEval and HumanEval+, FlexMDM outperforms Dream-Coder, our primary baseline, at Pass@16 (Figure~\ref{fig:flexmdm_tree}, right). This suggests that any-order inference may help by enabling more diverse algorithmic solution paths. On MBPP and MBPP+, lowering the insertion temperature to $0.6$ puts FlexMDM ahead of Dream-Coder at Pass@1, with overall performance matching the baseline (Table~\ref{tab:flexmdm_full_passk}); Appendix~\ref{app:insertion_temp} ablates this knob.

\subsection{LatentMDM} \label{sec:lmdm}
In this section, we explain the training and inference recipes of LatentMDM, and present empirical results on pre-training a 125M-scale LatentMDM on TinyGSM~\citep{liu2023tinygsm}.

\textbf{Training.} We partition each sequence $\rvx$ into at most $L_s$ (variable-length) segments, $\rvx=(\rvy^1, \dots, \rvy^{L_s})$, where each segment $\rvy^i$ belongs to the segment space $\mathcal{Y}$. An encoder $E_\phi:\mathcal{Y} \rightarrow \mathbb{R}^H$ maps each segment to a fixed-dimensional latent vector; therefore, a full sequence is represented as a tensor in $\mathbb{R}^{L_s\times H}$.  
We train a masked diffusion model in this latent space by applying masking \emph{at the segment level}. Concretely, we draw a partially masked segment sequence $\rvz=(\rvz^1,\dots,\rvz^{L_s})$ with masked segment indices $\mathcal{M}(\rvz) \subseteq \{1,\dots,L_s\}$.
For each $i\in\mathcal{M}(\rvz)$, the segment is replaced by $\mask$, while unmasked segments remain unchanged, i.e., $\rvz^i=\rvy^i$ for $i\notin\mathcal{M}(\rvz)$. 
The latent input is then formed by encoding unmasked segments with $E_\phi$ and replacing each masked segment with a learnable mask embedding $\mathbf{e}_{\mask}\in\mathbb{R}^H$. 

The LatentMDM, parameterized as $f_\theta:\mathbb{R}^{L_s \times H}  \to\mathbb{R}^{L_s \times H'}$, takes this partially masked latent sequence, conditioned on the prompt, and predicts a latent vector in $\mathbb{R}^{H'}$ for every masked segment. The LatentMDM preserves and projects each segment into $\mathbb{R}^{H'}$, which serves as the conditioning space for the decoder.
Finally, an \emph{autoregressive} decoder $D_\psi$
reconstructs each latent representation in $\mathbb{R}^{H'}$ token by token.
The training objective is the average autoregressive cross-entropy loss over the masked segments:

\vspace{-0.22in}
{\small\begin{equation*}
\mathcal{L}(\theta,\phi,\psi)
=
\mathbb{E}_{\rvx,\rvz}
\left[\frac{1}{|\mathcal{M}(\rvz)|} \sum_{i \in \mathcal{M}(\rvz)}
\sum_{j=1}^{\mathrm{len}(\rvy^i)}
-\log D_\psi\!\left(\rvy^i_j \mid \rvy^i_{<j}, \mathbf{h}_{\theta,\phi}^i(\rvz)\right)
\right],
\end{equation*}}
\vspace{-0.11in}

where $\mathbf{h}_{\theta,\phi}(\rvz)\in\mathbb{R}^{L_s \times H'}$ denotes the latent conditioning produced by LatentMDM, $\mathrm{len}(\rvy^i)$ denotes the length of $\rvy^i$, and the expectation is taken over the masking procedure defined above. Importantly, we use an autoregressive decoder for local segment sampling, since next-token prediction modeling remains effective for local sequence modeling~\citep{kim2025train}.

\textbf{Inference.}
At inference time, the model is given only the prompt and maintains a partially filled segment sequence $\rvz$. Initially, all segment positions are masked. At each iteration, the revealed segments are encoded by $E_\phi$, while masked positions are represented by the learned mask embedding. The LatentMDM then predicts a contextual latent $\mathbf{h}_{\theta,\phi}^i(\rvz)$ for each masked segment position $i$.

The key design knob is which segment to reveal next: the autoregressive decoder $D_\psi$ tentatively generates a candidate segment $\hat{\rvy}^i$ token-by-token from $\mathbf{h}_{\theta,\phi}^i(\rvz)$.
This tentative decoding is performed in parallel across all masked segments.
We score each candidate by its length-normalized token log-likelihood under $D_\psi$, commit the highest-scoring segment, re-encode it with $E_\phi$, and update the corresponding masked position in $\rvz$ for the next iteration. Thus, the inference-time decoding strategy is driven by \emph{segment-level confidence}, which we view as \emph{any-order inference} over semantic space: the model reveals the latent segment whose prediction can be most confidently realized.

\vspace{-0.05in}
\begin{figure}[h]
\includegraphics[width=0.98\textwidth,trim={0 2.0cm 0 2.4cm}]{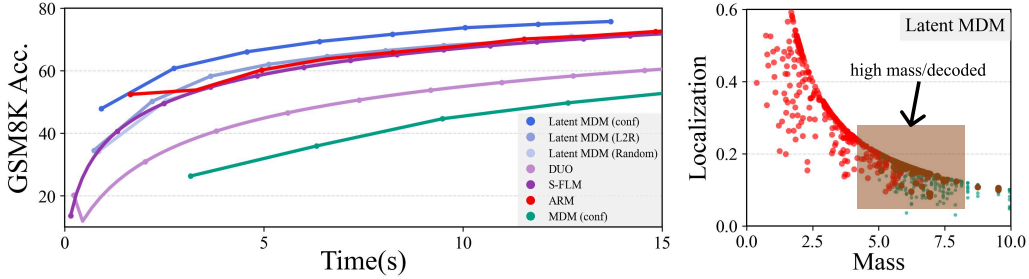}
\vspace{-0.05in}
\caption{\textbf{(Left)} LatentMDM outperforms autoregressive and other baselines at matched wall-clock time. \textbf{(Right)} It mitigates positional uncertainty, decoding tokens with high aggregate mass.}
\label{fig:LatentMDM_main_exp}
\vspace{-0.1in}
\end{figure}

\textbf{Experimental setup.}
Since LatentMDM introduces fundamentally different modeling from MDMs, we had to train it from scratch. In particular, we target a setup where (1) models pretrained from scratch can achieve reasonable performance on downstream benchmarks, e.g., GSM8K~\citep{cobbe2021training}, while (2) the compute budget remains feasible for running several controlled experiments. We adopt TinyGSM~\citep{liu2023tinygsm} as the pretraining corpus.
TinyGSM converts GSM8K-style natural-language solutions into structured Python programs, making the task substantially more learnable at small model scales.
We train LatentMDM and all baselines from scratch, using the same 125M-parameter scale with 500k optimizer steps and a global batch size of 256.

\xred{\texttt{A1.}}
To answer \xblue{\texttt{Q1}}, we revisit the \textit{positional-uncertainty} diagnosis in Section~\ref{sec:pos_uncer}. We examine the same mass-localization space defined by $m(v;\rvx_t)$ and $\mathrm{LOC}(v;\rvx_t)$. Figure~\ref{fig:LatentMDM_main_exp} (right) shows that LatentMDM commits a large fraction of high-mass but low-localization tokens (\coloredul{xsienna}{brown}). This contrasts with MDM, where tokens in the same region remain undecoded (Figure~\ref{fig:pos_uncert}). This demonstrates that moving the any-order interface from token positions to segment-level latents relaxes the positional bottleneck of fixed-canvas MDMs.

\xred{\texttt{A2.}}
To answer \xblue{\texttt{Q2}}, we scrutinize whether LatentMDM enables a useful form of any-order inference over semantic segments, in particular on GSM8K. We compare LatentMDM against representative discrete generative models: masked diffusion models (MDMs), autoregressive models, hyperspherical flow language model~\citep{deschenaux2026language} ($\mathbb{S}$-FLM), and DUO~\citep{sahoo2025diffusion}.\footnote{We omit Flow Map Language Models~\cite{lee2026flow}, as \citet{deschenaux2026language} benchmark against it under an almost identical experimental setup.}

For hyperspherical flow language model and DUO, we largely follow the implementation details in the codebase of \citet{deschenaux2026language}. Both methods support few-step sampling, which reduces latency but may come at the cost of lower accuracy. We therefore evaluate their latency–accuracy trade-offs across sampling steps in $\{8,16,\ldots,512\}$ and report the Pareto-optimal configurations: 16 steps for Spherical Flow Map Language Models and 8 steps for DUO. Moreover, we include two decoding-policy variants with the same LatentMDM: LatentMDM-random, which selects segments uniformly at random, and LatentMDM-L2R, which reveals segments in left-to-right order.

Fig.~\ref{fig:LatentMDM_main_exp} (left) reports pass@K evaluation results, together with the corresponding wall-clock decoding time.
The results show that LatentMDM consistently outperforms both LatentMDM-L2R and LatentMDM-random, by more than 6$\%$.
Since these variants differ only in their segment-selection policy, this result shows that the gain does not come merely from latent modeling, but from the ability to \emph{choose which semantic segment to reveal next}, indicating that 
LatentMDM performs genuinely any-order decoding over segments.

We next compare against modeling baselines.
Under a matched wall-clock budget, LatentMDM outperforms all other models by a gap of 5.5$\%$ at a sampling batch size of $1$. This efficiency arises because one LatentMDM inference pass takes roughly half the latency of an autoregressive model, even with KV caching, as LatentMDM first samples latent segments, which are then decoded by a lightweight decoder. The gain further suggests that compact latent-space search preserves useful exploration, in contrast to token-level confidence decoding, where confidence over-selects locally easy positions and compresses sample diversity.
We defer experimental details and additional results to Appendix~\ref{app:latent_mdm}.

Together, these results support our central claim: moving the any-order interface from token positions to latent segments \xblue{\textbf{relaxes the positional bottleneck, thus enabling semantic-order search through a distinct and practically advantageous form of any-order inference.}}

\textbf{Advances over prior work.} For FlexMDM, we advance both the theory and the empirical scope of \citet{kim2025any}. Theoretically, we show that the schedules can be changed at inference time without retraining via a time reparameterization trick, yielding further flexibility at inference time (Appendix~\ref{app:flexmdm_training}). We further derive a count-preserving \emph{insertion temperature} — an insertion-side analogue of the token temperature that gives flexible inference-time control over the structural stochasticity of insertion (Appendix~\ref{app:insertion_temp}). We also reformulate insertion-length prediction in log space, which is geometrically more natural and leads to more stable training. Empirically, we move beyond the task-specific LoRA fine-tuning and fully fine-tune Dream-Coder-7B into a general-purpose Python FlexMDM. Thus, in our work, FlexMDM serves both as a theoretically extended model and a repurposed empirical vehicle for investigating how insertion can elicit genuinely any-order generation. For LatentMDM, we note that prior work has not studied masked diffusion over semantic latent spaces; we provide a detailed discussion of related work in Appendix~\ref{app:related_works}.

\textbf{Discussion.} FlexMDM and LatentMDM relax the positional bottleneck in complementary ways rather than serving as competing designs. FlexMDM stays in the token space, making it easier to \emph{retrofit} from a pretrained MDM. LatentMDM enables segment-level order search but typically requires training from scratch. Combining insertion-based refinement with latent-space order search may realize the benefits of both forms of any-order inference; we leave this direction to future work.
\vspace{-0.12in}
\section{Conclusion}
\vspace{-0.10in}
Our two approaches, FlexMDM and LatentMDM, relax the fixed-canvas, token-level bottleneck of MDMs and thereby enable distinct forms of any-order inference. These approaches, however, also introduce trade-offs. More flexible inference procedures require additional design choices and may require extra care to ensure training and sampling stability. Moreover, our empirical study focuses primarily on Python code generation, leaving open whether similar forms of any-order inference transfer to broader domains, modalities, and scales.

Beyond masked-token diffusion, recent approaches based on continuous-state modeling or uniform-state diffusions~\citep{von2025scaling,sahoo2025diffusion,lee2026flow,roos2026categorical,potaptchik2026discrete} may also support non-causal generation in principle. However, they lack the explicit any-order interface found in mask-based models, leaving the elicitation of genuine any-order inference in these alternative diffusion frameworks as an important open question.\footnote{We note that very recently, \citet{agarwal2026posterior} proposed training an any-order model in a continuous embedding space for discrete generation, which enables any-order inference.} Another important direction is to investigate whether equipping masked diffusion models with self-correction mechanisms~\citep{wang2026remasking,kim2025fine,huang2025don,schiff2026learn} can elicit any-order inference behavior.

\textbf{Outlook.}
We, the authors, want the paper to be viewed as a \emph{foundational work} for generative modeling on discrete spaces, asking whether a model actually realizes the capabilities suggested by its modeling design. Recent progress in generative modeling has largely been driven by architectures and training procedures that scale effectively. Our work asks whether scalability must be the sole organizing principle for generative model design, or whether models can be designed so that their inference procedures better reflect the structure of the objects they generate. This perspective treats the structure of inference as a central modeling choice rather than a fixed sampling procedure, opening up a broader design space for models that more naturally support planning, refinement, and reasoning.

\section*{Acknowledgements} JK thanks Jiaxin Shi for insightful discussions that inspired this work. JK is also grateful to Kiwhan Song for sharing the idea of any-order models in continuous spaces. SK thanks Minkai Xu for insightful discussions and valuable feedback on this work. JK and YC thank Brian Lee and Michael Albergo for fruitful discussions on FlexMDM and parameterization via insertion schedules. JK acknowledges support from the Kempner Institute. SK and TL acknowledge support from the Texas Advanced Computing Center. JK and YC are grateful to the Kempner Institute for providing the compute. SC is supported in part by NSF CAREER award CCF2441635.

\section*{Contribution statement}
SK, JK, TL, and YC are co-first authors, and we include this contribution statement solely for clarity. While all authors contributed to every aspect of the project, the primary leads for each component are as follows: Preliminary idea and paper presentation: JK, Similarity analysis: SK, TL, LatentMDM: SK, FlexMDM: JK, YC.
\bibliographystyle{plainnat}
\bibliography{main}

@article{wewer2025spatial,
  title={Spatial reasoning with denoising models},
  author={Wewer, Christopher and Pogodzinski, Bart and Schiele, Bernt and Lenssen, Jan Eric},
  journal={arXiv preprint arXiv:2502.21075},
  year={2025}
}

@inproceedings{loshchilov2019decoupled,
  title     = {Decoupled Weight Decay Regularization},
  author    = {Loshchilov, Ilya and Hutter, Frank},
  booktitle = {International Conference on Learning Representations},
  year      = {2019},
  url       = {https://openreview.net/forum?id=Bkg6RiCqY7}
}

@article{gong2025diffucoder,
  title={DiffuCoder: Understanding and Improving Masked Diffusion Models for Code Generation},
  author={Gong, Shansan and Zhang, Ruixiang and Zheng, Huangjie and Gu, Jiatao and Jaitly, Navdeep and Kong, Lingpeng and Zhang, Yizhe},
  journal={arXiv preprint arXiv:2506.20639},
  year={2025}
}

@article{agarwal2026posterior,
  title={Posterior Refinement: Fast Language Generation via Any-Order Flow Maps},
  author={Agarwal, Manan and Shah, Sheel and Lee, Chanhyuk and Yoo, Jaehoon and Huang, Jerry and Hong, Seunghoon and Raghunathan, Aditi and Kim, Jinwoo and Boffi, Nicholas M},
  journal={arXiv preprint arXiv:2606.24773},
  year={2026}
}

@article{shi2024simplified,
  title={Simplified and generalized masked diffusion for discrete data},
  author={Shi, Jiaxin and Han, Kehang and Wang, Zhe and Doucet, Arnaud and Titsias, Michalis},
  journal={Advances in neural information processing systems},
  volume={37},
  pages={103131--103167},
  year={2024}
}

@misc{peng2025pathplanningmaskeddiffusion,
      title={Path Planning for Masked Diffusion Model Sampling}, 
      author={Fred Zhangzhi Peng and Zachary Bezemek and Sawan Patel and Jarrid Rector-Brooks and Sherwood Yao and Avishek Joey Bose and Alexander Tong and Pranam Chatterjee},
      year={2025},
      eprint={2502.03540},
      archivePrefix={arXiv},
      primaryClass={cs.LG},
      url={https://arxiv.org/abs/2502.03540}, 
}

@article{jain2024livecodebench,
  title={Livecodebench: Holistic and contamination free evaluation of large language models for code},
  author={Jain, Naman and Han, King and Gu, Alex and Li, Wen-Ding and Yan, Fanjia and Zhang, Tianjun and Wang, Sida and Solar-Lezama, Armando and Sen, Koushik and Stoica, Ion},
  journal={arXiv preprint arXiv:2403.07974},
  year={2024}
}

@inproceedings{devlin2019bert,
  title={{BERT}: Pre-training of deep bidirectional transformers for language understanding},
  author={Devlin, Jacob and Chang, Ming-Wei and Lee, Kenton and Toutanova, Kristina},
  booktitle={Proceedings of the 2019 conference of the North American chapter of the association for computational linguistics: human language technologies, volume 1 (long and short papers)},
  pages={4171--4186},
  year={2019}
}

@article{ou2024your,
  title={Your absorbing discrete diffusion secretly models the conditional distributions of clean data},
  author={Ou, Jingyang and Nie, Shen and Xue, Kaiwen and Zhu, Fengqi and Sun, Jiacheng and Li, Zhenguo and Li, Chongxuan},
  journal={arXiv preprint arXiv:2406.03736},
  year={2024}
}

@article{zheng2024masked,
  title={Masked diffusion models are secretly time-agnostic masked models and exploit inaccurate categorical sampling},
  author={Zheng, Kaiwen and Chen, Yongxin and Mao, Hanzi and Liu, Ming-Yu and Zhu, Jun and Zhang, Qinsheng},
  journal={arXiv preprint arXiv:2409.02908},
  year={2024}
}

@article{xie2025dream,
  title={{Dream-Coder 7B}: An open diffusion language model for code},
  author={Xie, Zhihui and Ye, Jiacheng and Zheng, Lin and Gao, Jiahui and Dong, Jingwei and Wu, Zirui and Zhao, Xueliang and Gong, Shansan and Jiang, Xin and Li, Zhenguo and others},
  journal={arXiv preprint arXiv:2509.01142},
  year={2025}
}

@article{nie2025large,
  title={Large language diffusion models},
  author={Nie, Shen and Zhu, Fengqi and You, Zebin and Zhang, Xiaolu and Ou, Jingyang and Hu, Jun and Zhou, Jun and Lin, Yankai and Wen, Ji-Rong and Li, Chongxuan},
  journal={arXiv preprint arXiv:2502.09992},
  year={2025}
}

@article{ye2024beyond,
  title={Beyond autoregression: Discrete diffusion for complex reasoning and planning},
  author={Ye, Jiacheng and Gao, Jiahui and Gong, Shansan and Zheng, Lin and Jiang, Xin and Li, Zhenguo and Kong, Lingpeng},
  journal={arXiv preprint arXiv:2410.14157},
  year={2024}
}

@article{kim2025any,
  title={Any-Order Flexible Length Masked Diffusion},
  author={Kim, Jaeyeon and Cheuk-Kit, Lee and Domingo-Enrich, Carles and Du, Yilun and Kakade, Sham and Ngotiaoco, Timothy and Chen, Sitan and Albergo, Michael},
  journal={arXiv preprint arXiv:2509.01025},
  year={2025}
}

@article{kim2025train,
  title={Train for the worst, plan for the best: Understanding token ordering in masked diffusions},
  author={Kim, Jaeyeon and Shah, Kulin and Kontonis, Vasilis and Kakade, Sham and Chen, Sitan},
  journal={arXiv preprint arXiv:2502.06768},
  year={2025}
}

@article{sahoo2024simple,
  title={Simple and effective masked diffusion language models},
  author={Sahoo, Subham and Arriola, Marianne and Schiff, Yair and Gokaslan, Aaron and Marroquin, Edgar and Chiu, Justin and Rush, Alexander and Kuleshov, Volodymyr},
  journal={Advances in Neural Information Processing Systems},
  volume={37},
  pages={130136--130184},
  year={2024}
}

@article{song2025history,
  title={History-guided video diffusion},
  author={Song, Kiwhan and Chen, Boyuan and Simchowitz, Max and Du, Yilun and Tedrake, Russ and Sitzmann, Vincent},
  journal={arXiv preprint arXiv:2502.06764},
  year={2025}
}

@article{chen2024diffusion,
  title={Diffusion forcing: Next-token prediction meets full-sequence diffusion},
  author={Chen, Boyuan and Mart{\'\i} Mons{\'o}, Diego and Du, Yilun and Simchowitz, Max and Tedrake, Russ and Sitzmann, Vincent},
  journal={Advances in Neural Information Processing Systems},
  volume={37},
  pages={24081--24125},
  year={2024}
}

@article{jazbec2025learning,
  title={Learning Unmasking Policies for Diffusion Language Models},
  author={Jazbec, Metod and Olausson, Theo X and B{\'e}thune, Louis and Ablin, Pierre and Kirchhof, Michael and Monterio, Joao and Turrisi, Victor and Ramapuram, Jason and Cuturi, Marco},
  journal={arXiv preprint arXiv:2512.09106},
  year={2025}
}

@article{ben2025accelerated,
  title={Accelerated Sampling from Masked Diffusion Models via Entropy Bounded Unmasking},
  author={Ben-Hamu, Heli and Gat, Itai and Severo, Daniel and Nolte, Niklas and Karrer, Brian},
  journal={arXiv preprint arXiv:2505.24857},
  year={2025}
}

@article{cobbe2021training,
  title={Training verifiers to solve math word problems},
  author={Cobbe, Karl and Kosaraju, Vineet and Bavarian, Mohammad and Chen, Mark and Jun, Heewoo and Kaiser, Lukasz and Plappert, Matthias and Tworek, Jerry and Hilton, Jacob and Nakano, Reiichiro and others},
  journal={arXiv preprint arXiv:2110.14168},
  year={2021}
}

@misc{chen2021evaluating,
      title={Evaluating Large Language Models Trained on Code},
      author={Mark Chen and Jerry Tworek and Heewoo Jun and Qiming Yuan and Henrique Ponde de Oliveira Pinto and Jared Kaplan and Harri Edwards and Yuri Burda and Nicholas Joseph and Greg Brockman and Alex Ray and Raul Puri and Gretchen Krueger and Michael Petrov and Heidy Khlaaf and Girish Sastry and Pamela Mishkin and Brooke Chan and Scott Gray and Nick Ryder and Mikhail Pavlov and Alethea Power and Lukasz Kaiser and Mohammad Bavarian and Clemens Winter and Philippe Tillet and Felipe Petroski Such and Dave Cummings and Matthias Plappert and Fotios Chantzis and Elizabeth Barnes and Ariel Herbert-Voss and William Hebgen Guss and Alex Nichol and Alex Paino and Nikolas Tezak and Jie Tang and Igor Babuschkin and Suchir Balaji and Shantanu Jain and William Saunders and Christopher Hesse and Andrew N. Carr and Jan Leike and Josh Achiam and Vedant Misra and Evan Morikawa and Alec Radford and Matthew Knight and Miles Brundage and Mira Murati and Katie Mayer and Peter Welinder and Bob McGrew and Dario Amodei and Sam McCandlish and Ilya Sutskever and Wojciech Zaremba},
      year={2021},
      eprint={2107.03374},
      archivePrefix={arXiv},
      primaryClass={cs.LG}
}

@article{austin2021program,
  title={Program Synthesis with Large Language Models},
  author={Austin, Jacob and Odena, Augustus and Nye, Maxwell and Bosma, Maarten and Michalewski, Henryk and Dohan, David and Jiang, Ellen and Cai, Carrie and Terry, Michael and Le, Quoc and others},
  journal={arXiv preprint arXiv:2108.07732},
  year={2021}
}

@misc{bie2025llada20scalingdiffusionlanguage,
      title={{LLaDA2.0}: Scaling Up Diffusion Language Models to {100B}}, 
      author={Tiwei Bie and Maosong Cao and Kun Chen and Lun Du and Mingliang Gong and Zhuochen Gong and Yanmei Gu and Jiaqi Hu and Zenan Huang and Zhenzhong Lan and Chengxi Li and Chongxuan Li and Jianguo Li and Zehuan Li and Huabin Liu and Ling Liu and Guoshan Lu and Xiaocheng Lu and Yuxin Ma and Jianfeng Tan and Lanning Wei and Ji-Rong Wen and Yipeng Xing and Xiaolu Zhang and Junbo Zhao and Da Zheng and Jun Zhou and Junlin Zhou and Zhanchao Zhou and Liwang Zhu and Yihong Zhuang},
      year={2025},
      eprint={2512.15745},
      archivePrefix={arXiv},
      primaryClass={cs.LG},
      url={https://arxiv.org/abs/2512.15745}, 
}

@article{hong2025improving,
  title={Improving Discrete Diffusion Unmasking Policies Beyond Explicit Reference Policies},
  author={Hong, Chunsan and An, Seonho and Kim, Min-Soo and Ye, Jong Chul},
  journal={arXiv preprint arXiv:2510.05725},
  year={2025}
}

@article{chen2025dultra,
  title={{dUltra}: Ultra-Fast Diffusion Language Models via Reinforcement Learning},
  author={Chen, Shirui and Jiao, Jiantao and Ratliff, Lillian J and Zhu, Banghua},
  journal={arXiv preprint arXiv:2512.21446},
  year={2025}
}

@article{wu2025fast,
  title={Fast-d{LLM}: Training-free acceleration of diffusion {LLM} by enabling {KV} cache and parallel decoding},
  author={Wu, Chengyue and Zhang, Hao and Xue, Shuchen and Liu, Zhijian and Diao, Shizhe and Zhu, Ligeng and Luo, Ping and Han, Song and Xie, Enze},
  journal={arXiv preprint arXiv:2505.22618},
  year={2025}
}

@article{hayakawa2025demystifying,
  title={Demystifying MaskGIT Sampler and Beyond: Adaptive Order Selection in Masked Diffusion},
  author={Hayakawa, Satoshi and Takida, Yuhta and Imaizumi, Masaaki and Wakaki, Hiromi and Mitsufuji, Yuki},
  journal={arXiv preprint arXiv:2510.04525},
  year={2025}
}

@article{li2026diffusion,
  title={Why Diffusion Language Models Struggle with Truly Parallel (Non-Autoregressive) Decoding?},
  author={Li, Pengxiang and Muhtar, Dilxat and Yin, Lu and Chen, Tianlong and Liu, Shiwei},
  journal={arXiv preprint arXiv:2602.23225},
  year={2026}
}

@article{olausson2026tale,
  title={A Tale of Two Temperatures: Simple, Efficient, and Diverse Sampling from Diffusion Language Models},
  author={Olausson, Theo X and Jazbec, Metod and Wang, Xi and Solar-Lezama, Armando and Naesseth, Christian A and Mandt, Stephan and Nalisnick, Eric},
  journal={arXiv preprint arXiv:2604.09921},
  year={2026}
}

@article{ni2026flexibility,
  title={The Flexibility Trap: Why Arbitrary Order Limits Reasoning Potential in Diffusion Language Models},
  author={Ni, Zanlin and Wang, Shenzhi and Yue, Yang and Yu, Tianyu and Zhao, Weilin and Hua, Yeguo and Chen, Tianyi and Song, Jun and Yu, Cheng and Zheng, Bo and others},
  journal={arXiv preprint arXiv:2601.15165},
  year={2026}
}

@article{lamont2026free,
  title={Free Lunch for Pass@ $ k $? Low Cost Diverse Sampling for Diffusion Language Models},
  author={Lamont, Sean and Walder, Christian and Montague, Paul and Dezfouli, Amir and Norrish, Michael},
  journal={arXiv preprint arXiv:2603.04893},
  year={2026}
}

@article{fang2026locally,
  title={Locally Confident, Globally Stuck: The Quality-Exploration Dilemma in Diffusion Language Models},
  author={Fang, Liancheng and Liu, Aiwei and Zou, Henry Peng and Chen, Yankai and Ma, Enze and Pan, Leyi and Miao, Chunyu and Huang, Wei-Chieh and Liu, Xue and Yu, Philip S},
  journal={arXiv preprint arXiv:2604.00375},
  year={2026}
}

@article{shen2026improving,
  title={Improving Diffusion Language Model Decoding through Joint Search in Generation Order and Token Space},
  author={Shen, Yangyi and Feng, Tianjian and Han, Jiaqi and Wang, Wen and Chen, Tianlang and Shen, Chunhua and Leskovec, Jure and Ermon, Stefano},
  journal={arXiv preprint arXiv:2601.20339},
  year={2026}
}

@article{song2025seed,
  title={Seed diffusion: A large-scale diffusion language model with high-speed inference},
  author={Song, Yuxuan and Zhang, Zheng and Luo, Cheng and Gao, Pengyang and Xia, Fan and Luo, Hao and Li, Zheng and Yang, Yuehang and Yu, Hongli and Qu, Xingwei and others},
  journal={arXiv preprint arXiv:2508.02193},
  year={2025}
}

@misc{gemini2025diffusion,
    year = {2025},
    url = {https://blog.google/technology/google-deepmind/gemini-diffusion/},
    title = {Gemini Diffusion},
    author = {Google DeepMind}
}

@article{labs2025mercury,
  title={Mercury: Ultra-Fast Language Models Based on Diffusion},
  author={Inception--Lab and Khanna, Samar and Kharbanda, Siddhant and Li, Shufan and Varma, Harshit and Wang, Eric and Birnbaum, Sawyer and Luo, Ziyang and Miraoui, Yanis and Palrecha, Akash and others},
  journal={arXiv preprint arXiv:2506.17298},
  year={2025}
}

@article{bie2026llada2,
  title={{LLaDA}2.1: Speeding up text diffusion via token editing},
  author={Bie, Tiwei and Cao, Maosong and Cao, Xiang and Chen, Bingsen and Chen, Fuyuan and Chen, Kun and Du, Lun and Feng, Daozhuo and Feng, Haibo and Gong, Mingliang and others},
  journal={arXiv preprint arXiv:2602.08676},
  year={2026}
}

@inproceedings{schusterbauer2026denoising,
  title={Denoising, fast and slow: Difficulty-aware adaptive sampling for image generation},
  author={Schusterbauer, Johannes and Gui, Ming and Li, Yusong and Ma, Pingchuan and Krause, Felix and Ommer, Bj{\"o}rn},
  booktitle={Proceedings of the IEEE/CVF Conference on Computer Vision and Pattern Recognition},
  pages={43260--43270},
  year={2026}
}

@article{patel2026insertion,
  title={Insertion Based Sequence Generation with Learnable Order Dynamics},
  author={Patel, Dhruvesh and Rozonoyer, Benjamin and Pandey, Gaurav and Naseem, Tahira and Astudillo, Ram{\'o}n Fernandez and McCallum, Andrew},
  journal={arXiv preprint arXiv:2602.18695},
  year={2026}
}

@article{patel2025insertion,
  title={Insertion language models: Sequence generation with arbitrary-position insertions},
  author={Patel, Dhruvesh and Sahoo, Aishwarya and Amballa, Avinash and Naseem, Tahira and Rudner, Tim GJ and McCallum, Andrew},
  journal={arXiv preprint arXiv:2505.05755},
  year={2025}
}

@article{zhang2026variational,
  title={Variational Learning for Insertion-based Generation},
  author={Zhang, Yangtian and Wang, Zhe and Gretton, Arthur and Ying, Rex and van Dijk, David and Titsias, Michalis K and Shi, Jiaxin},
  journal={arXiv preprint arXiv:2606.02133},
  year={2026}
}

@article{havasi2025edit,
  title={Edit flows: Flow matching with edit operations},
  author={Havasi, Marton and Karrer, Brian and Gat, Itai and Chen, Ricky TQ},
  journal={arXiv preprint arXiv:2506.09018},
  year={2025}
}

@article{tang2026a2d2,
  title={A2D2: Fine-Tuning Any-Length Discrete Diffusion for Adaptive Decoding},
  author={Tang, Sophia and Zhu, Yuchen and Tao, Molei and Chatterjee, Pranam},
  journal={arXiv preprint arXiv:2606.13565},
  year={2026}
}

@inproceedings{patel2025improved,
  title={Improved sampling from masked diffusion models with position contrastive guidance},
  author={Patel, Dhruvesh and Naseem, Tahira and Pandey, Gaurav and Sultan, Md Arafat and McCallum, Andrew and Astudillo, Ram{\'o}n Fernandez},
  booktitle={NeurIPS 2025 Workshop on Structured Probabilistic Inference $\{$$\backslash$\&$\}$ Generative Modeling},
  year={2025}
}

@article{schiff2026learn,
  title={Learn from your mistakes: Self-correcting masked diffusion models},
  author={Schiff, Yair and Belhasin, Omer and Uziel, Roy and Wang, Guanghan and Arriola, Marianne and Turok, Gilad and Zilberstein, Ran and Elad, Michael and Kuleshov, Volodymyr},
  journal={arXiv preprint arXiv:2602.11590},
  year={2026}
}

@article{kim2025fine,
  title={Fine-tuning masked diffusion for provable self-correction},
  author={Kim, Jaeyeon and Kim, Seunggeun and Lee, Taekyun and Pan, David Z and Kim, Hyeji and Kakade, Sham and Chen, Sitan},
  journal={arXiv preprint arXiv:2510.01384},
  year={2025}
}

@article{huang2025don,
  title={Don't Settle Too Early: Self-Reflective Remasking for Diffusion Language Models},
  author={Huang, Zemin and Wang, Yuhang and Chen, Zhiyang and Qi, Guo-Jun},
  journal={arXiv preprint arXiv:2509.23653},
  year={2025}
}

@article{wang2026remasking,
  title={Remasking discrete diffusion models with inference-time scaling},
  author={Wang, Guanghan and Schiff, Yair and Sahoo, Subham and Kuleshov, Volodymyr},
  journal={Advances in Neural Information Processing Systems},
  volume={38},
  pages={147282--147339},
  year={2026}
}

@article{deschenaux2026language,
  title={Language modeling with hyperspherical flows},
  author={Deschenaux, Justin and Gulcehre, Caglar},
  journal={arXiv preprint arXiv:2605.11125},
  year={2026}
}

@article{ye2025dream,
  title={Dream {7B}: Diffusion large language models},
  author={Ye, Jiacheng and Xie, Zhihui and Zheng, Lin and Gao, Jiahui and Wu, Zirui and Jiang, Xin and Li, Zhenguo and Kong, Lingpeng},
  journal={arXiv preprint arXiv:2508.15487},
  year={2025}
}

@article{wu2025fast2,
  title={Fast-d{LLM} v2: Efficient block-diffusion {LLM}},
  author={Wu, Chengyue and Zhang, Hao and Xue, Shuchen and Diao, Shizhe and Fu, Yonggan and Liu, Zhijian and Molchanov, Pavlo and Luo, Ping and Han, Song and Xie, Enze},
  journal={arXiv preprint arXiv:2509.26328},
  year={2025}
}

@article{trainin2026discrete,
  title={Discrete Diffusion Models Exploit Asymmetry to Solve Lookahead Planning Tasks},
  author={Trainin, Itamar and Ravfogel, Shauli and Abend, Omri and Feder, Amir},
  journal={arXiv preprint arXiv:2602.19980},
  year={2026}
}

@article{liu2023tinygsm,
  title={{TinyGSM}: achieving >80\% on {GSM8k} with small language models},
  author={Liu, Bingbin and Bubeck, Sebastien and Eldan, Ronen and Kulkarni, Janardhan and Li, Yuanzhi and Nguyen, Anh and Ward, Rachel and Zhang, Yi},
  journal={arXiv preprint arXiv:2312.09241},
  year={2023}
}

@inproceedings{gwak2025reward,
  title={Reward-weighted sampling: Enhancing non-autoregressive characteristics in masked diffusion {LLM}s},
  author={Gwak, Daehoon and Jung, Minseo and Park, Junwoo and Park, Minho and Park, ChaeHun and Hyung, Junha and Choo, Jaegul},
  booktitle={Proceedings of the 2025 Conference on Empirical Methods in Natural Language Processing},
  pages={34562--34582},
  year={2025}
}

@inproceedings{ghazvininejad2019mask,
  title={Mask-predict: Parallel decoding of conditional masked language models},
  author={Ghazvininejad, Marjan and Levy, Omer and Liu, Yinhan and Zettlemoyer, Luke},
  booktitle={Proceedings of the 2019 conference on empirical methods in natural language processing and the 9th international joint conference on natural language processing (EMNLP-IJCNLP)},
  pages={6112--6121},
  year={2019}
}

@inproceedings{wang2019bert,
  title={BERT has a mouth, and it must speak: BERT as a Markov random field language model},
  author={Wang, Alex and Cho, Kyunghyun},
  booktitle={Proceedings of the workshop on methods for optimizing and evaluating neural language generation},
  pages={30--36},
  year={2019}
}

@article{kim2026stop,
  title={Stop Training for the Worst: Progressive Unmasking Accelerates Masked Diffusion Training},
  author={Kim, Jaeyeon and Geuter, Jonathan and Alvarez-Melis, David and Kakade, Sham and Chen, Sitan},
  journal={arXiv preprint arXiv:2602.10314},
  year={2026}
}

@inproceedings{peebles2023scalable,
  title={Scalable diffusion models with transformers},
  author={Peebles, William and Xie, Saining},
  booktitle={Proceedings of the IEEE/CVF international conference on computer vision},
  pages={4195--4205},
  year={2023}
}

@article{ahmad2025opencodeinstruct,
  title={{OpenCodeInstruct}: A large-scale instruction tuning dataset for code {LLM}s},
  author={Ahmad, Wasi Uddin and Ficek, Aleksander and Samadi, Mehrzad and Huang, Jocelyn and Noroozi, Vahid and Majumdar, Somshubra and Ginsburg, Boris},
  journal={arXiv preprint arXiv:2504.04030},
  year={2025}
}

@article{liu2025rstarcoder,
  title={{rStar-Coder}: Scaling Competitive Code Reasoning with a Large-Scale Verified Dataset},
  author={Liu, Yifei and Zhang, Li Lyna and Zhu, Yi and Dong, Bingcheng and Zhou, Xudong and Shang, Ning and Yang, Fan and Yang, Mao},
  journal={arXiv preprint arXiv:2505.21297},
  year={2025}
}

@inproceedings{xu2025kodcode,
  title={{KodCode}: A diverse, challenging, and verifiable synthetic dataset for coding},
  author={Xu, Zhangchen and Liu, Yang and Yin, Yueqin and Zhou, Mingyuan and Poovendran, Radha},
  booktitle={Findings of the Association for Computational Linguistics: ACL 2025},
  pages={6980--7008},
  year={2025}
}

@inproceedings{huang2025opencoder,
  title={{OpenCoder}: The open cookbook for top-tier code large language models},
  author={Huang, Siming and Cheng, Tianhao and Liu, Jason Klein and Xu, Weidi and Hao, Jiaran and Song, Liuyihan and Xu, Yang and Yang, Jian and Liu, Jiaheng and Zhang, Chenchen and others},
  booktitle={Proceedings of the 63rd Annual Meeting of the Association for Computational Linguistics (Volume 1: Long Papers)},
  pages={33167--33193},
  year={2025}
}

@article{ye2024diffusion,
  title={Diffusion of thought: Chain-of-thought reasoning in diffusion language models},
  author={Ye, Jiacheng and Gong, Shansan and Chen, Liheng and Zheng, Lin and Gao, Jiahui and Shi, Han and Wu, Chuan and Jiang, Xin and Li, Zhenguo and Bi, Wei and others},
  journal={Advances in Neural Information Processing Systems},
  volume={37},
  pages={105345--105374},
  year={2024}
}

@article{potaptchik2026discrete,
  title={Discrete Flow Maps},
  author={Potaptchik, Peter and Yim, Jason and Saravanan, Adhi and Holderrieth, Peter and Vanden-Eijnden, Eric and Albergo, Michael S},
  journal={arXiv preprint arXiv:2604.09784},
  year={2026}
}

@article{lee2026flow,
  title={Flow Map Language Models: One-step Language Modeling via Continuous Denoising},
  author={Lee, Chanhyuk and Yoo, Jaehoon and Agarwal, Manan and Shah, Sheel and Huang, Jerry and Raghunathan, Aditi and Hong, Seunghoon and Boffi, Nicholas M and Kim, Jinwoo},
  journal={arXiv preprint arXiv:2602.16813},
  year={2026}
}

@article{roos2026categorical,
  title={Categorical flow maps},
  author={Roos, Daan and Davis, Oscar and Eijkelboom, Floor and Bronstein, Michael and Welling, Max and Ceylan, {\.I}smail {\.I}lkan and Ambrogioni, Luca and van de Meent, Jan-Willem},
  journal={arXiv preprint arXiv:2602.12233},
  year={2026}
}

@article{von2025scaling,
  title={Scaling behavior of discrete diffusion language models},
  author={von R{\"u}tte, Dimitri and Fluri, Janis and Pooladzandi, Omead and Sch{\"o}lkopf, Bernhard and Hofmann, Thomas and Orvieto, Antonio},
  journal={arXiv preprint arXiv:2512.10858},
  year={2025}
}

@article{sahoo2025diffusion,
  title={The diffusion duality},
  author={Sahoo, Subham Sekhar and Deschenaux, Justin and Gokaslan, Aaron and Wang, Guanghan and Chiu, Justin and Kuleshov, Volodymyr},
  journal={arXiv preprint arXiv:2506.10892},
  year={2025}
}

@article{zheng2025continuously,
  title={Continuously augmented discrete diffusion model for categorical generative modeling},
  author={Zheng, Huangjie and Gong, Shansan and Zhang, Ruixiang and Chen, Tianrong and Gu, Jiatao and Zhou, Mingyuan and Jaitly, Navdeep and Zhang, Yizhe},
  journal={arXiv preprint arXiv:2510.01329},
  year={2025}
}

@article{hao2024training,
  title={Training large language models to reason in a continuous latent space},
  author={Hao, Shibo and Sukhbaatar, Sainbayar and Su, DiJia and Li, Xian and Hu, Zhiting and Weston, Jason and Tian, Yuandong},
  journal={arXiv preprint arXiv:2412.06769},
  year={2024}
}

@article{kang2025ladir,
  title={{LaDiR}: Latent diffusion enhances {LLM}s for text reasoning},
  author={Kang, Haoqiang and Zhang, Yizhe and Kuang, Nikki Lijing and Majamaki, Nicklas and Jaitly, Navdeep and Ma, Yi-An and Qin, Lianhui},
  journal={arXiv preprint arXiv:2510.04573},
  year={2025}
}

@article{dai2025context,
  title={Context-level language modeling by learning predictive context embeddings},
  author={Dai, Beiya and Liu, Yuliang and Xue, Daozheng and Song, Yunchong and Guo, Qipeng and Chen, Kai and Wang, Xinbing and Zhou, Bowen and Lin, Zhouhan},
  journal={arXiv preprint arXiv:2510.20280},
  year={2025}
}

@article{li2022diffusion,
  title={Diffusion-lm improves controllable text generation},
  author={Li, Xiang and Thickstun, John and Gulrajani, Ishaan and Liang, Percy S and Hashimoto, Tatsunori B},
  journal={Advances in neural information processing systems},
  volume={35},
  pages={4328--4343},
  year={2022}
}

@inproceedings{lin2023text,
  title={Text generation with diffusion language models: A pre-training approach with continuous paragraph denoise},
  author={Lin, Zhenghao and Gong, Yeyun and Shen, Yelong and Wu, Tong and Fan, Zhihao and Lin, Chen and Duan, Nan and Chen, Weizhu},
  booktitle={International Conference on Machine Learning},
  pages={21051--21064},
  year={2023},
  organization={PMLR}
}

@article{lovelace2023latent,
  title={Latent diffusion for language generation},
  author={Lovelace, Justin and Kishore, Varsha and Wan, Chao and Shekhtman, Eliot and Weinberger, Kilian Q},
  journal={Advances in Neural Information Processing Systems},
  volume={36},
  pages={56998--57025},
  year={2023}
}

@article{zhang2023planner,
  title={Planner: Generating diversified paragraph via latent language diffusion model},
  author={Zhang, Yizhe and Gu, Jiatao and Wu, Zhuofeng and Zhai, Shuangfei and Susskind, Joshua and Jaitly, Navdeep},
  journal={Advances in Neural Information Processing Systems},
  volume={36},
  pages={80178--80190},
  year={2023}
}

@inproceedings{zhu2025segment,
  title={Segment-Level Diffusion: A Framework for Controllable Long-Form Generation with Diffusion Language Models},
  author={Zhu, Xiaochen and Karadzhov, Georgi and Whitehouse, Chenxi and Vlachos, Andreas},
  booktitle={Proceedings of the 63rd Annual Meeting of the Association for Computational Linguistics (Volume 1: Long Papers)},
  pages={4163--4183},
  year={2025}
}

@article{barrault2024large,
  title={Large concept models: Language modeling in a sentence representation space},
  author={Barrault, Lo{\"\i}c and Duquenne, Paul-Ambroise and Elbayad, Maha and Kozhevnikov, Artyom and Alastruey, Belen and Andrews, Pierre and Coria, Mariano and Couairon, Guillaume and Costa-juss{\`a}, Marta R and Dale, David and others},
  journal={arXiv preprint arXiv:2412.08821},
  year={2024}
}

@article{qu2025dynamic,
  title={Dynamic Large Concept Models: Latent Reasoning in an Adaptive Semantic Space},
  author={Qu, Xingwei and Wang, Shaowen and Huang, Zihao and Hua, Kai and Yin, Fan and Zhu, Rui-Jie and Zhou, Jundong and Min, Qiyang and Wang, Zihao and Li, Yizhi and others},
  journal={arXiv preprint arXiv:2512.24617},
  year={2025}
}

@article{zhang2025flexible,
  title={Flexible language modeling in continuous space with transformer-based autoregressive flows},
  author={Zhang, Ruixiang and Zhai, Shuangfei and Gu, Jiatao and Zhang, Yizhe and Zheng, Huangjie and Chen, Tianrong and Bautista, Miguel Angel and Susskind, Josh and Jaitly, Navdeep},
  journal={arXiv preprint arXiv:2507.00425},
  year={2025}
}

@article{qu2026vdlm,
  title={{VDLM}: Variable Diffusion {LM}s via Robust Latent-to-Text Rendering},
  author={Qu, Shuhui},
  journal={arXiv preprint arXiv:2602.15870},
  year={2026}
}

@article{bavarian2022efficient,
  title={Efficient training of language models to fill in the middle},
  author={Bavarian, Mohammad and Jun, Heewoo and Tezak, Nikolas and Schulman, John and McLeavey, Christine and Tworek, Jerry and Chen, Mark},
  journal={arXiv preprint arXiv:2207.14255},
  year={2022}
}

@article{shah2024causal,
  title={Causal language modeling can elicit search and reasoning capabilities on logic puzzles},
  author={Shah, Kulin and Dikkala, Nishanth and Wang, Xin and Panigrahy, Rina},
  journal={Advances in Neural Information Processing Systems},
  volume={37},
  pages={56674--56702},
  year={2024}
}

@article{lu2025next,
  title={Next edit prediction: Learning to predict code edits from context and interaction history},
  author={Lu, Ruofan and Huo, Yintong and Zhang, Meng and Li, Yichen and Lyu, Michael R},
  journal={arXiv preprint arXiv:2508.10074},
  year={2025}
}

@article{yang2025qwen3,
  title={Qwen3 technical report},
  author={Yang, An and Li, Anfeng and Yang, Baosong and Zhang, Beichen and Hui, Binyuan and Zheng, Bo and Yu, Bowen and Gao, Chang and Huang, Chengen and Lv, Chenxu and others},
  journal={arXiv preprint arXiv:2505.09388},
  year={2025}
}

@article{meshchaninov2025cosmos,
  title={Cosmos: Compressed and Smooth Latent Space for Text Diffusion Modeling},
  author={Meshchaninov, Viacheslav and Chimbulatov, Egor and Shabalin, Alexander and Abramov, Aleksandr and Vetrov, Dmitry},
  journal={arXiv preprint arXiv:2506.21170},
  year={2025}
}

@article{chen2026langflow,
  title={{LangFlow}: Continuous Diffusion Rivals Discrete in Language Modeling},
  author={Chen, Yuxin and Liang, Chumeng and Sui, Hangke and Guo, Ruihan and Cheng, Chaoran and You, Jiaxuan and Liu, Ge},
  journal={arXiv preprint arXiv:2604.11748},
  year={2026}
}

@article{shen2026codar,
  title={{CoDAR}: Continuous Diffusion Language Models are More Powerful Than You Think},
  author={Shen, Junzhe and Zhao, Jieru and He, Ziwei and Lin, Zhouhan},
  journal={arXiv preprint arXiv:2603.02547},
  year={2026}
}

@article{pynadath2025candi,
  title={{CANDI}: Hybrid discrete-continuous diffusion models},
  author={Pynadath, Patrick and Shi, Jiaxin and Zhang, Ruqi},
  journal={arXiv preprint arXiv:2510.22510},
  year={2025}
}

@article{zhou2025coevolutionary,
  title={Coevolutionary continuous discrete diffusion: Make your diffusion language model a latent reasoner},
  author={Zhou, Cai and Yang, Chenxiao and Hu, Yi and Wang, Chenyu and Zhang, Chubin and Zhang, Muhan and Mackey, Lester and Jaakkola, Tommi and Bates, Stephen and Zhang, Dinghuai},
  journal={arXiv preprint arXiv:2510.03206},
  year={2025}
}

@article{liu2025tidar,
  title={Tidar: Think in diffusion, talk in autoregression},
  author={Liu, Jingyu and Dong, Xin and Ye, Zhifan and Mehta, Rishabh and Fu, Yonggan and Singh, Vartika and Kautz, Jan and Zhang, Ce and Molchanov, Pavlo},
  journal={arXiv preprint arXiv:2511.08923},
  year={2025}
}

@article{li2025refusion,
  title={{ReFusion}: A Diffusion Large Language Model with Parallel Autoregressive Decoding},
  author={Li, Jia-Nan and Guan, Jian and Wu, Wei and Li, Chongxuan},
  journal={arXiv preprint arXiv:2512.13586},
  year={2025}
}

@article{hersche2026locally,
  title={Locally Coherent Parallel Decoding in Diffusion Language Models},
  author={Hersche, Michael and Menet, Nicolas and Tanios, Ronan and Rahimi, Abbas},
  journal={arXiv preprint arXiv:2603.20216},
  year={2026}
}

@article{li2024autoregressive,
  title={Autoregressive image generation without vector quantization},
  author={Li, Tianhong and Tian, Yonglong and Li, He and Deng, Mingyang and He, Kaiming},
  journal={Advances in Neural Information Processing Systems},
  volume={37},
  pages={56424--56445},
  year={2024}
}

@article{yoo2026self,
  title={Self-conditioned Flow Map Language Models via Fixed-point Flows},
  author={Yoo, Jaehoon and Kim, Wonjung and Eijkelboom, Floor and Lee, Chanhyuk and Boffi, Nicholas M and Hong, Seunghoon and Kim, Jinwoo},
  journal={arXiv preprint arXiv:2607.00714},
  year={2026}
}

@article{hu2026elf,
  title={Elf: Embedded language flows},
  author={Hu, Keya and Qiu, Linlu and Lu, Yiyang and Zhao, Hanhong and Li, Tianhong and Kim, Yoon and Andreas, Jacob and He, Kaiming},
  journal={arXiv preprint arXiv:2605.10938},
  year={2026}
}

@article{batzolis2026cobit,
  title={CoBit: Language Modeling with Bitstream Diffusion},
  author={Batzolis, Georgios and Girolami, Mark and Ambrogioni, Luca},
  journal={arXiv preprint arXiv:2605.07013},
  year={2026}
}
\newpage
\appendix
\tableofcontents
\newpage
\section{Related Works} \label{app:related_works}
\textbf{Causal collapse in MDMs.}
Recent work has identified two related but distinct limitations of confidence-based decoding in masked diffusion models: reduced sample diversity and the collapse of flexible generation orders toward AR-like trajectories~\citep{ni2026flexibility,olausson2026tale,shen2026improving,lamont2026free,fang2026locally,gong2025diffucoder,li2026diffusion}. As we make it clear in Section~\ref{sec:similarity}, the former concerns the entropy and coverage of the resulting samples, whereas the latter concerns the structure of the generation process itself.

A first line of work studies why confidence-based any-order decoding produces less diverse outputs than fixed left-to-right sampling with a nonzero token-sampling temperature. \citet{ni2026flexibility} characterize this phenomenon as the \emph{flexibility trap}. Since confidence-based decoding can postpone uncertain tokens and instead resolve low-entropy positions, it may bypass the branching points at which autoregressive trajectories would emerge. Motivated by a similar observation, \citet{fang2026locally} describe a \emph{quality--exploration dilemma}: greedily committing high-confidence tokens can improve the quality of an individual trajectory while suppressing sequence-level entropy and harming multi-sample exploration at Pass@K.

Several works develop inference-time methods to mitigate this diversity degradation. \citet{olausson2026tale} distinguish the temperature used to sample token values from the temperature used to select or remask token positions. \citet{lamont2026free} instead coordinate multiple samples within a batch, discouraging each new trajectory from reproducing feature-space directions already explored by previous samples. Most closely related, \citet{patel2025improved} reweights a token's probability at a given position relative to a local or global cross-position reference. Their notion of \emph{cross-position reference} is closely related to our aggregate probability mass. Whereas they leverage this signal for an inference-time algorithm, we use it as a diagnostic of positional uncertainty, making it complementary.

A second line of work more directly examines the generation trajectory itself. \citet{gong2025diffucoder} introduce local and global measures of \emph{AR-ness}, quantifying how closely the order in which an MDM finalizes tokens follows a left-to-right trajectory. They show that this behavior depends on training and inference choices and that increasing sampling temperature can diversify not only token values but also generation orders. \citet{gwak2025reward} similarly observe that independently selecting positions according to token-level confidence often yields sequential, AR-like trajectories, and propose reward-weighted sampling to favor groups of tokens that better preserve non-autoregressive generation. \citet{li2026diffusion} attribute the tendency toward causal generation to biases inherited from the training data and propose a parallel-inference design intended to reduce this bias. Other methods treat generation order as an inference-time search problem. Rather than accepting the single trajectory induced by greedy confidence ranking, \citet{shen2026improving} jointly search over both which positions to decode and which token values to assign.

Our work focuses on this remaining mechanistic question: we ask why standard MDM inference, despite its \emph{any-order interface}, fails to go meaningfully beyond AR-like reasoning in the first place. We quantify how closely confidence-based inference structurally approximates causal decoding on Python programs (Section~\ref{sec:similarity})  and attribute this collapse to a more fundamental bottleneck of token-level, fixed-canvas inference (Section~\ref{sec:pos_uncer}). This mechanism also reveals broader limitations for tasks requiring globally structured generation.

\textbf{Continuous latent reasoning in autoregressive LLMs.}
Reasoning in continuous representation space has recently gained attention as a way to reduce the cost of explicit chain-of-thought generation while preserving intermediate computation. Since such methods avoid verbalizing every reasoning step, they can improve efficiency, though at the cost of reduced interpretability.
Coconut~\citep{hao2024training} enables LLMs to perform intermediate reasoning in a continuous hidden-state space by feeding the last hidden state back as the next input embedding, rather than decoding each reasoning step into language tokens.
ContextLM~\citep{dai2025context} augments autoregressive LMs with latent context prediction, where fixed-size context embeddings guide subsequent token generation.
DLCM~\citep{qu2025dynamic} learns to parse sequences into variable-length concept segments, runs a causal Transformer over the compressed concept sequence, and decodes the resulting representations back to tokens, thereby reallocating computation around semantic boundaries.
These works demonstrate the usefulness of continuous representations for efficient reasoning, but remain fundamentally autoregressive at the high-level inference interface.

\textbf{Continuous generative models for language.}
Another line of work develops continuous generative models for text, often avoiding discrete-token realization during intermediate generation.
Early approaches~\citep{li2022diffusion, lin2023text} apply diffusion models to token-level continuous embeddings, mainly for controllable generation or sequence-to-sequence generation.
TarflowLM~\citep{zhang2025flexible} models token-level Gaussian embeddings using Transformer-based autoregressive normalizing flows.
A common issue in direct embedding-space generation is the rounding problem: predicted continuous embeddings must eventually be mapped back to discrete tokens, and this projection can introduce rounding errors.

More recent approaches~\citep{chen2026langflow, lee2026flow} indirectly mitigate this issue by matching token posteriors rather than directly regressing word embeddings.
This is especially explicit in FMLM~\citep{lee2026flow}, which operates in one-hot token space and decodes by taking the argmax over predicted token probabilities.

This flow-map approach to few-step generation of discrete data has advanced
rapidly. Categorical Flow Maps~\citep{roos2026categorical} learn maps toward the probability simplex using distillation and endpoint-consistency objectives, while Discrete Flow Maps~\citep{potaptchik2026discrete} modify flow-map training to better reflect the discrete structure of language and support full-sequence generation in a single model evaluation. Building on FMLM, \citet{yoo2026self} interpret self-conditioning in flow-based language models as a fixed-point iteration and use this perspective to improve few-step generation. Other works replace one-hot token representations with more compact
continuous parameterizations: the hyperspherical flow language model
($\mathbb{S}$-FLM)~\citep{deschenaux2026language} runs latent flows on the unit hypersphere and narrows the gap to MDMs on mathematics and code (we adopt it as a representative continuous-flow baseline in Section~\ref{sec:lmdm}); ELF~\citep{hu2026elf} performs flow matching in word-embedding space and converts the resulting continuous states to token predictions through a learned output map; and CoBit~\citep{batzolis2026cobit}, applies diffusion to analog bitstream representations of tokens, reducing the dimensional dependence of each token prediction from linear to logarithmic in the vocabulary size.
Closest to our focus,
\citet{agarwal2026posterior} equip flow-map language models with masking-style noise
schedules, yielding an any-order continuous-space sampler with an inference-time
self-correction mechanism (posterior refinement); their goal, however, is to improve the speed--quality trade-off of parallel generation rather than to analyze the structure of the generation orders that the any-order interface induces.

A complementary way to avoid token-level rounding is to separate high-level continuous planning from discrete token realization.
LD4PG~\citep{lovelace2023latent}, PLANNER~\citep{zhang2023planner}, Segment-Level Diffusion~\citep{zhu2025segment}, VDLM~\citep{qu2026vdlm}, and CoDAR~\citep{shen2026codar} use continuous diffusion models over higher-level semantic latents (i.e., sentence-, paragraph-, or whole-sequence-level representations), and rely on an autoregressive token-level decoder or renderer to iteratively decode each denoised latent back into discrete text in the final stage of generation.
COSMOS~\citep{meshchaninov2025cosmos} similarly constructs a compressed continuous latent space for text diffusion, but decodes tokens in parallel.
LCM~\citep{barrault2024large} and LaDiR~\citep{kang2025ladir} both use diffusion over continuous high-level semantic representations beyond tokens while performing autoregressive generation at the semantic level: LCM denoises sentence-level concept embeddings for semantic continuation, whereas LaDiR denoises structured latent chain-of-thought blocks for reasoning.

Related hybrid continuous--discrete diffusion methods, including CADD~\citep{zheng2025continuously}, CANDI~\citep{pynadath2025candi}, and CCDD~\citep{zhou2025coevolutionary}, combine continuous and discrete diffusion to reduce rounding mismatch, though the continuous variables differ across methods, including word embeddings, one-hot vectors, and LLM-contextualized latents.

Our LatentMDM shares the broad motivation of using continuous latent variables to represent higher-level semantic units. However, our focus is different: we study whether moving the masked-diffusion interface from token positions to segment-level latents can elicit genuinely any-order inference. Most prior continuous or latent-space language models primarily target likelihood modeling, controllability, or fast generation, and typically do not analyze whether their generation traces differ structurally from ARM or token-level MDM decoding.

\textbf{Hybrid autoregressive and masked-diffusion.}
Our LatentMDM also relates to hybrid architectures that combine non-autoregressive (NAR) prediction with autoregressive token generation.
TiDAR~\citep{liu2025tidar} uses a token-level MDM as a parallel drafter and an ARM as a verifier or sampler within a unified module.
ReFusion~\citep{li2025refusion} fine-tunes an ARM for block-level any-order generation while preserving token-level autoregressivity, enabling efficient inference with KV caching.
CoDiLA~\citep{hersche2026locally} uses a token-level MDM as a parallel drafter and a smaller local ARM as a conditional sampler given soft draft conditions.
These approaches are promising for improving sampling efficiency, but their any-order structure is generally defined over fixed token blocks while preserving substantial token-level autoregressive-ness. In contrast, our LatentMDM moves the reveal decision to variable-length semantic segments and asks whether segment-level confidence can induce generation orders that escape the left-to-right bias observed in token-level MDMs.

\textbf{Any-order models in continuous latent spaces.}
MAR~\citep{li2024autoregressive} and Diffusion Forcing~\citep{song2025history,chen2024diffusion} propose any-order interfaces for denoising-based diffusion models in continuous latent spaces. Their use of this flexibility, however, is mainly aimed at improving generation efficiency~\citep{li2024autoregressive} or enabling conditional and long-horizon video generation~\citep{song2025history,chen2024diffusion}, rather than searching over semantic generation orders.

The works with the greatest conceptual overlap with ours are those of~\citep{wewer2025spatial,schusterbauer2026denoising}. They train diffusion-forcing-style, any-order continuous diffusion models in which each patch or frame can be assigned a separate time step, enabling flexible generation orders. Although our work shares this high-level conceptual connection, it differs substantially in formulation: these works focus on continuous-space diffusion modeling, whereas we adopt a masked-modeling formulation.

\textbf{Insertion-based discrete generative models.} Along with FlexMDM~\citep{kim2025any}, a growing body of work has sought to develop discrete generative models that support insertion~\citep{havasi2025edit,tang2026a2d2,patel2025insertion,patel2026insertion,zhang2026variational}. We leave comparisons with these models, as well as investigating how they could support any-order inference, for future work.
\section{Details of Section~\ref{sec:similarity}} \label{app:sim}

This appendix describes the code-similarity metrics used in Section~\ref{sec:similarity}. Each metric converts a generated Python code into a rooted ordered tree and then compares two such trees by tree edit distance. A tree node represents a syntactic component of the program, and an edge represents immediate containment: a parent node is a larger construct, and its children are the constructs that appear inside it. Children are ordered by their order in the source code.

For two rooted ordered labeled trees $\mathcal{T}$ and $\mathcal{T}'$, let $\mathrm{TED}_{\rho}(\mathcal{T},\mathcal{T}')$ denote the tree edit distance between them, computed with unit insertion cost, unit deletion cost, and rename cost $\rho$. We normalize the distance into a similarity score:
\[\mathrm{Sim}_{\rho}(\mathcal{T},\mathcal{T}') = 1 - \frac{\mathrm{TED}_{\rho}(\mathcal{T},\mathcal{T}')
}{\max\{|\mathcal{T}|,|\mathcal{T}'|\}},\] where $|\mathcal{T}|$ is the number of nodes in $\mathcal{T}$. Scores are clipped to $[0,1]$ when necessary. Larger values indicate a more similar tree structure. The three metrics below differ only in how they construct the tree before applying this normalized tree-edit similarity.

\paragraph{Abstract Syntax Tree Distance (ASTD).} 
Our main metric, ASTD, uses Python's built-in parser to convert each program into an abstract syntax tree. In this tree, nodes correspond to semantic syntactic units used by the Python interpreter, such as \texttt{Module}, \texttt{FunctionDef}, \texttt{For}, \texttt{If}, \texttt{Assign}, \texttt{Return}, \texttt{Name}, \texttt{Constant}, and \texttt{Add}. The parent-child relation is given by Python's AST fields: for example, a \texttt{FunctionDef} contains its arguments and body statements, an \texttt{If} contains its condition and branches, and an \texttt{Assign} contains its target and value.

For example, for the code
\[\begin{array}{l} \texttt{def f(x):}\\
\quad \texttt{y = x + 1}\\
\quad \texttt{return y}
\end{array}
\]

the AST-based representation contains nodes such as \texttt{Module}, \texttt{FunctionDef}, \texttt{arguments}, \texttt{arg}, \texttt{Assign}, \texttt{Name}, \texttt{BinOp}, \texttt{Add}, \texttt{Constant}, and \texttt{Return}. The \texttt{FunctionDef} node contains the argument list and the two body statements; the \texttt{Assign} node contains the target \texttt{Name} and the value \texttt{BinOp}; and the \texttt{BinOp} node contains the left operand, operator, and right operand.

Before comparing ASTs, we anonymize identifiers so that arbitrary names do not dominate the score. Variable and argument names are replaced by canonical placeholders, function names are canonicalized, and top-level docstrings are ignored. The resulting tree still preserves the program's control flow, statement structure, operators, literals, and expression nesting.

Let $\mathcal{A}(c)$ be this anonymized Python AST for code $c$. We define
\[\mathrm{ASTD}(c,c') = \mathrm{Sim}_{1}\bigl(\mathcal{A}(c), \mathcal{A}(c')\bigr).\]
The rename cost is $\rho=1$, so changing a node label, such as replacing \texttt{Add} with \texttt{Sub} or \texttt{Constant(1)} with \texttt{Constant(2)}, contributes edit cost. Therefore, ASTD measures both the global program skeleton and local syntactic choices inside that skeleton.

\paragraph{Tree Similarity of Edit Distance (TSED).}
TSED uses the same tree-edit-distance normalization, but builds the tree with \texttt{tree-sitter} rather than Python's AST parser. Python's AST is an interpreter-level abstraction, whereas tree-sitter follows the surface grammar of the written program. Its nodes correspond to grammar constructs such as \texttt{function\_definition}, \texttt{parameters}, \texttt{block}, \texttt{assignment}, \texttt{return\_statement}, and \texttt{call}. Edges again mean immediate containment in the parsed program, and children are ordered by source order.

For the same example above, the tree-sitter representation contains grammar nodes such as \texttt{module}, \texttt{function\_definition}, \texttt{identifier}, \texttt{parameters}, \texttt{block}, \texttt{expression\_statement}, \texttt{assignment}, \texttt{binary\_operator}, and \texttt{return\_statement}. Compared with the Python AST, this tree follows the written grammar more directly: for instance, the function body appears under a \texttt{block} node, and the assignment may appear under an \texttt{expression\_statement} node.

Let $\mathcal{G}(c)$ be the tree-sitter parse tree of code $c$. We define
\[\mathrm{TSED}(c,c') = \mathrm{Sim}_{0}\bigl(\mathcal{G}(c), \mathcal{G}(c')\bigr).\]
Here the rename cost is $\rho=0$. Thus, TSED does not charge for changing a node label; it mainly charges for inserting or deleting nodes to make the two parse trees align. In other words, ASTD asks whether the two ASTs have matching node types and values, while TSED asks a weaker question: whether the two programs have similarly shaped parse trees. This makes TSED a more permissive structural metric than ASTD.

\paragraph{Coarse ASTD (Coarse).}
Coarse ASTD is a lower-resolution variant of ASTD. It starts from the same anonymized Python AST, but keeps only statement-level structure. Nodes such as \texttt{FunctionDef}, \texttt{For}, \texttt{While}, \texttt{If}, \texttt{Assign}, and \texttt{Return} are retained, while fine-grained expression subtrees are collapsed into the label of the nearest statement node.

For the same example, Coarse ASTD keeps a much smaller tree: a root \texttt{Module} node, a \texttt{FunctionDef} node, and statement-level children such as \texttt{Assign} and \texttt{Return}. The internal expression \texttt{x + 1} is not expanded into separate \texttt{Name}, \texttt{Add}, and \texttt{Constant} nodes. Instead, it is treated as part of the assignment's statement-level label. Thus, Coarse ASTD ignores many expression-level differences and focuses on the high-level statement layout.

Let $\mathcal{C}(c)$ be this coarsened statement-level tree. We define
\[\mathrm{Coarse}(c,c') = \mathrm{Sim}_{1}\bigl(\mathcal{C}(c), \mathcal{C}(c')\bigr).\]
The rename cost is again $\rho=1$, so mismatched statement labels contribute to the edit cost. However, because expression internals are not expanded into separate nodes, Coarse ASTD focuses more directly on high-level control flow and statement layout. It tests whether two decoding strategies produce programs with similar global skeletons even after ignoring much of the expression-level detail.

\paragraph{Sampling protocol.}
For each model and prompt, all decoding policies use temperature 0.2 and nucleus sampling top-$p=0.95$. We generate one sample for each confidence-based policy, since confidence-based decoding at this temperature produces little diversity across repeated samples. We generate 32 samples for strict L2R and random decoding. The 32 L2R samples form the reference set used in best-match aggregation, while the 32 random samples provide a calibration baseline and help obtain enough valid or correct programs despite the lower success rates of random and L2R decoding.

\paragraph{Best-match aggregation over samples.}
The metrics above define the similarity between a pair of code snippets. In our experiments, however, each prompt has multiple generated samples from each decoding strategy. We therefore aggregate pairwise similarities using a best-match rule. For a prompt $q$, let $\mathcal{D}^{\pi}_q$ be the samples generated by the decoding strategy being evaluated, and let $\mathcal{D}^{\mathrm{L2R}}_q$ be the reference L2R samples. For a chosen similarity metric $s(\cdot,\cdot)$, we compute
\[\mathrm{BM}_q(s) = \frac{1}{|\mathcal{D}^{\pi}_q|}\sum_{y \in \mathcal{D}^{\pi}_q} \max_{x \in \mathcal{D}^{\mathrm{L2R}}_q} s(y,x).
\]
That is, each sample from the evaluated decoding strategy is matched to its most similar L2R sample, and we average these best-match scores. We then report the mean of $\mathrm{BM}_q(s)$ across prompts.

We apply the same aggregation separately under the valid--valid and success--success filters. For valid--valid, both $y$ and its candidate matches $x$ must be syntactically valid. For success--success, both must pass the unit tests. This aggregation asks whether each generated program has a close L2R counterpart.

\subsection{Additional similarity results}
Tables~\ref{tab:humaneval-similarity-bestmatch-all} and \ref{tab:mbpp-similarity-bestmatch-all} report the L2R comparison results for DiffuCoder-base, DiffuCoder-instruct, DiffuCoder-cpGRPO, Dream-Base, Dream-instruct, LLaDA-instruct, LLaDA-cpGRPO, and \texttt{dUltra-coding-b128} on HumanEval and MBPP. Table~\ref{tab:dclcb-similarity-bestmatch-all} reports the comparison results for Dream-Coder-instruct on LiveCodeBench. We report TSED, ASTD, and Coarse ASTD under two filtering regimes: valid--valid pairs, where both codes are syntactically valid, and correct--correct pairs, where both codes pass the unit tests.

Across all models and both datasets, Random vs L2R consistently yields much lower similarity than Entropy, Top-$K$ Margin, and Top-$K$ vs L2R. This trend holds not only for syntactically valid codes, but also for correct codes. In other words, random decoding does not merely produce structurally different incorrect code. Even among successful generations, structures are meaningfully less similar to L2R outputs than those produced by confidence-based decoding. This supports our claim that confidence-based decoding remains structurally close to L2R, whereas random reveal orders provide a substantially more non-left-to-right reference.

\begin{table*}[h]
  \centering
  \small
  \setlength{\tabcolsep}{5pt}
  \begin{tabular}{ll ccc ccc}
  \toprule
   & & \multicolumn{3}{c}{Valid pairs} & \multicolumn{3}{c}{Correct pairs} \\
  \cmidrule(lr){3-5} \cmidrule(lr){6-8}
  Model & Comparison & TSED & ASTD & Coarse & TSED & ASTD & Coarse \\
  \midrule
  \multirow{4}{*}{DiffuCoder-base}
   & Entropy vs L2R      & 0.9994 & 0.9995 & 0.9994 & 1.0000 & 1.0000 & 1.0000 \\
   & Top-$K$ Margin vs L2R & 0.8890 & 0.8427 & 0.8527 & 0.9070 & 0.8627 & 0.8670 \\
   & Top-$K$ vs L2R        & 0.8816 & 0.8320 & 0.8466 & 0.9070 & 0.8652 & 0.8708 \\
   \rowcolor{xxpurple!20}
   \cellcolor{white}
   & Random vs L2R       & 0.6857 & 0.5583 & 0.5963 & 0.8108 & 0.7077 & 0.7087 \\
  \midrule
  \multirow{4}{*}{DiffuCoder-instruct}
   & Entropy vs L2R      & 0.9637 & 0.9458 & 0.9424 & 0.9741 & 0.9628 & 0.9647 \\
   & Top-$K$ Margin vs L2R & 0.9128 & 0.8739 & 0.8801 & 0.9381 & 0.9126 & 0.9156 \\
   & Top-$K$ vs L2R        & 0.9128 & 0.8739 & 0.8801 & 0.9381 & 0.9126 & 0.9156 \\
   \rowcolor{xxpurple!20}
   \cellcolor{white}
   & Random vs L2R       & 0.7070 & 0.5981 & 0.6494 & 0.7897 & 0.6902 & 0.7090 \\
  \midrule
  \multirow{4}{*}{DiffuCoder-cpGRPO}
   & Entropy vs L2R      & 0.9795 & 0.9706 & 0.9791 & 0.9883 & 0.9794 & 0.9801 \\
   & Top-$K$ Margin vs L2R & 0.9795 & 0.9706 & 0.9791 & 0.9883 & 0.9794 & 0.9801 \\
   & Top-$K$ vs L2R        & 0.9795 & 0.9706 & 0.9791 & 0.9883 & 0.9794 & 0.9801 \\
   \rowcolor{xxpurple!20}
   \cellcolor{white}
   & Random vs L2R       & 0.8049 & 0.7245 & 0.7813 & 0.8592 & 0.7909 & 0.8173 \\
  \midrule
  \multirow{4}{*}{Dream-base}
   & Entropy vs L2R      & 0.8329 & 0.7717 & 0.7831 & 0.9047 & 0.8649 & 0.8691 \\
   & Top-$K$ Margin vs L2R & 0.8329 & 0.7715 & 0.7823 & 0.9035 & 0.8632 & 0.8673 \\
   & Top-$K$ vs L2R        & 0.8329 & 0.7717 & 0.7831 & 0.9047 & 0.8649 & 0.8691 \\
   \rowcolor{xxpurple!20}
   \cellcolor{white}
   & Random vs L2R       & 0.5973 & 0.4578 & 0.5277 & 0.7325 & 0.6328 & 0.6482 \\
  \midrule
  \multirow{4}{*}{Dream-instruct}
   & Entropy vs L2R      & 0.9074 & 0.8706 & 0.8769 & 0.9714 & 0.9585 & 0.9554 \\
   & Top-$K$ Margin vs L2R & 0.9074 & 0.8706 & 0.8769 & 0.9714 & 0.9585 & 0.9554 \\
   & Top-$K$ vs L2R        & 0.8913 & 0.8481 & 0.8576 & 0.9556 & 0.9366 & 0.9235 \\
   \rowcolor{xxpurple!20}
   \cellcolor{white}
  & Random vs L2R       & 0.6823 & 0.5626 & 0.6227 & 0.7881 & 0.7001 & 0.7288 \\
  \midrule
  \multirow{4}{*}{LLaDA-instruct}
  & Entropy vs L2R        & 0.9758 & 0.9470 & 0.9508 & 0.9718 & 0.9594 & 0.9555 \\
  & Top-$K$ Margin vs L2R & 0.9017 & 0.8494 & 0.8709 & 0.9151 & 0.8737 & 0.8959 \\
  & Top-$K$ vs L2R        & 0.9246 & 0.8843 & 0.8984 & 0.9507 & 0.9323 & 0.9374 \\
  \rowcolor{xxpurple!20}
  \cellcolor{white}
  & Random vs L2R         & 0.7672 & 0.6611 & 0.6947 & 0.8595 & 0.7958 & 0.8097 \\
  \midrule
  \multirow{4}{*}{LLaDA-cpGRPO}
  & Entropy vs L2R        & 0.8927 & 0.8434 & 0.8437 & 0.9145 & 0.8739 & 0.8757 \\
  & Top-$K$ Margin vs L2R & 0.8780 & 0.8226 & 0.8277 & 0.9192 & 0.8802 & 0.8798 \\
  & Top-$K$ vs L2R        & 0.8940 & 0.8416 & 0.8481 & 0.9287 & 0.8917 & 0.8878 \\
  \rowcolor{xxpurple!20}
  \cellcolor{white}
  & Random vs L2R         & 0.7306 & 0.6377 & 0.6451 & 0.8107 & 0.7355 & 0.7086 \\
  \midrule
  \multirow{5}{*}{dUltra-coding-b128}
   & Planner vs L2R      & 0.9632 & 0.8962 & 0.8993 & 0.9761 & 0.9689 & 0.9648 \\
   & Entropy vs L2R      & 1.0000 & 0.9295 & 0.9295 & 1.0000 & 1.0000 & 1.0000 \\
   & Top-$K$ Margin vs L2R & 0.9220 & 0.8386 & 0.8431 & 0.9358 & 0.9095 & 0.8918 \\
   & Top-$K$ vs L2R        & 0.9261 & 0.8416 & 0.8460 & 0.9499 & 0.9272 & 0.9195 \\
   \rowcolor{xxpurple!20}
   \cellcolor{white}
   & Random vs L2R       & 0.6525 & 0.3085 & 0.3505 & 0.8966 & 0.8605 & 0.8822 \\
  \bottomrule
  \end{tabular}
  \caption{HumanEval code-similarity to strict L2R baselines using best-match aggregation. Dream, DiffuCoder, LLaDA-instruct, and LLaDA-cpGRPO results are compared against L2R under the Pass@32 setup, while \texttt{dUltra-coding-b128} is evaluated under its Pass@64 setup. Valid pairs require both samples to parse; correct pairs require both samples to pass unit tests.}
  \label{tab:humaneval-similarity-bestmatch-all}
\end{table*}

\begin{table*}[h]
  \centering
  \small
  \setlength{\tabcolsep}{5pt}
  \begin{tabular}{ll ccc ccc}
  \toprule
   & & \multicolumn{3}{c}{Valid pairs} & \multicolumn{3}{c}{Correct pairs} \\
  \cmidrule(lr){3-5} \cmidrule(lr){6-8}
  Model & Comparison & TSED & ASTD & Coarse & TSED & ASTD & Coarse \\
  \midrule
  \multirow{4}{*}{DiffuCoder-base}
   & Entropy vs L2R      & 0.9960 & 0.9931 & 0.9915 & 0.9992 & 0.9987 & 0.9982 \\
   & Top-$K$ Margin vs L2R & 0.8717 & 0.8216 & 0.8781 & 0.9099 & 0.8668 & 0.9036 \\
   & Top-$K$ vs L2R        & 0.8697 & 0.8190 & 0.8764 & 0.9116 & 0.8670 & 0.9019 \\
   \rowcolor{xxpurple!20}
   \cellcolor{white}
   & Random vs L2R       & 0.7107 & 0.5842 & 0.6916 & 0.7822 & 0.6856 & 0.7597 \\
  \midrule
  \multirow{4}{*}{DiffuCoder-instruct}
   & Entropy vs L2R      & 0.8552 & 0.8055 & 0.8306 & 0.8932 & 0.8539 & 0.8638 \\
   & Top-$K$ Margin vs L2R & 0.8552 & 0.8055 & 0.8306 & 0.8932 & 0.8539 & 0.8638 \\
   & Top-$K$ vs L2R        & 0.8552 & 0.8055 & 0.8306 & 0.8932 & 0.8539 & 0.8638 \\
   \rowcolor{xxpurple!20}
   \cellcolor{white}
   & Random vs L2R       & 0.7445 & 0.6494 & 0.7363 & 0.8128 & 0.7421 & 0.7903 \\
  \midrule
  \multirow{4}{*}{DiffuCoder-cpGRPO}
   & Entropy vs L2R      & 0.9851 & 0.9808 & 0.9877 & 0.9916 & 0.9872 & 0.9952 \\
   & Top-$K$ Margin vs L2R & 0.9851 & 0.9808 & 0.9877 & 0.9916 & 0.9872 & 0.9952 \\
   & Top-$K$ vs L2R       & 0.9851 & 0.9808 & 0.9877 & 0.9916 & 0.9872 & 0.9952 \\
   \rowcolor{xxpurple!20}
   \cellcolor{white}
   & Random vs L2R       & 0.8337 & 0.7676 & 0.8470 & 0.8592 & 0.8093 & 0.8717 \\
  \midrule
  \multirow{4}{*}{Dream-base}
   & Entropy vs L2R      & 0.8941 & 0.8406 & 0.8802 & 0.9276 & 0.8933 & 0.9236 \\
   & Top-$K$ Margin vs L2R & 0.8950 & 0.8421 & 0.8802 & 0.9289 & 0.8952 & 0.9250 \\
   & Top-$K$ vs L2R        & 0.8950 & 0.8421 & 0.8802 & 0.9289 & 0.8952 & 0.9250 \\
   \rowcolor{xxpurple!20}
   \cellcolor{white}
   & Random vs L2R       & 0.6761 & 0.5537 & 0.6819 & 0.7870 & 0.7004 & 0.7739 \\
  \midrule
  \multirow{4}{*}{Dream-instruct}
   & Entropy vs L2R      & 0.9441 & 0.9211 & 0.9235 & 0.9656 & 0.9519 & 0.9417 \\
   & Top-$K$ Margin vs L2R & 0.9441 & 0.9211 & 0.9235 & 0.9656 & 0.9519 & 0.9417 \\
   & Top-$K$ vs L2R        & 0.9441 & 0.9211 & 0.9235 & 0.9656 & 0.9519 & 0.9417 \\
   \rowcolor{xxpurple!20}
   \cellcolor{white}
   & Random vs L2R       & 0.7593 & 0.6683 & 0.7545 & 0.8383 & 0.7774 & 0.8359 \\
  \midrule
  \multirow{4}{*}{LLaDA-instruct}
  & Entropy vs L2R        & 0.9757 & 0.9573 & 0.9602 & 0.9760 & 0.9653 & 0.9661 \\
  & Top-$K$ Margin vs L2R & 0.8971 & 0.8407 & 0.8587 & 0.9328 & 0.8934 & 0.9035 \\
  & Top-$K$ vs L2R        & 0.8991 & 0.8488 & 0.8595 & 0.9284 & 0.8981 & 0.9092 \\
  \rowcolor{xxpurple!20}
  \cellcolor{white}
  & Random vs L2R         & 0.7421 & 0.6301 & 0.6771 & 0.8286 & 0.7479 & 0.7838 \\
  \midrule
  \multirow{4}{*}{LLaDA-cpGRPO}
  & Entropy vs L2R        & 0.9033 & 0.8541 & 0.8677 & 0.9320 & 0.9023 & 0.9099 \\
  & Top-$K$ Margin vs L2R & 0.8902 & 0.8341 & 0.8517 & 0.9257 & 0.8915 & 0.9010 \\
  & Top-$K$ vs L2R        & 0.9033 & 0.8541 & 0.8721 & 0.9417 & 0.9123 & 0.9205 \\
  \rowcolor{xxpurple!20}
  \cellcolor{white}
  & Random vs L2R         & 0.7435 & 0.6399 & 0.6921 & 0.8227 & 0.7417 & 0.7636 \\
  \midrule
  \multirow{5}{*}{dUltra-coding-b128}
   & Planner vs L2R     & 0.9731 & 0.9343 & 0.9401 & 0.9772 & 0.9677 & 0.9710 \\
   & Entropy vs L2R      & 1.0000 & 0.9552 & 0.9552 & 1.0000 & 1.0000 & 1.0000 \\
   & Top-$K$ Margin vs L2R & 0.9215 & 0.8513 & 0.8680 & 0.9461 & 0.9198 & 0.9272 \\
   & Top-$K$ vs L2R        & 0.9208 & 0.8483 & 0.8661 & 0.9457 & 0.9199 & 0.9294 \\
   \rowcolor{xxpurple!20}
   \cellcolor{white}
   & Random vs L2R       & 0.5414 & 0.2646 & 0.3054 & 0.8603 & 0.8057 & 0.8356 \\
  \bottomrule
  \end{tabular}
  \caption{MBPP code-similarity to strict L2R baselines using best-match aggregation. Dream, DiffuCoder, LLaDA-instruct, and LLaDA-cpGRPO results are compared against L2R under the Pass@32 setup, while \texttt{dUltra-coding-b128} is evaluated under its Pass@64 setup. Valid pairs require both samples to parse; correct pairs require both samples to pass unit tests.}
  \label{tab:mbpp-similarity-bestmatch-all}
\end{table*}

\newcommand{\cp}{\cellcolor{xxpurple!20}}

\begin{table*}[h]
  \centering
  \small
  \setlength{\tabcolsep}{5pt}
  \begin{tabular}{ll ccc ccc}
  \toprule
   & & \multicolumn{3}{c}{Valid pairs} & \multicolumn{3}{c}{Correct pairs} \\
  \cmidrule(lr){3-5} \cmidrule(lr){6-8}
  Benchmark & Comparison & TSED & ASTD & Coarse & TSED & ASTD & Coarse \\
  \midrule
  \multirow{2}{*}{LCB-v5}
   & Confidence vs L2R & 0.7339 & 0.5960 & 0.5358 & 0.8488 & 0.7598 & 0.7030 \\
   & \cp Random vs L2R & \cp 0.6032 & \cp 0.4075 & \cp 0.3681 & \cp 0.7324 & \cp 0.6098 & \cp 0.6025 \\
  \midrule
  \multirow{2}{*}{LCB-v6}
   & Confidence vs L2R & 0.7272 & 0.5866 & 0.5261 & 0.8488 & 0.7624 & 0.7111 \\
   & \cp Random vs L2R & \cp 0.6000 & \cp 0.4065 & \cp 0.3632 & \cp 0.7431 & \cp 0.6254 & \cp 0.6126 \\
  \bottomrule
  \end{tabular}
  \caption{Dream-Coder-instruct's code-similarity to the L2R baseline on LiveCodeBench v5 and v6, using best-match aggregation. For Dream-Coder, the entropy, top-$K$ margin, and top-$K$ generations on LCB are identical to each other under temperature 0.2; we thus report all three as ``confidence vs L2R'' instead of repeating three times.}
  \label{tab:dclcb-similarity-bestmatch-all}
\end{table*}

\subsection{Semantic collapse in dUltra planner} \label{app:dultra}

\textbf{Experimental setup.} \texttt{dUltra-coding-b128} \citep{chen2025dultra} uses LLaDA as the base masked diffusion language model and augments it with an unmasking planner head. The base LLaDA model predicts the token values, whereas the planner head predicts which masked positions to reveal. The planner is first pretrained with the frozen LLaDA backbone, and then the base model and the planner are jointly optimized with GRPO. The goal is to learn an efficient reveal policy that selects weakly dependent masked positions for parallel unmasking while preserving the token-prediction quality.

Our goal is to test whether the learned dUltra unmasking planner produces structurally different outputs from standard decoding policies. We therefore apply the same similarity analysis as in Section~\ref{sec:similarity} and compare each policy against strict L2R decoding under a Pass@64 evaluation setup. For a controlled comparison, all policies follow the inference configuration of \texttt{dUltra-coding-b128}, using the same LLaDA backbone, prompt formatting, maximum generation length, diffusion step schedule, block size, temperature, candidate sampling budget, and evaluation pipeline. For the planner-based policy, we use the trained dUltra unmasking head to select masked positions for reveal and keep the planner decoding temperature at its default value, $0$. The planner-based sampling scheme still introduces stochasticity through its learned reveal-position selection procedure. For planner-free baselines, we disable the planner head and decode the same LLaDA token predictor with hand-designed reveal policies: Top-$K$, Top-$K$ margin, entropy-based selection, random selection, and strict L2R. In these backbone-only baselines, we use a temperature $0.2$ following the standard sampling setup. Thus, apart from this policy-specific temperature convention, the comparison controls for the backbone, prompts, decoding budget, and evaluation pipeline, and isolates how the reveal-position selection rule affects the resulting code structure. The dUltra experiments follow a Pass@64 protocol and compare all policies against the L2R baseline.

\textbf{Results.} 
Tables~\ref{tab:humaneval-similarity-bestmatch-all} and \ref{tab:mbpp-similarity-bestmatch-all} show that the learned dUltra planner produces outputs that remain highly similar to strict L2R decoding. On both HumanEval and MBPP, planner decoding has substantially higher valid-pair similarity to L2R than the original random-style decoding policy, indicating that the learned planner does not induce a structurally diverse any-order generation pattern. Moreover, the planner is more L2R-like than Top-$K$ and Top-$K$ margin on most similarity metrics. This suggests that explicitly learning an unmasking schedule is not sufficient to avoid semantic collapse, and the planner can still converge to reveal orders whose final code structures closely match those produced by L2R-style decoding.

\section{Details of Section~\ref{sec:pos_uncer}}
\label{app:pos_unc}
\textbf{Experimental setup.} We evaluate positional uncertainty on large-scale masked diffusion language models for code generation: Dream-7B-Base \citep{xie2025dream}, and DiffuCoder-7B-Base, DiffuCoder-7B-Instruct, DiffuCoder-7B-cpGRPO \citep{gong2025diffucoder}, and LLaDA-8B-instruct  \citep{nie2025large}. We use problem prompts from HumanEval \citep{chen2021evaluating} and MBPP \citep{austin2021program}.

For Dream and DiffuCoder, the logged runs use a 256-token canvas and 256 diffusion steps with temperature 0.2 and top-$p=0.95$. LLaDA runs use a 512-token canvas and 512 denoising steps. For the released bundle visualizations, we export point data on the full HumanEval-164 and MBPP-500 task sets and render standalone plots at a step stride of 16.

All generated sequences include non-code text around them, including an instruction/prompt prefix before the code region and a natural language explanation after the code. We therefore extract the code span before computing positional-uncertainty statistics. Since generation is performed at the token level, the generated sequence can be truncated in the middle of a code region or continue beyond the end of the program into natural language text. In such cases, we discard tokens outside the extracted code span and compute all quantities only on the retained in-code token positions. Thus, the masked positions appearing in the definitions of $m(v;\rvx_t)$ and $\mathrm{LOC}(v;\rvx_t)$ are restricted to this trimmed code domain. 

Because storing the full vocabulary distribution at every step is too expensive, we do not log the full per-position vocabulary posterior. Instead, at each analyzed step, we retain up to 200 selected token IDs, chosen primarily by aggregate token mass, and we store their positional mass profiles over the trimmed code domain. Committed token IDs are additionally forced into the logged set when necessary, and committed-local statistics are recorded separately so that committed-position uncertainty can still be computed even when a committed token would otherwise fall outside the generic 200-token selection. Unless otherwise specified, the positional-uncertainty results in this subsection are reported under a Pass@32 setup against the strict L2R baseline.

\begin{figure*}[h]
\centering
\begin{subfigure}[t]{0.32\textwidth}
    \centering
    \includegraphics[width=\linewidth]{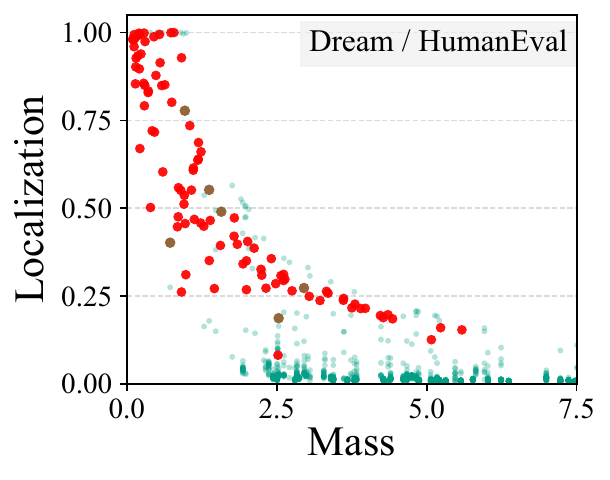}
    \caption{Dream, HumanEval, step 16}
\end{subfigure}
\hfill
\begin{subfigure}[t]{0.32\textwidth}
    \centering
    \includegraphics[width=\linewidth]{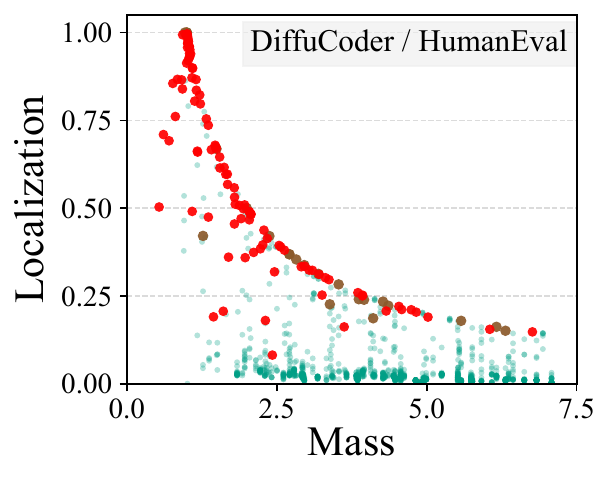}
    \caption{DiffuCoder, HumanEval, step 16}
\end{subfigure}
\hfill
\begin{subfigure}[t]{0.32\textwidth}
    \centering
    \includegraphics[width=\linewidth]{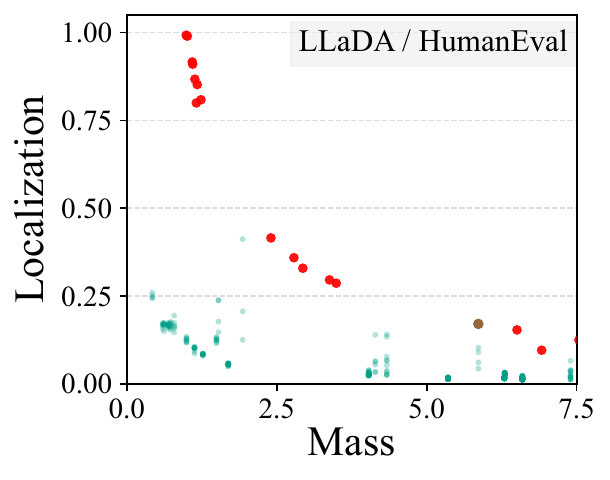}
    \caption{LLaDA, HumanEval, step 32}
\end{subfigure}
\vspace{0.08in}
\begin{subfigure}[t]{0.32\textwidth}
    \centering
    \includegraphics[width=\linewidth]{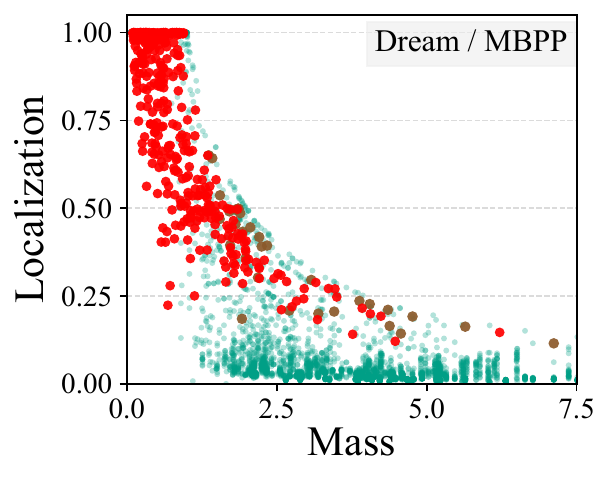}
    \caption{Dream, MBPP, step 16}
\end{subfigure}
\hfill
\begin{subfigure}[t]{0.32\textwidth}
    \centering
    \includegraphics[width=\linewidth]{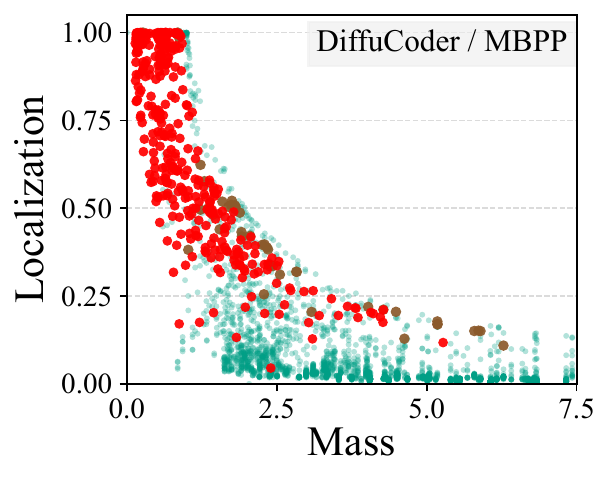}
    \caption{DiffuCoder, MBPP, step 16}
\end{subfigure}
\hfill
\begin{subfigure}[t]{0.32\textwidth}
    \centering
    \includegraphics[width=\linewidth]{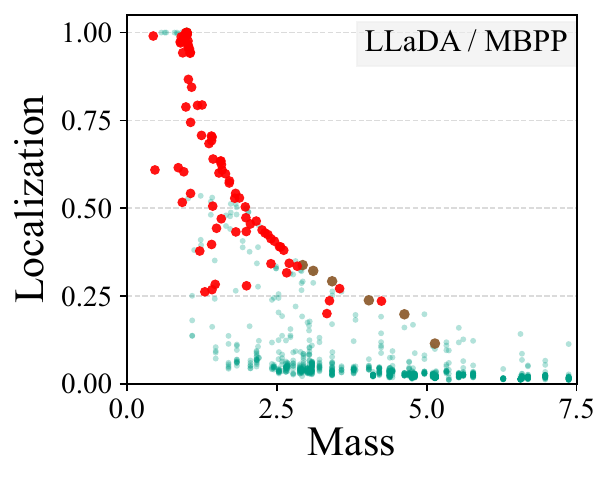}
    \caption{LLaDA, MBPP, step 32}
\end{subfigure}
\vspace{-0.05in}
\caption{Additional positional-uncertainty visualizations across model families and benchmarks. Dream and DiffuCoder are visualized at step 16 out of 256 denoising steps, while LLaDA is visualized at step 32 out of 512 denoising steps. Across settings, high aggregate token mass does not necessarily imply high localization, and committed tokens tend to be more localized.}
\label{fig:pos-unc-additional}
\vspace{-0.10in}
\end{figure*}

\textbf{Additional Results.}
Figure~\ref{fig:pos-unc-additional} shows representative positional-uncertainty plots across model families and benchmarks. Since Dream and DiffuCoder use a 256-step denoising schedule, whereas LLaDA uses a 512-step schedule, we visualize step 16 for Dream and DiffuCoder and step 32 for LLaDA. These choices correspond to the same relative denoising progress, i.e., $16/256 = 32/512$. Across both HumanEval and MBPP, we observe the same qualitative pattern as in Section~\ref{sec:pos_uncer}: many high-mass tokens remain poorly localized, while committed tokens are biased toward higher localization. This supports that positional uncertainty is not specific to a single model family or benchmark.

\begin{figure*}[h]
\centering
\begin{subfigure}[t]{0.32\textwidth}
    \centering
    \includegraphics[width=\linewidth]{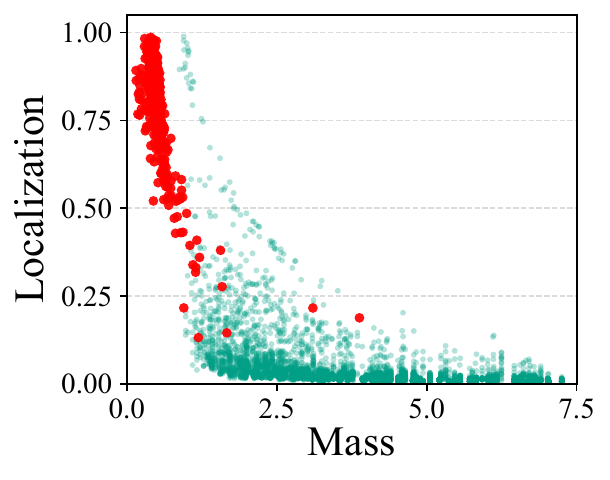}
    \caption{Step 0}
\end{subfigure}
\hfill
\begin{subfigure}[t]{0.32\textwidth}
    \centering
    \includegraphics[width=\linewidth]{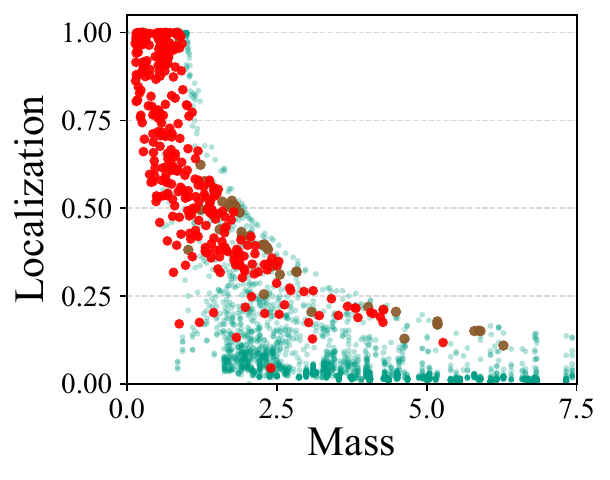}
    \caption{Step 16}
\end{subfigure}
\hfill
\begin{subfigure}[t]{0.32\textwidth}
    \centering
    \includegraphics[width=\linewidth]{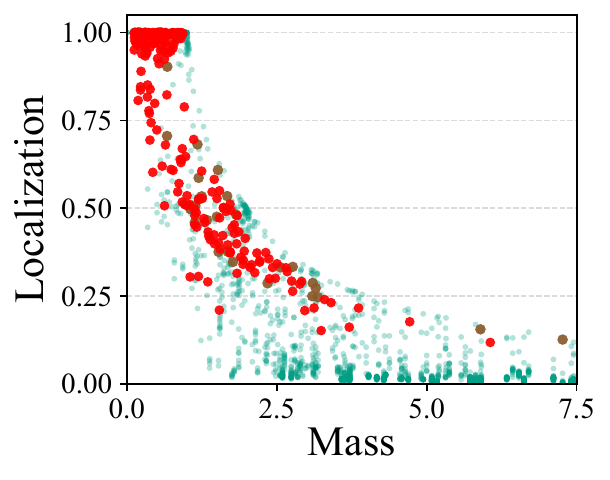}
    \caption{Step 32}
\end{subfigure}
\vspace{0.08in}
\begin{subfigure}[t]{0.32\textwidth}
    \centering
    \includegraphics[width=\linewidth]{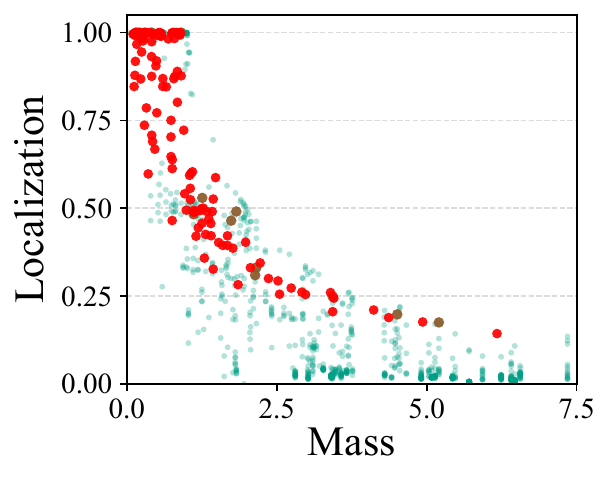}
    \caption{Step 48}
\end{subfigure}
\hfill
\begin{subfigure}[t]{0.32\textwidth}
    \centering
    \includegraphics[width=\linewidth]{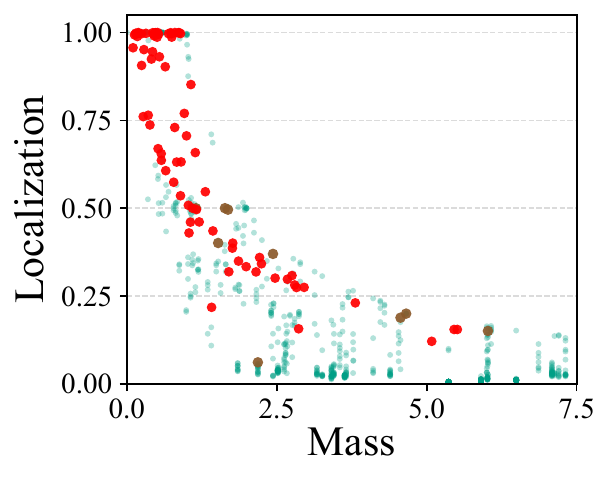}
    \caption{Step 64}
\end{subfigure}
\hfill
\begin{subfigure}[t]{0.32\textwidth}
    \centering
    \includegraphics[width=\linewidth]{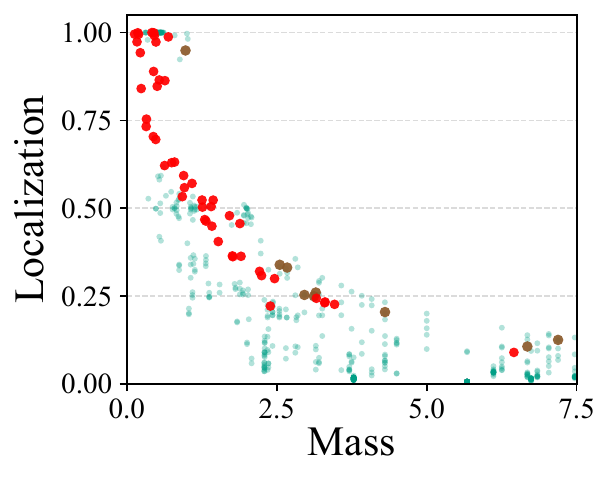}
    \caption{Step 80}
\end{subfigure}
\vspace{-0.05in}
\caption{Step-wise positional-uncertainty visualization for DiffuCoder on MBPP. Each panel plots aggregate token mass against localization at a different denoising step. Across the trajectory, high-mass tokens are not necessarily localized, while committed tokens tend to have higher localization.}
\label{fig:pos-unc-diffucoder-mbpp-steps}
\vspace{-0.10in}
\end{figure*}

We further visualize how positional uncertainty evolves across denoising steps for DiffuCoder on MBPP. Figure~\ref{fig:pos-unc-diffucoder-mbpp-steps} shows the mass-localization plots at steps $0,16,32,48,64,$ and $80$. Across the trajectory, high-mass tokens can remain broadly dispersed over candidate positions, while committed tokens tend to concentrate in regions with higher localization. This step-wise view illustrates that positional uncertainty persists over a substantial portion of the inference trajectory.

\section{Details of Section~\ref{sec:flexmdm}} \label{app:flexmdm}
In this appendix, we provide the theoretical foundation of FlexMDM, details on tree metrics for quantifying any-order inference, and an evaluation pipeline.

\subsection{FlexMDM Training}\label{app:flexmdm_training}
\paragraph{Preliminary theory.} Let $\alpha_t,\beta_t$ be smooth and monotone schedules with $\alpha_0=\beta_0=0$ and $\alpha_1=\beta_1=1$. They will be the insertion and unmasking schedules, respectively. Given a clean sequence $\rvx_1=(\rvx_1^1,...,\rvx_1^L) \sim p_{\mathrm{data}}$ of length $L$, for each position $i\in[L]$, define its insertion and unmasking times $T_1^i$, $T_2^i$ as \[T_1^i\sim \dot\alpha_tdt,\quad T_2^i\sim \mathrm{1}_{t\geq T_1^i}\cdot\frac{\dot\beta_t}{1-\beta_{T_1^i}}dt.\]
For $t\in[0,1]$, define the (sorted) set $s_t$ of inserted indices as \[s_t=\{i\in[L]\mid t\geq T^i_1\}\]
with $s_t[1]<s_t[2]<\cdots<s_t[\mathrm{len}(s_t)]$, and with the boundary convention that $s_t[0]=0$ and $s_t[\mathrm{len}(s_t)+1]=L+1$. Intuitively, this is the set of indices that have been inserted by time $t$. The stochastic interpolant $\rvx_t$, a sequence of length $|s_t|$, is then constructed by the following masking and deletion procedure: \begin{equation*}
    \rvx_t^i=\begin{cases}
        \mask & \text{if } t<T_2^{s_t[i]},\\
        \rvx_1^{s_t[i]} & \text{if } t\geq T_2^{s_t[i]}.
    \end{cases}
\end{equation*}
Similar to a standard MDM, a FlexMDM parametrizes an unmasking posterior $f_\theta(\rvx,t)[i,v]\approx \mathbb P[\rvx_1^{s_t[i]}=v\mid \rvx_t=\rvx]$; however, it also parametrizes an insertion prediction \[g_\theta(\rvx,t)[i]\approx \log\mathbb E[s_t[i]-s_t[i-1]-1\mid \rvx_t=\rvx].\]
Intuitively, this is the log of the expected number of tokens to be inserted between the $i$-th and $(i-1)$-th positions in the current $\rvx_t$.
At training time, we minimize the following loss function: \begin{equation}\label{eq:flexmdm_loss}
\begin{aligned}
\mathcal L(\theta)&=
\underbrace{
\int_{0}^{1}
\mathbb{E}_{\rvx_1,s_t,\rvx_t}
\left[
-\frac{\dot{\beta}_t}{1-\beta_t}
\sum_{i=1}^{\operatorname{len}(\rvx_t)+1}
\mathbf{1}_{\rvx_t^i=\mask}
\log f_{\theta}(\rvx_t,t)[i,\rvx_1^{s_t[i]}]
\right]\,dt
}_{\text{unmasking loss}}
\\[0.5em]
&\quad+
\underbrace{
\int_{0}^{1}
\mathbb{E}_{\rvx_1,s_t,\rvx_t}
\left[
-\frac{\dot{\alpha}_t}{1-\alpha_t}
\sum_{i=1}^{\mathrm{len}(\rvx_1)+1}
\phi\!\left(s_t[i]-s_t[i-1]-1,\,
g_{\theta}(\rvx_t,t)[i]\right)
\right]\,dt
}_{\text{insertion loss}},
\end{aligned}
\end{equation}
where $\phi(x,y)=e^y-xy$.

Notably, this differs from the original setup in \cite{kim2025any}, where the insertion head directly predicts the expected insertion length, whereas we found that empirically predicting in log space yields stabler training. Geometrically, insertion counts live on a positive, multiplicative scale and enter inference as Poisson intensities, so the log map converts variation in expected lengths into an unconstrained additive coordinate. This makes the insertion head predict the natural parameter of the Poisson rate, which better matches a neural network’s real-valued outputs and avoids the boundary/scale issues of directly regressing a nonnegative expectation.

\paragraph{Implementation details.} We choose the following insertion and unmasking schedules: \begin{equation}\label{eq:power_schedules}
    \alpha_t=1-(1-t)^a,\qquad\beta_t=1-(1-t)^{ab},
\end{equation}
where $a=b=1.7$. We explicitly note that this parameterization is part of a more general class of schedules considered in prior work~\citep{patel2026insertion}.

\begin{figure}[h]
\centering
\includegraphics[width=\linewidth]{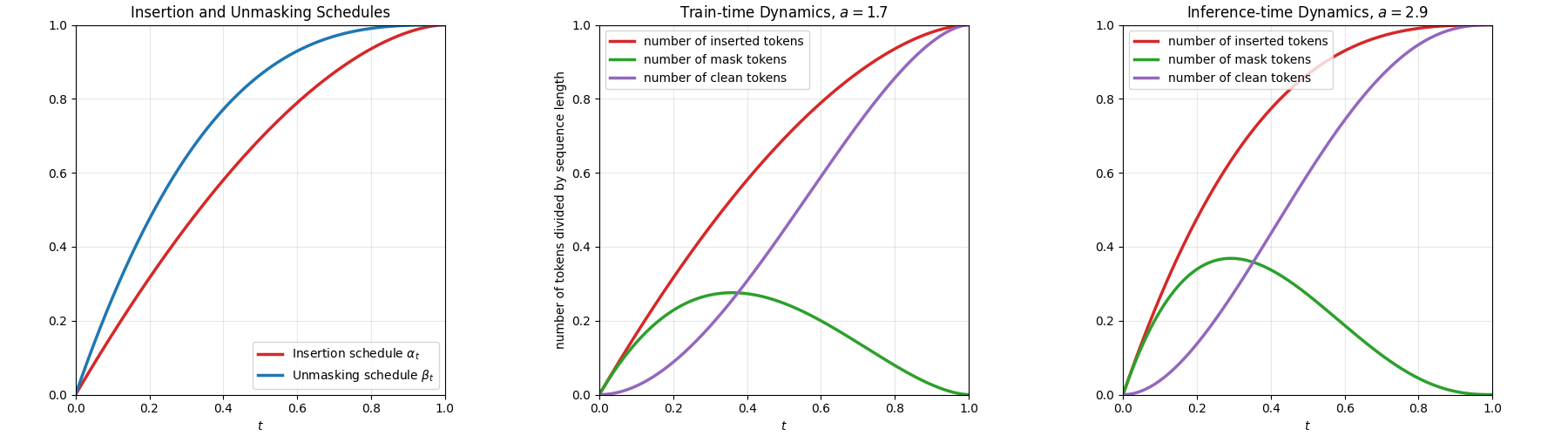}
\caption{\textbf{(Left)} graphs of the insertion and unmasking schedules in \eqref{eq:power_schedules}. \textbf{(Center)} the number of inserted, mask, and clean tokens as a function of time during training, normalized by sequence length. \textbf{(Right)} the number of inserted, mask, and clean tokens as a function of time during inference, normalized by sequence length.}
\label{fig:schedules}
\end{figure}

An important practical observation beyond the original FlexMDM formulation in \cite{kim2025any} is that, at inference time, we are not locked to the insertion schedule used during training. Let $\alpha_t$ denote the insertion schedule used during training, and let $\tilde{\alpha}_t$ denote a desired insertion schedule at inference time. We define the monotone time reparameterization
\[
\tau(t) := \alpha^{-1}(\tilde{\alpha}_t).
\]
At inference step $t$, we query the model at the reparameterized time $\tau(t)$, i.e., we use the predictions $f_{\theta}(x,\tau(t))$ and $g_{\theta}(x,\tau(t))$, while evolving the insertion process with the inference-time schedule $\tilde{\alpha}_t$. In this way, the model is always evaluated at the training-time corruption level whose insertion fraction matches the desired inference-time insertion fraction, since
\[
\alpha_{\tau(t)}=\tilde{\alpha}_t.
\]
Equivalently, the insertion Poisson intensity at gap $i$ is taken to be
\[
\frac{\dot{\tilde{\alpha}}_t}{1-\tilde{\alpha}_t}\exp\bigl(g_{\theta}(x,\tau(t))[i]\bigr),
\]
while the unmasking process continues to use the original unmasking schedule $\beta_t$.

For the power-family schedules in \eqref{eq:power_schedules}, this reparameterization is even more explicit: if the model is trained with $\alpha_t = 1-(1-t)^a$ but we want to sample with $\tilde{\alpha}_t = 1-(1-t)^{\tilde a}$, then
\[
\tau(t)=\alpha^{-1}(\tilde{\alpha}_t)=1-(1-t)^{\tilde a/a}.
\]
This flexibility is practically important. We found that any-order inference requires a sufficiently strong scaffold to be inserted early in the trajectory; otherwise, each currently visible position must summarize too broad a region of the eventual sequence, and token predictions become clogged amalgamations of several plausible future locations. In our experiments, we therefore use a more aggressive insertion schedule at inference time with $\tilde a=2.9$, while keeping the same unmasking schedule. This front-loads scaffold construction \emph{without} retraining the model parameters.

The unmasking and insertion dynamics are visualized in Figure~\ref{fig:schedules}. While we insert tokens more aggressively at the early stage of inference, the number of clean tokens remains sub-linear in this early phase, avoiding premature token commitments while the layout is still uncertain; we also retain a long tail of later insertions so the model can still add missing local details after partial content has been revealed, preserving the main flexibility that mitigates positional uncertainty.

We obtain the training dataset by processing open-source Python data from OpenCodeInstruct~\citep{ahmad2025opencodeinstruct}, rStar-Coder~\citep{liu2025rstarcoder}, KodCode-V1-SFT-4o~\citep{xu2025kodcode}, and opc-sft-stage2-educational~\citep{huang2025opencoder}, resulting in $\approx$2.6M training sequences. For OpenCodeInstruct, we keep only those samples whose \texttt{average\_test\_score} is at least 0.9 and from which valid Python code is extracted, resulting in 1.85M samples out of the total 5M. From rStar-Coder, we use the \texttt{seed\_sft} and \texttt{synthetic\_sft} splits, keeping only the samples verified as passing their tests, which yields 81K and 369K samples respectively. We use the entirety of KodCode-V1-SFT-4o (209K samples) and opc-sft-stage2-educational (118K samples) without additional filtering. During training, we up-sample rStar-Coder, KodCode-V1-SFT-4o, and opc-sft-stage2-educational by a factor of two relative to OpenCodeInstruct.

We initialize from Dream-Coder 7B~\citep{xie2025dream}, while attaching an auxiliary head and an AdaLN time embedding~\citep{peebles2023scalable} for insertion prediction. We use the AdamW~\citep{loshchilov2019decoupled} optimizer with $\beta_1=0.9$, $\beta_2=0.95$, and weight decay 0.01. We use a learning rate schedule with linear warmup and cosine decay, with warmup ratio 0.1 and peak learning rate of $10^{-5}$ for the backbone and $2\times10^{-5}$ for the insertion head. We train with the loss in \eqref{eq:flexmdm_loss} for 50000 optimizer steps, with global batch size 576, which took 3 days on 16 H100's.

\subsection{Training FlexMDM in Insertion-Progress Coordinates}
\label{sec:alpha-coordinate-flexmdm}

The standard FlexMDM construction conditions the model on a raw time variable $t\in[0,1]$, together with an insertion schedule $\alpha_t$ and an unmasking schedule $\beta_t$. We instead propose to condition the model directly on the \emph{insertion progress} \[\tau := \alpha_t \in [0,1].\]
This removes the dependence of the model interface on the particular parameterization of time. Raw time $t$ is then used only as an external sampler clock, while the network always receives the schedule-invariant coordinate $\tau$.

Let $\alpha:[0,1]\to[0,1]$ be a strictly increasing insertion schedule and let $\beta:[0,1]\to[0,1]$ be the unmasking schedule used in the original raw-time formulation. Define the corresponding unmasking schedule in insertion-progress coordinates by \[B(\tau):=\beta\left(\alpha^{-1}(\tau)\right).\]
Thus, after changing variables from $t$ to $\tau$, insertion progress is the identity schedule, while unmasking is governed by $B$.

For a clean sequence $x_1$, let $x_\tau$ denote the partially corrupted sequence at insertion progress $\tau$, and let $s_\tau$ be the alignment map from positions in $x_\tau$ to positions in $x_1$. The model is trained as \[f_\theta(x_\tau,\tau)[i,v]\approx\mathbb P\left(x^1_{s_\tau[i]}=v \mid x_\tau\right),\]
and \[g_\theta(x_\tau,\tau)[i]\approx\log\mathbb E\left[s_\tau[i]-s_\tau[i-1]-1\mid x_\tau\right],\]
where $f_\theta$ predicts clean token identities and $g_\theta$ predicts the log of the expected number of missing tokens in each gap.

In the $\tau$-coordinate, the FlexMDM training objective becomes
\[\begin{aligned}
    \mathcal L(\theta)=\int_0^1\mathbb E\Bigg[&-\frac{B'(\tau)}{1-B(\tau)}\sum_{i:x_\tau^i=\mask}\log f_\theta(x_\tau,\tau)\left[i,x_1^{s_\tau[i]}\right]\\
    &+\frac{1}{1-\tau}\sum_i\phi\left(s_\tau[i]-s_\tau[i-1]-1,\,g_\theta(x_\tau,\tau)[i]\right)\Bigg]\,d\tau.
\end{aligned}
\]
where $\phi(x,y)=e^y-xy$. The insertion weight is $1/(1-\tau)$ because \[\frac{\dot\alpha_t}{1-\alpha_t}\,dt=\frac{1}{1-\tau}\,d\tau,\]
and the unmasking weight is $B'(\tau)/(1-B(\tau))$ because \[\frac{\dot\beta_t}{1-\beta_t}\,dt=\frac{B'(\tau)}{1-B(\tau)}\,d\tau.\]
Thus this objective is simply the original FlexMDM objective written in the coordinate $\tau$.

Equivalently, training data can be generated directly in the $\tau$-coordinate. For each clean token $j$, sample an insertion coordinate \[A_j \sim \mathrm{Unif}(0,1).\]
Conditional on $A_j$, sample an unmasking coordinate $U_j\ge A_j$ by drawing \[Z_j \sim \mathrm{Unif}(B(A_j),1),\qquad U_j = B^{-1}(Z_j).\]
At progress level $\tau$, token $j$ is deleted if $\tau<A_j$, masked if $A_j\le \tau<U_j$, and revealed if $U_j\le \tau$. The model is then given $(x_\tau,\tau)$ rather than $(x_t,t)$.

At inference time, choose any desired event-time insertion schedule \[\Gamma:[0,1]\to[0,1].\]
At sampler time $t$, set $\tau_t:=\Gamma(t)$ and query the model as \[f_\theta(x_t,\tau_t),\qquad g_\theta(x_t,\tau_t).\]
The insertion CTMC in event time is then \[
    R_t^{\mathrm{ins}}(x,x\triangleleft_i m)=\frac{\dot\Gamma(t)}{1-\Gamma(t)}\exp\bigl(g_\theta(x,\Gamma(t))[i]\bigr).
\]
Thus the sampler uses the event-time hazard of the chosen inference schedule, but the model itself is always conditioned on the insertion-progress value $\tau_t=\Gamma(t)$.

With an event-time unmasking schedule $\Delta(t)$, the unmasking transition rate is \[
    R_t^{\mathrm{unmask}}(x,x[i\leftarrow v])=\frac{\dot\Delta(t)}{1-\Delta(t)}f_\theta(x,\Gamma(t))[i,v].
\]
If one wants to preserve the original training unmasking dynamics in $\tau$-coordinates, then the event-time unmasking schedule should just be $\Delta(t)=B(\Gamma(t))$.

The key distinction from a raw-time-conditioned model is that no inverse-time correction is required at inference. A model trained on raw time under $\alpha_{\mathrm{train}}$ must be queried at
\[
    \alpha_{\mathrm{train}}^{-1}(\Gamma(t))
\]
when using a new inference schedule $\Gamma$. In contrast, an insertion-progress-conditioned model is queried directly at
\[
    \tau_t=\Gamma(t).
\]
Therefore changing the insertion schedule at inference only changes how quickly the sampler moves through insertion-progress space; it does not change the semantic meaning of the model's conditioning variable.

\subsection{Any-Order Metrics}\label{app:flexmdm_anyorder}
To study the any-order inference ability of FlexMDM, we evaluate both FlexMDM and Dream-Coder on the three tree-based metrics, each measuring the any-order performance from a different perspective. Below, we first describe how we parse the extracted Python code into a tree, and then define the metrics precisely.
\paragraph{Parsing into trees.}
We base our metrics on the natural syntactic structure already exposed by the Python AST (Abstract Syntax Tree) parser. For each generated sample, we first extract the Python code span and parse it into an \texttt{ast.Module}, using standard fallbacks for fenced code blocks, closing fences, and longest valid prefixes when direct parsing fails. We then coarsen the full Python AST into a simplified statement-level tree: expression-level nodes are collapsed into their nearest containing statement, docstrings are collapsed into their owner node, and trailing generated tests or examples are optionally grouped into a synthetic reference node. Each Python token is assigned to the deepest compatible AST node, so every tree node is associated with the set of generated tokens in its subtree.

\paragraph{Coverage Before Commitment (CBC).}
CBC measures whether the model previews multiple sibling subtrees before committing to finishing one of them. This distinguishes genuinely any-order behavior from a merely serial traversal: at a branching point in the program tree, a model with higher CBC begins work on several child blocks before completing any one.

More precisely, let $v$ be a non-leaf node in the syntax tree with $k$ children ($k\geq1$). We say $v$ is \textit{split} if $k\geq 2$, and \textit{non-split} if $k=1$. Write $\mathrm{child}(v)=\{c_1,\ldots,c_k\}$, where each $c_i$ represents the subtree rooted at the $i$-th child node of $v$. Let $t_{\mathrm{commit}}(v)$ be the first generation step at which any child block $c_i\in\mathrm{child}(v)$ becomes complete, and let $N_{\mathrm{started}}(v)$ be the number of child blocks of $v$ that has at least one generated token by that time. We define
\[
\mathrm{CBC}(v)=\frac{N_{\mathrm{started}}(v)}{k}.
\]

The normalization by $k$ makes CBC comparable across different branching factors and makes the score measure how much of the \emph{available} sibling structure was explored before commitment. At a split node, $\mathrm{CBC}(v)=1/k$ is the strictly sequential baseline, while values closer to $1$ indicate that the model opened most sibling blocks before completing the first one. For a non-split node, there is no branching opportunity in the inherent code structure, so sequential generation is the natural behavior; setting $\mathrm{CBC}(v)=1$ ensures that the metric does not penalize the model for being sequential when the program structure itself is sequential.

Lastly, after computing $\mathrm{CBC}(v)$ for all non-leaf nodes $v$, we aggregate by reporting both the overall average over all non-leaf nodes and the split-only average over nodes with $k\geq 2$; the former measures alignment with the full tree structure, while the latter isolates behavior at genuine branching points.

\paragraph{Return to Unfinished Blocks (RUB).}
RUB measures whether the model returns to a partially generated child block after moving away from it. This captures a stronger form of any-order inference than CBC: rather than only checking whether several sibling blocks were previewed before commitment, RUB asks whether generation actually moves back and forth across unfinished parts of the same local tree region.

More precisely, let $v$ be a non-leaf split node in the syntax tree. For a child block $c_i\in\mathrm{child}(v)$, let $t_{\mathrm{open}}(c_i)$ be the first generation step at which any token in $c_i$ is generated, and let $t_{\mathrm{finish}}(c_i)$ be the first generation step at which this subtree becomes complete. We say that $c_i$ is \textit{returned to} if, after $c_i$ is opened and before it is finished, the model generates at least one token belonging to another sibling block $c_j\in\mathrm{child}(v)$, $j\neq i$. Let $N_{\mathrm{returned}}(v)$ be the number of children of $v$ that are returned to. We define
\[
\mathrm{RUB}(v)=\frac{N_{\mathrm{returned}}(v)}{k}.
\]

The normalization by $k$ again makes RUB comparable across branching factors and measures what fraction of the available sibling blocks are revisited after being left unfinished. For a split node, $\mathrm{RUB}(v)=0$ indicates no back-and-forth behavior across unfinished sibling blocks, while values closer to $1$ indicate that many child blocks were revisited during generation. For a non-split node, there is no alternative sibling block to move to, so we set $\mathrm{RUB}(v)=1$ to avoid penalizing sequential generation when the underlying code structure offers no branching opportunity.

\textbf{Graded Return to Unfinished Blocks (RUB$^+$).} RUB is binary at each child: it saturates after a single return, so it cannot distinguish a trace that opens every sibling once and then closes them in turn from one that alternates between siblings many times. RUB$^+$ refines RUB into a three-level graded score, while capping the credit at two returns so that it does not reward pathological alternation.

More precisely, let $v$ be a non-leaf split node and let $c_i \in \mathrm{child}(v)$. 
Restricting attention to generation steps that reveal a token in some child of $v$, define $\mathrm{visits}(c_i)$ to be the number of maximal contiguous runs of steps during which the model unmasks tokens belonging to $c_i$'s subtree; equivalently, $\mathrm{visits}(c_i) - 1$ counts the number of times the model leaves $c_i$, works on a sibling, and returns. We define the per-child score \[s(c_i)=\frac{\min\bigl(\mathrm{visits}(c_i) - 1,2\bigr)}{2}\in\{0, 0.5, 1\},\]
so that $s(c_i)$ takes value $0$, $0.5$, or $1$ according to whether the model never returns to $c_i$, returns exactly once, or returns at least twice. We then set \[\mathrm{RUB}^{+}(v)=\frac{1}{k} \sum_{i=1}^{k} s(c_i).\]
The cap at two returns reflects the view that genuine any-order generation should exhibit a small number of revisits per branch — the kind a human writing code might exhibit — rather than frantic alternation; replacing the binary ``did the model ever return'' signal with this three-level score lets us separate single-return traces from richer back-and-forth without rewarding pathological interleaving. As with RUB, $\mathrm{RUB}^{+}(v) = 0$ indicates strictly serial generation at $v$, and we set $\mathrm{RUB}^{+}(v) = 1$ for non-split nodes by convention.

Lastly, same as above, we aggregate $\mathrm{RUB}(v)$ and $\mathrm{RUB}^+(v)$ by reporting both the overall average over all non-leaf nodes and the split-only average over nodes with $k\geq 2$.

\paragraph{Open-Block Width (OBW).}
OBW measures how many sibling blocks the model keeps simultaneously unfinished during generation. This captures the local breadth of any-order inference: at a branching point in the program tree, a model with higher OBW maintains progress on several child blocks at once rather than completing one child block before opening the next.

More precisely, let $v$ be a non-leaf node in the syntax tree. For each generation step $t$, let $N_{\mathrm{open}}(v,t)$ be the number of children blocks $c_i\in\mathrm{child}(v)$ that has at least one generated token but is not yet complete at time $t$. We define
\[
\mathrm{OBW}(v)=\frac{\max_t N_{\mathrm{open}}(v,t)}{k}.
\]

Again, the normalization by $k$ makes OBW comparable across branching factors and measures the largest fraction of available sibling blocks that are simultaneously active. For a non-split node, there is only one child block and we automatically get $\mathrm{OBW}(v)=1$, which also avoids penalizing sequential generation when the code structure itself is sequential.

Same as the above two metrics, we report both overall and split-only average.
\newcommand{\numhl}[1]{\cellcolor{xxpurple!20}#1}
\begin{table}[t]
\centering
\small
\setlength{\tabcolsep}{2.5pt}
\resizebox{\linewidth}{!}{%
\begin{tabular}{@{}lllcrrrrrrrr@{}}
\toprule
Benchmark & Model & Temp. & Scope
& \multicolumn{4}{c}{Overall}
& \multicolumn{4}{c}{Split-only} \\
\cmidrule(lr){5-8}\cmidrule(lr){9-12}
& & & & CBC & RUB & RUB$^+$ & OBW & CBC & RUB & RUB$^+$ & OBW \\
\midrule
\multirow{4}{*}{HumanEval}
& \multirow{2}{*}{Dream-Coder}
& \multirow{2}{*}{0.2}
& \texttt{full\_output}
& 0.634 & 0.402 & 0.379 & 0.641
& 0.431 & 0.081 & 0.041 & 0.442 \\

& & 
& \texttt{code\_only}
& 0.742 & 0.596 & 0.583 & 0.745
& 0.388 & 0.066 & 0.034 & 0.396 \\

\cmidrule(lr){2-12}
& \multirow{2}{*}{FlexMDM}
& \multirow{2}{*}{0.1}
& \texttt{full\_output}
& \numhl{0.833} & \numhl{0.780} & \numhl{0.716} & \numhl{0.855}
& \numhl{0.706} & \numhl{0.620} & \numhl{0.510} & \numhl{0.746} \\

& &
& \texttt{code\_only}
& \numhl{0.835} & \numhl{0.780} & \numhl{0.720} & \numhl{0.857}
& \numhl{0.582} & \numhl{0.468} & \numhl{0.329} & \numhl{0.640} \\

\midrule
\multirow{4}{*}{MBPP}
& \multirow{2}{*}{Dream-Coder}
& \multirow{2}{*}{0.1}
& \texttt{full\_output}
& 0.798 & 0.666 & {0.664} & 0.798
& 0.382 & 0.011 & {0.006} & 0.383 \\

& &
& \texttt{code\_only}
& 0.798 & 0.666 & {0.664} & 0.798
& 0.382 & 0.011 & {0.006} & 0.383 \\

\cmidrule(lr){2-12}
& \multirow{2}{*}{FlexMDM}
& \multirow{2}{*}{0.1}
& \texttt{full\_output}
& \numhl{0.876} & \numhl{0.829} & \numhl{0.774} & \numhl{0.892}
& \numhl{0.766} & \numhl{0.691} & \numhl{0.588} & \numhl{0.799}\\
& &
& \texttt{code\_only}
& \numhl{0.876} & \numhl{0.823} & \numhl{0.768} & \numhl{0.891}
& \numhl{0.601} & \numhl{0.464} & \numhl{0.306} & \numhl{0.657} \\
\bottomrule
\end{tabular}%
}
\vspace{0.05in}
\caption{
Tree-based any-order metrics for Dream-Coder-7B-Base and FlexMDM (ours) under both \texttt{full\_output} evaluation, without sanitizing, and \texttt{code\_only} evaluation, after discarding generated test-case nodes. Overall scores average over all non-leaf nodes, while split-only scores average only over branching nodes.
}
\label{tab:flexmdm_any_order_metrics}
\vspace{-0.2in}
\end{table}
\paragraph{Complete any-order metrics results.}
We use the default inference setup in the original Dream-Coder paper \citep{xie2025dream}, with temperature 0.2 on HumanEval and 0.1 on MBPP, and with confidence based on negative entropy on both benchmarks.

We observed that the models sometimes generate some test cases after the function body itself, such as standalone \texttt{print}, \texttt{assert}, or \texttt{\_\_main\_\_} statements, which may confound the evaluation on the actual function body. Thus, we sanitize the code and discard all nodes corresponding to the test cases. We report the any-order metrics both before and after the sanitizing, labeled as \texttt{full\_output} and \texttt{code\_only}, respectively.

As shown in Table~\ref{tab:flexmdm_any_order_metrics}, with both variants, our FlexMDM significantly outperforms Dream-Coder. Notably, the gap on split-only RUB shows that at branching points in the program tree, Dream-Coder almost never revisits a sibling block once it has moved away (split RUB $\approx 0.07$ on HumanEval and $\approx 0.01$ on MBPP), generating in a near-autoregressive manner, whereas FlexMDM revisits roughly half of all sibling branches on HumanEval (split RUB $0.62$ and $0.47$ under \texttt{full\_output} and \texttt{code\_only} respectively). The ratio between split-only RUB$^+$ and split-only RUB further sharpens this picture: a child returned to exactly once contributes $1$ to RUB but only $\tfrac{1}{2}$ to RUB$^+$, while a child returned to at least twice contributes $1$ to both, so this ratio reads off the fraction of returned children that were revisited at least twice. For Dream-Coder the ratio is $\approx 0.50$ across both benchmarks, meaning that on the rare occasions it does return to an unfinished sibling, it does so exactly once; for FlexMDM the ratio rises to $\approx 0.70$–$0.82$, meaning that a substantial share of revisits involve two or more genuine returns rather than a single one-pass ABA excursion.

\subsection{FlexMDM Inference and Details on Evaluation Setup} \label{app:flexmdm_eval}
At inference time, we follow the standard FlexMDM inference algorithm with top-K for unmasking, as detailed in Alg.~\ref{alg:flexmdm_inference}. For all models, we use 512 sampling steps; Dream-Coder generates up to 512 new tokens, while FlexMDM caps the total sequence length (prompt plus generation) at 768 for HumanEval and 1100 for MBPP. For Dream-Coder, since using low temperature with the confidence-based sampling strategy collapses to one single generation, we use temperature 1.0 for all baseline models for diversity at pass@$k$. For FlexMDM, due to the inherent stochasticity in the insertion, we use a token temperature of 0.1; the insertion temperature (Appendix~\ref{app:insertion_temp}) is $T_{\mathrm{ins}}=0.6$ on MBPP/MBPP+ and the neutral $T_{\mathrm{ins}}=1$ on HumanEval/HumanEval+. We use the same code extraction and grading criteria from Dream-Coder's official codebase across all models and benchmarks, executing each sample against the benchmark's full test suite under a 30-second wall-clock limit. Results are summarized in Table~\ref{tab:flexmdm_full_passk}.

\textbf{Discussion on FlexMDM's downstream performance.}
FlexMDM does not dominate Dream-Coder-7B uniformly: it trails at low $k$ on HumanEval and by ${\le}1.6$ points at $k\ge2$ on MBPP. Two factors are at play. First, Dream-Coder-7B-Base is already a strong model — the best or second-best on HumanEval and MBPP among comparable baselines~\citep{xie2025dream} — so the headroom left to fine-tuning is limited. Second, the token temperature never touches the insertion process, whose stochasticity perturbs \emph{where} code is laid out and therefore matters most at small $k$. This axis has its own dial: as Appendix~\ref{app:insertion_temp} shows, sharpening insertion placement improves every Pass@$k$ on MBPP/MBPP+, whereas on HumanEval we keep the neutral setting, trading Pass@1 for the placement diversity that drives the Pass@16 gains.

\begin{figure}[!t]
\captionsetup{type=algorithm}
\centering
\begin{minipage}[t]{0.55\textwidth}
\vspace{-0.15in}
\begin{algorithm}[H]
\caption*{\textbf{Subroutine 1:} FlexMDM inference}
\begin{algorithmic}[1]
\Require Learned functions $(f_\theta, g_\theta)$
\Require Discretization $0 = t_1 < \dots < t_N = 1$
\Require Insertion, Unmasking schedule $\alpha_t,\beta_t$
\State Initialize \(X_{t_1} \gets \texttt{empty\_sequence}\)
\For{$j=1$ to $N-1$}
    \State \(\tau \gets t_{j+1} - t_j\)
    \State \textbf{Invoke Subroutine 2 for unmasking}
    \For{\xxgreen{\textbf{\(i\) in $[\mathrm{len}(X_{t_j})]+1$}}}
        \State Set insertion rate $r_\alpha\gets\tfrac{\dot{\alpha}_{t_j}}{1-\alpha_{t_j}}\cdot\tau$
        \State Sample $\ell\sim\mathrm{Poi}\big(r_\alpha \cdot \exp(g_\theta(X_{t_j}, t_j)[i]) \big)$
        \State \xxgreen{\textbf{Insert $\ell$ masks between $X_{t_j}^{i-1}$ and $X_{t_j}^i$}}
    \EndFor
\EndFor
\State \Return $X_{t_N}$
\end{algorithmic}
\end{algorithm}
\end{minipage}\hfill
\begin{minipage}[t]{0.43\textwidth}
\vspace{-0.15in}
\begin{algorithm}[H]
\caption*{\textbf{Subroutine 2: }Unmasking step (top-$k$)}
\begin{algorithmic}[1]
\State Set unmasking rate $r_\beta\gets \tfrac{\dot{\beta}_{t_j}}{1-\beta_{t_j}}\cdot\tau$
\State Sample $k\sim \mathrm{Poi}(r_\beta\cdot |\{i\mid X_{t_j}^i=\mask\}|)$
\For{$i \in \{i\mid X_{t_j}^i=\mask\}$}
\State Sample $v^i \sim \mathrm{Cat}(f_\theta(X_{t_j},t_j)[i])$
\State Compute $\mathcal{C}^i=f_\theta(X_{t_j},t_j)[i,v^i]$
\vspace{-0.025in}
\EndFor
\vspace{-0.03in}
\For{\(i\) in $\mathrm{argmaxk}(\mathcal{C})$}
\State  Commit $X_{t_j}^i\gets v^i$
\EndFor
\end{algorithmic}
\end{algorithm}
\end{minipage}
\caption{\textbf{FlexMDM inference.} At each step we perform \xblue{\textbf{unmasking}} and \xxgreen{\textbf{insertion}}. For \xblue{\textbf{unmasking}}, we use confidence-based top-$k$ selection. The number of mask tokens to \xxgreen{\textbf{insert}} and the number of tokens to \xblue{\textbf{unmask}} are drawn from a Poisson distribution. \textbf{Notation}: $\mathrm{Cat}$, $\mathrm{Poi}$ denote the categorical and Poisson distribution, respectively. $\mathrm{argmaxk}(\mathcal{C})$ is the indices set of the $k$ largest components of $\mathcal{C}$. Adapted from \cite{kim2025any}.} \label{alg:flexmdm_inference}
\end{figure}

\begin{table}[t]
\centering
\small
\setlength{\tabcolsep}{5pt}
\renewcommand{\arraystretch}{1.08}
\begin{tabular}{llccccc}
\toprule
Benchmark & Model & pass@1 & pass@2 & pass@4 & pass@8 & pass@16 \\
\midrule
\multirow{2}{*}{HumanEval}
  & FlexMDM (ours)        & 50.65 & 66.60 & 78.69 & 86.86 & 92.07 \\
  & Dream-Coder-7B-Base   & 58.65 & 72.22 & 81.34 & 86.94 & 90.85 \\
\midrule
\multirow{2}{*}{HumanEval+}
  & FlexMDM (ours)        & 46.61 & 61.89 & 73.83 & 82.07 & 87.80  \\
  & Dream-Coder-7B-Base   & 53.89 & 67.05 & 75.96 & 81.53 & 85.37 \\
\midrule
\multirow{2}{*}{MBPP}
  & FlexMDM (ours)        & 64.70 & 76.73 & 83.80 & 88.20 & 91.27 \\
  & Dream-Coder-7B-Base   & 64.53 & 77.33 & 84.66 & 89.29 & 92.86 \\
\midrule
\multirow{2}{*}{MBPP+}
  & FlexMDM (ours)        & 54.98 & 66.11 & 73.06 & 77.38 & 80.69 \\
  & Dream-Coder-7B-Base   & 53.98 & 66.13 & 73.55 & 78.25 & 82.01 \\
\bottomrule
\end{tabular}
\vspace{0.05in}
\caption{
Full pass@k results for FlexMDM and Dream-Coder-7B-Base on HumanEval, HumanEval+, MBPP, and MBPP+. FlexMDM decodes with insertion temperature $T_{\mathrm{ins}}=0.6$ on MBPP/MBPP+ and $T_{\mathrm{ins}}=1$ on HumanEval/HumanEval+ (Appendix~\ref{app:insertion_temp}).}
\label{tab:flexmdm_full_passk}
\end{table}

\subsection{Insertion Temperature} \label{app:insertion_temp}
Just as the token temperature shapes the unmasking posterior, the insertion process carries its own natural temperature. At each step, the insertion step of Alg.~\ref{alg:flexmdm_inference} (Subroutine 1) draws an independent Poisson count in every gap $i$ with rate $\lambda_i = r\,e^{g_i}$, where $r$ is the scalar schedule hazard and $g_i = g_\theta(\rvx_t,t)[i]$. By Poisson superposition and thinning, this is exactly equivalent to first drawing the \emph{total} number of insertions $N \sim \mathrm{Poisson}(\Lambda)$, $\Lambda = r\sum_i e^{g_i}$, and then \emph{placing} the $N$ masks multinomially with probabilities $p=\mathrm{softmax}(g)$. The single head $g_\theta$ thus encodes two separately tunable quantities — \emph{how many} tokens to insert and \emph{where} — and we define the \emph{insertion temperature} $T_{\mathrm{ins}}$ by tempering the placement while holding the total fixed:
\[
\lambda_i(T_{\mathrm{ins}}) \;=\; \Lambda\cdot \mathrm{softmax}(g/T_{\mathrm{ins}})_i,
\]
which reduces to $\lambda_i$ at $T_{\mathrm{ins}}=1$. Because the placement distribution is normalized for every $T_{\mathrm{ins}}$, the expected number of insertions — and hence the length statistics the model was trained on — is invariant to the knob. Lowering $T_{\mathrm{ins}}$ makes structural commitments more decisive by concentrating insertions in the gaps the model is most confident about; raising it diversifies placement. Count preservation is what makes placement the right axis to temper: naively rescaling the rates as $e^{g_i/T}$ changes the expected length (the softmax denominator is exactly the missing normalization), while suppressing the stochasticity of the count channel instead drives the sampler off the stochastic-insertion process it was trained on — in our experiments collapsing generation length. The knob is inference-only and adds no compute.

Table~\ref{tab:insertion_temp_sweep} sweeps $T_{\mathrm{ins}}$ under the identical protocol as Table~\ref{tab:flexmdm_full_passk}, and the two benchmark families respond in opposite ways, tracking their prompt structure. MBPP prompts give a one-line description with no code scaffold, so the model must lay out the program structure itself, and placement noise perturbs exactly these least-reversible early commitments. Sharpening to $T_{\mathrm{ins}}=0.6$ accordingly improves on the neutral setting at \emph{every} $k$: it is best or tied at every $k$ on MBPP, within half a point of the sweep-best on MBPP+, and lifts Pass@1 by $+2.5$ (MBPP) and $+2.1$ (MBPP+) — past Dream-Coder-7B-Base. HumanEval prompts instead pin the scaffold in advance (signature, docstring, worked examples), so placement stochasticity acts as useful exploration over valid realizations — the source of the Pass@16 advantage; sharpening recovers less than a point of Pass@1 while giving up performance at larger $k$, so the neutral $T_{\mathrm{ins}}=1$ is preferable. We therefore select $T_{\mathrm{ins}}$ per benchmark in Table~\ref{tab:flexmdm_full_passk} ($0.6$ on MBPP/MBPP+, $1$ on HumanEval/HumanEval+), just as token temperatures are routinely chosen per benchmark~\citep{xie2025dream}. In the language of Section~\ref{sec:pos_uncer}, $T_{\mathrm{ins}}$ is a dial on the exploration–commitment trade-off of any-order inference: it sharpens placement precisely in the high-positional-uncertainty regime where the prompt does not anchor it.

\begin{table}[t]
\centering
\small
\setlength{\tabcolsep}{5pt}
\renewcommand{\arraystretch}{1.05}
\begin{tabular}{llccccc}
\toprule
Benchmark & $T_{\mathrm{ins}}$ & pass@1 & pass@2 & pass@4 & pass@8 & pass@16 \\
\midrule
\multirow{5}{*}{MBPP}
  & \textit{Dream-Coder-7B-Base} & 64.53 & 77.33 & 84.66 & 89.29 & 92.86 \\
  & 1.0 & 62.22 & 74.61 & 81.81 & 86.19 & 89.68 \\
  & 0.75 & 62.85 & 75.72 & 83.22 & 87.69 & 91.27 \\
  & \textbf{0.60} & \textbf{64.70} & \textbf{76.73} & \textbf{83.80} & \textbf{88.20} & \textbf{91.27} \\
  & 0.50 & 64.30 & 75.84 & 82.41 & 86.42 & 89.42 \\
\midrule
\multirow{5}{*}{MBPP+}
  & \textit{Dream-Coder-7B-Base} & 53.98 & 66.13 & 73.55 & 78.25 & 82.01 \\
  & 1.0 & 52.86 & 64.38 & 72.04 & 77.00 & 80.16 \\
  & 0.75 & 53.54 & 65.26 & 72.85 & 77.44 & 80.69 \\
  & \textbf{0.60} & \textbf{54.98} & \textbf{66.11} & \textbf{73.06} & \textbf{77.38} & \textbf{80.69} \\
  & 0.50 & 55.47 & 66.36 & 73.29 & 77.38 & 80.16 \\
\midrule
\multirow{3}{*}{HumanEval}
  & \textit{Dream-Coder-7B-Base} & 58.65 & 72.22 & 81.34 & 86.94 & 90.85 \\
  & \textbf{1.0} & \textbf{50.65} & \textbf{66.60} & \textbf{78.69} & \textbf{86.86} & \textbf{92.07} \\
  & 0.70 & 51.64 & 67.02 & 77.90 & 85.49 & 91.46 \\
\midrule
\multirow{3}{*}{HumanEval+}
  & \textit{Dream-Coder-7B-Base} & 53.89 & 67.05 & 75.96 & 81.53 & 85.37 \\
  & \textbf{1.0} & \textbf{46.61} & \textbf{61.89} & \textbf{73.83} & \textbf{82.07} & \textbf{87.80} \\
  & 0.70 & 47.45 & 62.20 & 73.31 & 81.51 & 87.80 \\
\bottomrule
\end{tabular}
\vspace{0.05in}
\caption{Insertion-temperature sweep for FlexMDM (16 samples/task, identical protocol to Table~\ref{tab:flexmdm_full_passk}); Dream-Coder-7B-Base in italics for reference. \textbf{Bold} marks the per-benchmark setting adopted in Table~\ref{tab:flexmdm_full_passk}. Sharpening to $0.6$ helps at every $k$ on MBPP/MBPP+; on HumanEval/HumanEval+, where prompts already fix the scaffold, sharpening recovers little Pass@1 and costs performance at larger $k$.}
\label{tab:insertion_temp_sweep}
\end{table}

\section{Details of Section~\ref{sec:lmdm}}
\label{app:latent_mdm}
\subsection{Data Pre-processing}
\label{app:latent_mdm_data}

We use TinyGSM~\citep{liu2023tinygsm} as the training corpus for all models in Section~\ref{sec:lmdm}. Each example consists of a prompt $P$ and a target Python solution $\rvx$. For LatentMDM, we segment each target solution using \texttt{\textbackslash newline} as a delimiter, treating each line of code as a semantic segment.
We fix the maximum number of segments to $L_s=16$ during training; examples with more than $L_s$ segments are truncated, while examples with fewer segments are padded with empty all-\texttt{<EOS>} segments.

Each segment is tokenized and padded to a maximum length of $L_{\mathrm{seg}}=32$. If a segment exceeds this length, we truncate it and replace the final token with \texttt{<EOS>}. We define $\operatorname{len}(\rvy^i)$ as the position of the first \texttt{<EOS>} token in segment $\rvy^i$, which determines the effective length used in the loss and segment-level scoring during inference.

For the autoregressive and token-level MDM baselines, we do not apply segmentation.
Instead, we concatenate the prompt and target solution, truncate the resulting sequence to 512 tokens, and pad shorter sequences with \texttt{<EOS>}. All models use the Qwen tokenizer~\citep{yang2025qwen3} with vocabulary size 151{,}645.

\subsection{Architecture}
\label{app:latent_mdm_architecture}

\begin{figure}[h]
\centering
\includegraphics[width=0.85\linewidth]{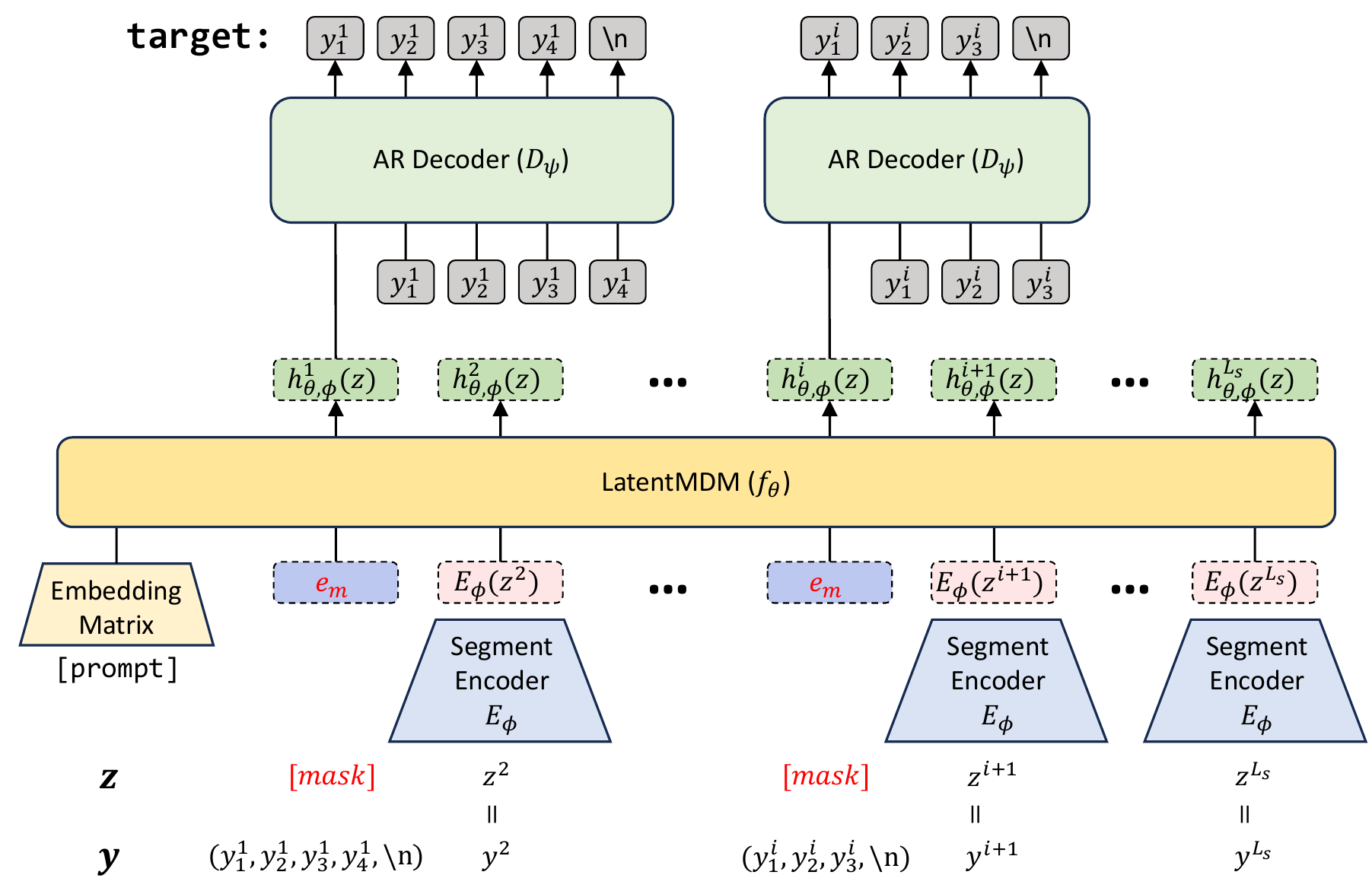}
\caption{\textbf{LatentMDM training pipeline.}}
\label{fig:LatentMDM training}
\vspace{-10pt}
\end{figure}

We use a 125M-parameter architecture for LatentMDM and all token-space baselines, following the setup of \citep{kim2026stop}. All Transformer blocks follow the Qwen2-style design~\citep{yang2025qwen3}. We modify the attention mask according to each module: the segment encoder, LatentMDM, and token-level MDM baseline use bidirectional attention, while the autoregressive decoder and AR baseline use causal attention. Unless otherwise specified, all modules use a hidden dimension $H=512$ and rotary positional embeddings (RoPE).

\textbf{LatentMDM.}
The LatentMDM consists of a segment encoder $E_\phi$, a bidirectional latent Transformer $f_\theta$, and an autoregressive segment decoder $D_\psi$.
The segment encoder $E_\phi$ is a bidirectional Transformer followed by mean pooling over non-\texttt{<EOS>} tokens, without an additional projection layer.
Unlike LaDiR~\citep{kang2025ladir} and LD4PG~\citep{lovelace2023latent}, which use learnable query embeddings to compress variable-length sequences, we use mean pooling for simplicity.
The LatentMDM $f_\theta$ takes the encoded segment sequence as input, replaces masked segments with a learned mask embedding $e_m \in \mathbb{R}^H$, and uses in-context conditioning by concatenating prompt token embeddings as a prefix to the segment-latent sequence.
The autoregressive decoder $D_\psi$ is a causal Transformer that reconstructs each segment autoregressively, using in-context conditioning on the corresponding LatentMDM output $\mathbf{h}_{\theta,\phi}^i(\rvz) \in \mathbb{R}^{H'}$, which is prepended as a one-token prefix.
We set $H'=H=512$ for all three modules, so no projection layer is required between the encoder, LatentMDM, and decoder.

For LatentMDM and all baselines, we tie the token embedding matrix to the output prediction head.
The same embedding matrix is shared across the segment encoder input embedding, the LatentMDM prompt embedding, the decoder input embedding, and the decoder output head.
Since the Qwen tokenizer has a vocabulary size of 151{,}645, embedding tying substantially reduces the parameter count and makes the comparison across models more controlled.

\textbf{Baselines.}
The autoregressive baseline is a 14-layer Qwen2-style causal Transformer, and the token-level MDM baseline is a 14-layer Qwen2-style bidirectional Transformer.
Both baselines use the same tokenizer, hidden dimension, RoPE positional encoding, and tied input/output embeddings as the LatentMDM.

\subsection{Training Details}
\label{app:latent_mdm_training}
Figure~\ref{fig:LatentMDM training} illustrates the resulting training pipeline.
For each training example, the target solution is split into variable-length code segments, and a random subset of segment positions is masked.
Unmasked segments are passed through the segment encoder $E_\phi$, whereas masked positions are represented by the learned mask embedding $\mathbf{e}_{\mask}$.
The LatentMDM $f_\theta$ then performs bidirectional contextual prediction over the prompt-conditioned segment-latent sequence, producing one latent conditioning vector for each masked segment.
Finally, the autoregressive decoder $D_\psi$ reconstructs each masked segment from its predicted latent representation using teacher forcing.
Thus, the model learns latent-segment-level masked prediction while retaining token-level autoregressive modeling within each segment.

In the training setup, we never mask the prompt $\mathbf{P}$ for either LatentMDM or the token-level MDM baseline. For LatentMDM, masking is applied only at the segment level over the target solution. Namely, we sample $n \sim \mathrm{Uniform}\{1,\ldots,L_s\}$ and uniformly choose $n$ segment positions to mask among the $L_s$ target segments. If a segment is selected for masking, it is not passed through the segment encoder $E_\phi$; instead, its latent representation is directly replaced by the learned mask embedding $\mathbf{e}_{\mask}$.
Unmasked segments are encoded by $E_\phi$ and provided to the LatentMDM together with the prompt prefix. Training LatentMDM took approximately three days on 4 80GB NVIDIA A100 GPUs. We defer the remaining optimization hyperparameters and training configurations to Table~\ref{tab:latentmdm-hyperparameters}.

For the token-level MDM baseline, masking is applied only to ground-truth response tokens, including \texttt{<EOS>} tokens, while prompt tokens remain unmasked. Concretely, we sample $r \sim \mathrm{Uniform}(0,1)$ and mask $\lceil r \cdot (512-\operatorname{len}(\mathbf{P})) \rceil$ tokens uniformly among the response-token positions. This matches the LatentMDM setup in that the conditional prompt is always visible, and the model is trained to reconstruct only the target solution.

\subsection{Inference Details}
\label{app:latent_mdm_inference}

\begin{figure}[!t]
\captionsetup{type=algorithm}
\centering
\begin{minipage}[t]{0.55\textwidth}
\vspace{-0.15in}
\begin{algorithm}[H]
\caption*{\textbf{Subroutine 1:} LatentMDM parallel decoding}
\label{alg:latent_mdm_inference}
\begin{algorithmic}[1]
\State \textit{Require:} LatentMDM \xblue{$f_\theta$}, segment encoder \xxgreen{$E_\phi$}, autoregressive decoder \xred{$D_\psi$}, prompt $\mathbf{P}$, number of segments $L_s$, reveal size $k$
\State Initialize latent sequence $\rvz \gets (\mathbf{e}_\mask,\ldots,\mathbf{e}_\mask)$
\State Initialize decoded segments $\hat{\rvy} \gets (\varnothing,\ldots,\varnothing)$
\State Set $N_s \gets L_s/k$
\For{$n = 1$ to $N_s$}
    \State $\mathcal{M} \gets \{i \mid \rvz^i = \mathbf{e}_\mask\}$
    \State $\mathbf{h} \gets \xblue{f_\theta(\rvz;\mathbf{P})}$
    \For{$i \in \mathcal{M}$} \Comment{Parallel over masked segments}
        \State \textcolor{gray}{\texttt{\# Sample tokens autoregressively}}
        \State \textcolor{gray}{\texttt{\# and compute its score}}
        \State $\hat{\rvy}^{i}, s^i \gets \textsc{SampleSegment}(\xred{D_\psi},\mathbf{h}^i)$
    \EndFor
    \State Sample top-$k$ indices $\mathcal{S} \subseteq \mathcal{M}$ according to $s^i$
    \For{$i \in \mathcal{S}$} \Comment{Parallel over selected segments}
        \State $\rvz^i \gets \xxgreen{E_\phi(\hat{\rvy}^i)}$
    \EndFor
\EndFor
\State \Return $\hat{\rvy} = (\hat{\rvy}^1,\ldots,\hat{\rvy}^{L_s})$
\end{algorithmic}
\end{algorithm}
\end{minipage}\hfill
\begin{minipage}[t]{0.43\textwidth}
\vspace{-0.15in}
\begin{algorithm}[H]
\caption*{\textbf{Subroutine 2:} \textsc{SampleSegment}}
\label{alg:ar_segment_sampling}
\begin{algorithmic}[1]
\Function{SampleSegment}{$D_\psi, \mathbf{h}^i$}
    \State Initialize: $\hat{\rvy}^i_0 \gets \varnothing$, $s^i \gets 0$, $l=0$
    \For{$t = 1$ to $L_{\mathrm{seg}}$}
        \State $l \gets l+1$
        \State $\mathbf{p}_t \gets \xred{D_\psi(\cdot \mid \hat{\rvy}^i_{<t}, \mathbf{h}^i)}$
        \State Sample $\hat{\rvy}^i_t \sim \mathrm{Cat}(\mathbf{p}_t)$
        \State $s^i \gets s^i + \log \mathbf{p}_t[\hat{\rvy}^i_t]$
        \If{$\hat{\rvy}^i_t = \texttt{<EOS>}$}
            \State \textbf{break}
        \EndIf
    \EndFor
    \State \Return $\hat{\rvy}^i, s^i/l$
\EndFunction
\end{algorithmic}
\end{algorithm}
\end{minipage}
\caption{\textbf{LatentMDM inference.} Subroutine 1 performs parallel segment-level decoding.}
\label{alg:latent_mdm_parallel_decoding}
\vspace{-0.2in}
\end{figure}

We further describe the LatentMDM inference procedure in a more general setting.
In Section~\ref{sec:lmdm}, we present the case where segments are revealed one at a time.
More generally, the same procedure can reveal multiple masked segments per iteration, analogous to multi-token decoding in token-level MDMs.
Decoding more than one segment per step can improve sampling speed, but introduces the usual speed--quality trade-off in diffusion language model sampling~\citep{zheng2024masked,kim2025train,peng2025pathplanningmaskeddiffusion,ben2025accelerated,nie2025large,wu2025fast,hayakawa2025demystifying}.

Algorithm~\ref{alg:latent_mdm_parallel_decoding} gives the resulting inference procedure with $k$-segment parallel decoding.
For simplicity, we assume that $k$ divides $L_s$ and define $N_s = L_s/k$.
At each iteration, the LatentMDM predicts contextual latent representations for all currently masked segment positions.
The autoregressive decoder then tentatively decodes a candidate segment for each masked position, in parallel across segment positions.
We score each candidate by its length-normalized token log-likelihood:
\begin{equation*}
s^i
\coloneqq
\frac{1}{\mathrm{len}(\hat{\rvy}^i)}
\sum_{j=1}^{\mathrm{len}(\hat{\rvy}^i)}
\log D_\psi\!\left(
\hat{\rvy}^i_j
\mid
\hat{\rvy}^i_{<j}, \mathbf{h}_{\theta,\phi}^i(\rvz)
\right).
\end{equation*}
We then commit the top-$k$ masked segments according to this score, re-encode the committed segments with $E_\phi$, and update the corresponding positions in the latent sequence.
We use the length-normalized token log-likelihood as the default segment-selection score, and ablate both alternative scoring rules and $k$ in Appendix~\ref{app:latent_mdm_additional_experiments}.
KV caching is enabled for the autoregressive decoder and for the autoregressive baseline during inference.
We use $L_s=8$ segment positions at inference time, while keeping the remaining settings consistent with training.
For wall-clock measurements, we use a single 80GB NVIDIA A100 GPU and a sampling batch size 1.

\subsection{Additional Experiments}
\label{app:latent_mdm_additional_experiments}

\begin{table}[h]
  \centering
  \small
  \setlength{\tabcolsep}{6pt}
  \begin{tabular}{llccc}
  \toprule
  Model & Segment-selection strategy & $k=1$ & $k=2$ & $k=4$ \\
  \midrule
  \rowcolor{xxpurple!20}
  \multirow{5}{*}{\cellcolor{white}LatentMDM}
   & Avg. log-likelihood & \textbf{45.5} & \textbf{33.8} & \textbf{20.8} \\
   & Min. log-likelihood & 45.0 & 30.3 & 15.2\\
   & First log-likelihood & 41.2 & 28.6 & 12.7\\
   & L2R & 41.3 & 28.0 & 12.1\\
   & Random & 40.3 & 33.1 & 20.3\\
  \bottomrule
  \end{tabular}
  \vspace{5pt}
  \caption{
  Zero-shot GSM8K performance of LatentMDM under different segment-selection policies and parallel decoding sizes ($k$). Token sampling within each segment is performed greedily $(T=0)$.
  }
  \label{tab:latentmdm-segment-scoring-parallel-ablation}
\end{table}

\textbf{Ablation on Segment Scoring.} 
We evaluate LatentMDM under several segment-selection policies in a zero-shot setting using greedy token sampling.
Avg. log-likelihood is our default scoring rule, which ranks each candidate segment by its length-normalized token log-likelihood.
First log-likelihood scores a segment using only the log-likelihood of its first generated token, following the primary slot-scoring strategy used in ReFusion~\citep{li2025refusion}.
Min. log-likelihood assigns each segment the minimum token log-likelihood within the generated segment, thereby emphasizing the least confident local decision.
L2R reveals segments strictly from left to right, removing the any-order segment-selection mechanism, while Random selects masked segments uniformly at random.
Table~\ref{tab:latentmdm-segment-scoring-parallel-ablation} shows that average log-likelihood performs best, suggesting that reliable segment-level confidence requires aggregating information across the whole generated segment.

\textbf{Parallel Decoding.}
We further evaluate LatentMDM under parallel segment decoding, where $k \in \{1,2,4\}$ denotes the number of masked segments committed at each decoding iteration.
As shown in Table~\ref{tab:latentmdm-segment-scoring-parallel-ablation}, the average log-likelihood score achieves the best performance across all values of $k$.
In contrast, the alternative scoring variants degrade more sharply as $k$ increases, and often fall behind random segment selection in the more parallel regimes.
This suggests that effective parallel segment decoding requires a robust segment-level confidence estimate; otherwise, committing multiple segments per iteration can amplify early selection errors.

\begin{table}[h]
  \centering
  \small
  \setlength{\tabcolsep}{5pt}
  \begin{tabular}{lccc}
  \toprule
   & Encoder $E_\phi$ & LatentMDM $f_\theta$ & Decoder $D_\psi$ \\
  \midrule
  Model size & 6.8M & 34.1M+77.6M & 6.8M \\
  Hidden dim & 512 & 512 & 512 \\
  MLP dim & 1536 & 1536 & 1536 \\
  Number of layers & 2 & 10 & 2 \\
  Number of heads & 8 & 8 & 8 \\
  Number of KV heads & 8 & 8 & 8 \\
  Attention mask & Bidirectional & Bidirectional & Causal \\
  RMSNorm epsilon & $10^{-6}$ & $10^{-6}$ & $10^{-6}$ \\
  Dropout & 0.0 & 0.0 & 0.0 \\
  \midrule
  Tokenizer & \multicolumn{3}{c}{Qwen/Qwen2-0.5B} \\
  Max segment number & \multicolumn{3}{c}{16} \\
  Max segment length & \multicolumn{3}{c}{32} \\
  \midrule
  Optimizer & \multicolumn{3}{c}{AdamW} \\ 
  Learning rate & \multicolumn{3}{c}{$3 \times 10^{-4}$} \\
  Warmup steps & \multicolumn{3}{c}{1000} \\
  Weight decay & \multicolumn{3}{c}{0.01} \\
  EMA value & \multicolumn{3}{c}{0.9999} \\
  Training Steps & \multicolumn{3}{c}{500k} \\
  Max grad norm & \multicolumn{3}{c}{1.0} \\
  Batch size per GPU & \multicolumn{3}{c}{64} \\
  Number of GPUs  & \multicolumn{3}{c}{4} \\
  \bottomrule
  \end{tabular}
  \vspace{5pt}
  \caption{
  Training hyperparameters and architecture details for LatentMDM on TinyGSM. The reported LatentMDM model size separates the Transformer parameters from the tied token-embedding matrix; the additional 77.6M parameters correspond to this shared embedding matrix.
  }
  \label{tab:latentmdm-hyperparameters}
\end{table}

\subsection{Generation Trace}
\begin{figure}[h]
\centering
\includegraphics[width=1.0\linewidth]{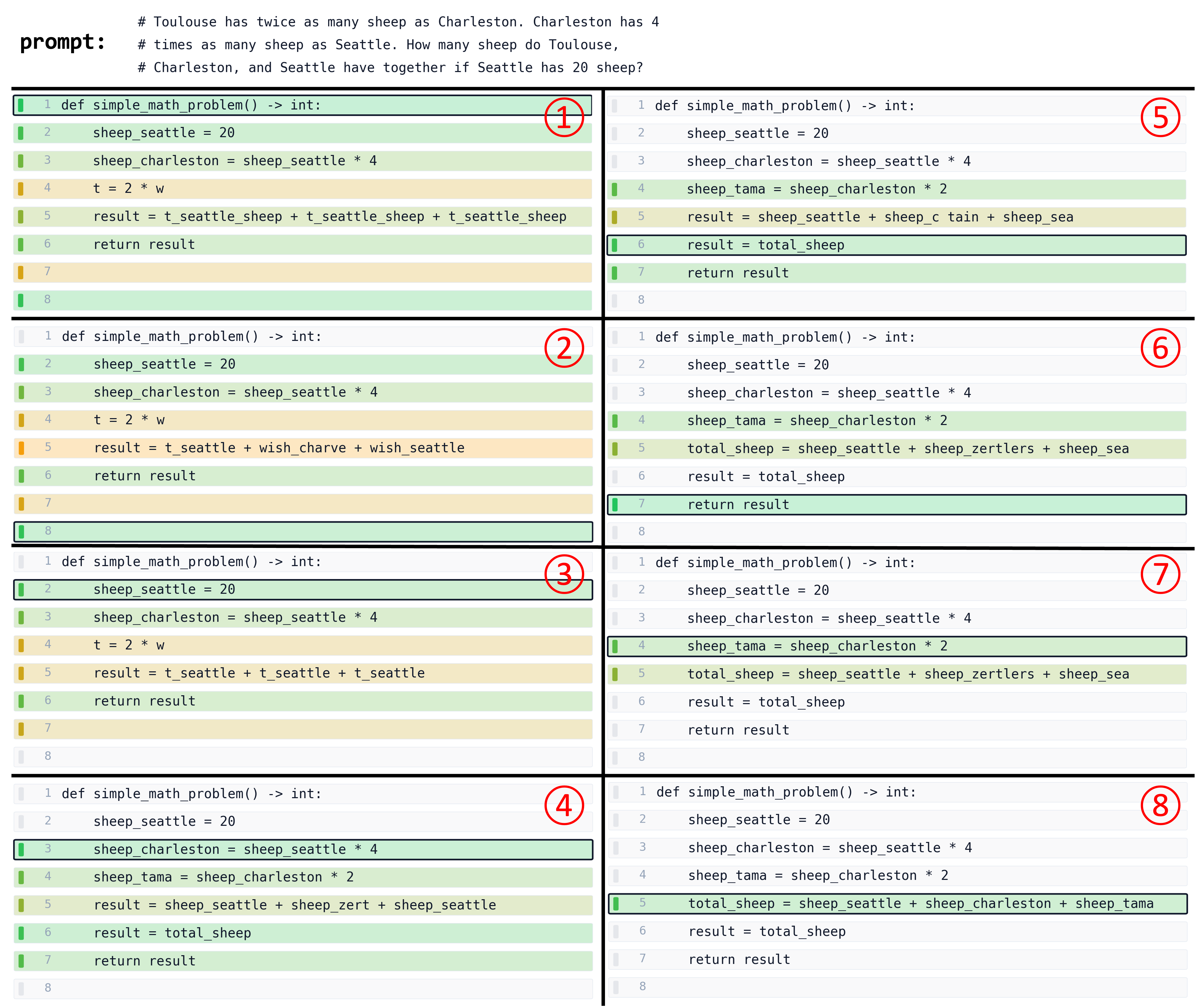}
\caption{\textbf{Generation trace.}
At each iteration, the model tentatively decodes candidate segments for all masked positions, scores them using the segment-selection criterion, and commits the highest-scoring segment.
The highlighted row indicates the segment committed at the current iteration, while previously revealed segments are shown in gray.
The row color reflects the segment score, with stronger colors corresponding to higher-scoring candidates.
This example illustrates that LatentMDM generates at the level of semantic code segments rather than individual token positions, enabling a non-left-to-right segment reveal order.}

\label{fig:latentmdm_generation_trace}
\end{figure}

We further provide a generation trace of LatentMDM in Fig.~\ref{fig:latentmdm_generation_trace}.

\clearpage
\end{document}